%% file: aurora.tex
\documentclass{custom}

\input{math_commands.tex}

\usepackage{float}
\usepackage{url}
\usepackage{amsfonts}
\usepackage{nicefrac}
\usepackage{colortbl}
\usepackage{wrapfig}
\usepackage{array}
\usepackage{tikz}
\usepackage{xspace}

\newcommand{\answerTODO}[1][]{\textcolor{red}{\bf [TODO]}}
\newcommand{\method}{Aurora\xspace}
\newcommand{\bench}{AgentEdit-Bench\xspace}

\definecolor{auroraA}{HTML}{1F4A8A}
\definecolor{auroraB}{HTML}{3D4FAD}
\definecolor{auroraC}{HTML}{6850C2}
\definecolor{auroraD}{HTML}{9650C7}
\definecolor{auroraE}{HTML}{C24FA8}
\definecolor{auroraF}{HTML}{D4528A}
\newcommand{\auroramark}{%
  \textcolor{auroraA}{A}%
  \textcolor{auroraB}{u}%
  \textcolor{auroraC}{r}%
  \textcolor{auroraD}{o}%
  \textcolor{auroraE}{r}%
  \textcolor{auroraF}{a}%
}
\definecolor{benchOurs}{HTML}{E6EDF7}
\definecolor{benchReference}{HTML}{F3F4F6}
\definecolor{benchReferenceText}{HTML}{5E6670}
\definecolor{benchHeader}{HTML}{D9E1EF}
\definecolor{benchAccent}{HTML}{001E5F}
\definecolor{benchBestText}{HTML}{001445}
\newcommand{\benchgroup}[1]{\textbf{\textcolor{benchAccent}{#1}}}
\newcommand{\bestscore}[1]{\textbf{\textcolor{benchBestText}{#1}}}
\newcommand{\refscore}[1]{\textcolor{benchReferenceText}{#1}}
\definecolor{benchTargetClr}{HTML}{001E5F}
\definecolor{benchRegionClr}{HTML}{1B5E20}
\newcommand{\benchTarget}[1]{\textbf{\textcolor{benchTargetClr}{#1}}}
\newcommand{\benchRegion}[1]{\emph{\textcolor{benchRegionClr}{#1}}}
\newcolumntype{L}[1]{>{\raggedright\arraybackslash}p{#1}}

\title{\auroramark{}: Unified Video Editing with a Tool-Using Agent}

\author[1]{Yongsheng Yu}
\author[1]{Ziyun Zeng}
\author[1]{Zhiyuan Xiao}
\author[1]{Zhenghong Zhou}
\author[2]{Hang Hua}
\author[3]{Wei Xiong}
\author[1]{Jiebo Luo}
\affiliation[1]{UofR}
\affiliation[2]{MIT-IBM}
\affiliation[3]{NVIDIA}
\metadata[Code]{\url{https://github.com/yeates/Aurora}}
\metadata[Homepage]{\url{https://yeates.github.io/Aurora-Page}}

\abstract{%
  Recent video editing models have converged on a unified conditioning design: a single diffusion transformer jointly consumes text, source video, and reference images, and one set of weights covers replacement, removal, style transfer, and reference-driven insertion. The design is flexible, but it assumes that the user already provides model-ready text, reference images, and spatial grounding for local edits, which real requests often omit. We present \method{}, an agentic video editing framework that pairs a tool-augmented vision-language model (VLM) agent with a unified video diffusion transformer. The VLM agent maps a raw user request to a structured edit plan aligned with the transformer's conditioning channels, thereby resolving textual and visual underspecification before generation. We train the VLM agent with supervised data for complete edit planning and reference-image selection, together with preference pairs for robust tool use and instruction refinement. We introduce \bench{} to evaluate agent-enhanced video editing under textual and visual underspecification. Experiments on \bench{} and two existing video editing benchmarks show that \method{} improves over instruction-only baselines and that the VLM agent transfers to compatible frozen video editing models.
}

\begin{document}

\input{figs/filmstrip}
\filmstripdraftfalse

\maketitle

\begin{figure}[H]
  \centering
  \vspace{-2mm}
  \includegraphics[width=0.975\linewidth]{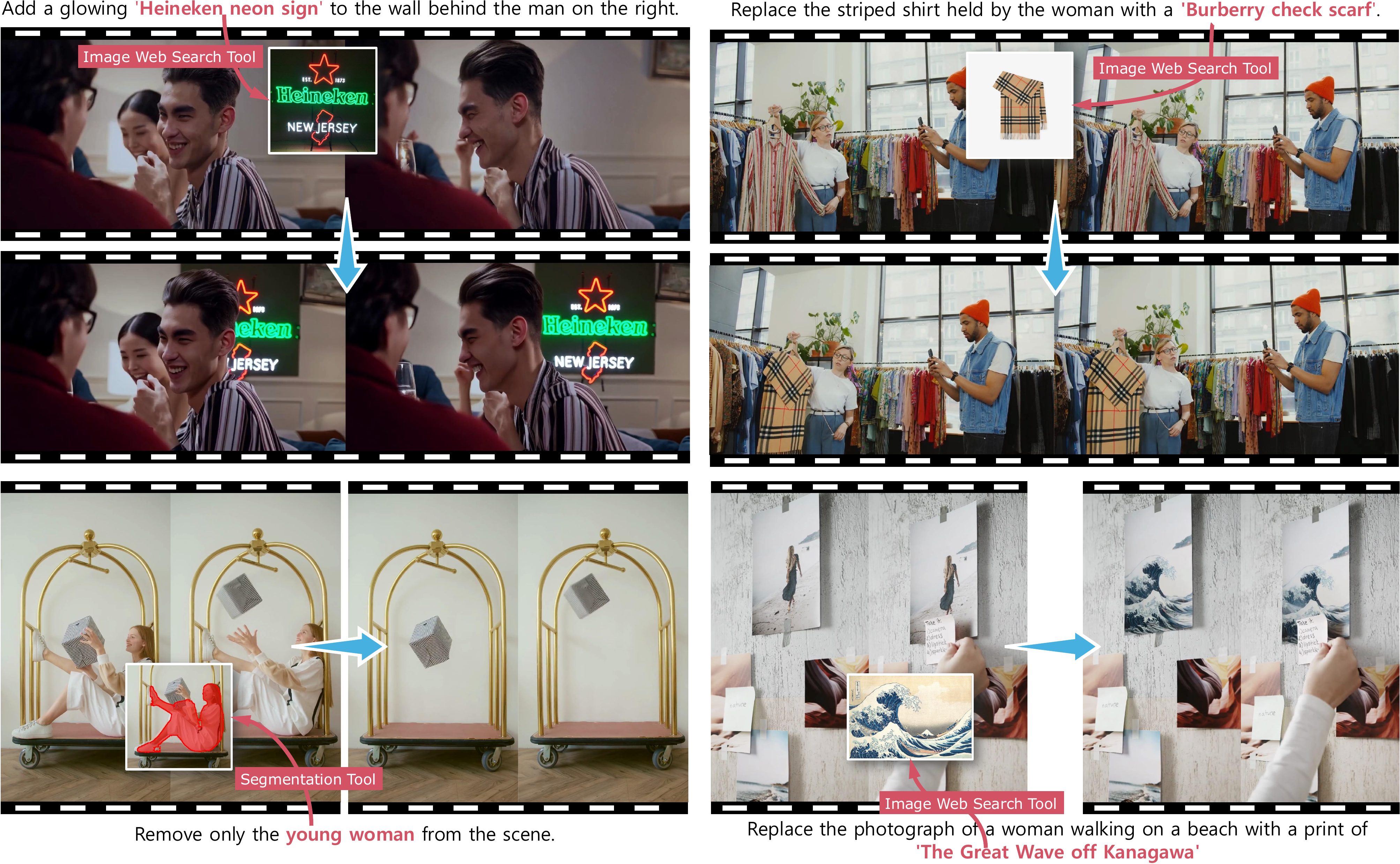}
  \caption{\textbf{\method{}} converts underspecified instructions into model-ready video-editing conditions. The VLM agent invokes web image search to recover the missing reference images for a Heineken neon sign, a Burberry check scarf, and a print of The Great Wave off Kanagawa, and invokes segmentation to recover the missing mask for a person removal. }
  \label{fig:teaser_examples}
\end{figure}

\section{Introduction}
\label{sec:intro}

Recent video editing models increasingly accept several conditioning inputs in one model: text specifies the edit, the source video supplies content to preserve, and optional reference images or masks specify appearance and location~\citep{univideo,editverse,kiwi,videocof}. This design lets one set of weights support replacement, removal, style transfer, and reference-driven insertion, but it also assumes that each edit request is provided with the appropriate conditioning inputs, including a precise text instruction, a reference image that specifies the visual characteristics of a requested real-world entity or visual pattern, and a mask when the edit is spatially localized. However, in practice, user requests are often incomplete with respect to these conditioning channels. We categorize this incompleteness along two axes. \emph{Visual underspecification} occurs when the instruction specifies an edit target or region without providing the visual input needed to ground it. For example, \emph{``Replace the striped shirt held by the woman with a Burberry check scarf''} names a specific visual pattern but provides no reference image. \emph{Textual underspecification} occurs when the desired edit is described only indirectly, as in \emph{``Change the blue item to a vehicle that runs much faster''}, where the model must infer the concrete editing goal. Thus, unified conditioning video editing models provide channels for text, reference images, and masks, but they do not completely interpret how missing conditions should be inferred, grounded, or filled from underspecified user requests. Figure~\ref{fig:teaser_examples} shows underspecified instructions that need additional conditions.

This gap is best understood as a condition-construction problem. Unified video editing models already define the conditioning tuple they expect, but what remains unresolved is how to construct that tuple from a user's raw request. The missing piece may be a more complete instruction, a reference image that specifies the visual characteristics of a requested real-world entity or visual pattern, or spatial grounding for a local edit. Each arises before video generation begins. Accordingly, we decouple condition construction from video generation.

\begin{figure}[t]
  \centering
  \includegraphics[width=\linewidth]{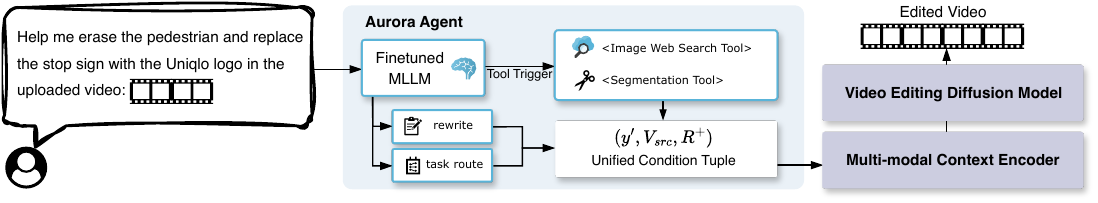}
  \caption{\textbf{\method{} pipeline.} The Aurora VLM agent (a LoRA-finetuned Qwen3-VL-8B-Instruct~\citep{qwen3vl}) parses the raw user request, triggers tools, rewrites the instruction, and routes the task. The output is the unified conditioning tuple $(y', V_{src}, R^+)$ that the \method{} video diffusion transformer consumes; the task label is part of the agent plan and supports the instruction rewrite, but is not fed to the video model.}
  \label{fig:teaser_pipeline}
\end{figure}

We present \method{}, an agentic video editing framework built around this split between the VLM agent and video model, shown in Figure~\ref{fig:teaser_pipeline}. Our design has two parts. First, a tool-augmented VLM agent maps a source video and a raw user request to a four-field edit plan: a rewritten instruction, a task label, an optional image-search query, and an optional mask phrase. The first two fields resolve textual underspecification through rewriting and task routing; the latter two recover missing visual conditions through web image search and grounded segmentation. When the source clip and the user's text already specify the edit, the tool fields remain empty and the framework reduces to ordinary instruction-conditioned editing.

Second, \method{} uses a unified video diffusion transformer (DiT) whose conditioning paths match those plan fields. Built on Wan2.2-TI2V-5B~\citep{wan}, the \method{} video DiT consumes the resulting conditions along two paths: a shared multimodal context for the rewritten instruction, sampled source frames, and reference images, and a latent token sequence where source and reference tokens accompany the noisy video tokens. The video DiT itself does not rewrite instructions or call tools. It consumes the conditioning tuple produced by the VLM agent.

We also introduce \bench{}, a video editing benchmark designed to evaluate this two-part framework jointly rather than each part in isolation. Each case is built around a request that is underspecified along either the textual or the visual axis, so the framework succeeds only if it fills in the missing visual input or instruction text with a concrete editing goal. This structure supports three evaluation protocols on the same source clips: our VLM agent with our video DiT, our VLM agent with any unified video editing model, and any video editing model under the raw user instruction alone. The first measures the full framework, the second isolates VLM agent transfer across video editing models, and the third measures how much the VLM agent helps over an instruction-only baseline. Our contributions are as follows.

\begin{itemize}[leftmargin=*]
  \item We present \method{}, an agentic video editing framework that pairs a tool-augmented VLM agent with a unified video diffusion transformer, so that raw user requests are mapped to more complete and model-ready conditions before generation.
  \item We construct VLM agent training data for complete edit planning, reference-image selection, and DPO-based preference alignment.
  \item We introduce \bench{}, a benchmark designed to evaluate video editing enhanced by a tool-augmented VLM agent under textual and visual underspecification.
\end{itemize}

\section{Related Work}
\label{sec:related}

\subsection{Instruction-conditioned and unified video editing}
\label{sec:related:video_edit}

Early instruction-conditioned video editing methods adapted image-editing techniques to video by sharing attention across frames~\citep{fatezero, stdf, tokenflow}. Senorita-2M~\citep{senorita} scales supervision with task-specialist synthesizers, and OpenVE~\citep{openve3m} and Ditto~\citep{ditto} place VLMs in the data-construction pipeline to generate instructions and filter low-quality edits. Recent methods move toward unified video editing models that accept richer conditioning inputs, including instructions, source videos, reference images, masks, in-context examples, or reasoning frames~\citep{vace,kiwi,univideo,editverse,videocof}. As conditioning becomes richer, the bottleneck shifts from what the model can ingest to what the user can reliably provide. Their conditioning interfaces accept references and masks as inputs, but the burden of producing those inputs, when the raw request omits them, is left to the user or to upstream pipelines. \method{} closes this gap with a trained tool-augmented VLM agent that reads the raw request, retrieves a reference image when needed, grounds a mask when a spatial region must be localized, and rewrites the instruction into a model-ready form, then hands a complete conditioning tuple to the video editing model.

\subsection{Agentic visual generation and video agents}
\label{sec:related:agents}

The closest related work is in image generation, where language-model agents retrieve external knowledge or reference images, decompose user requests, choose tools, and score intermediate results with visual feedback or learned rewards~\citep{gensearcher, agentbanana, jarvisevo, photoagent,zeng2025mira}. Their shared pattern is to keep planning outside the generator, so that the generator receives more complete conditions. Existing video-domain agents target complementary settings: generalist video generation, compositional text-to-video generation, model selection, or multi-step workflows over model libraries~\citep{yuan2024mora, huang2026genmac, tu2026spagent, liang2025univauniversalvideoagent,hua2026mmig}. The \method{} VLM agent addresses a different problem in this design space: rather than coordinating a library of video models, it translates a raw instruction into the conditions required by a single unified video DiT.

\section{Method}
\label{sec:method}

\method{} pairs a tool-augmented VLM agent with a unified video DiT.
\method{} uses two separate Qwen-family checkpoints: the VLM agent is a
LoRA-finetuned Qwen3-VL-8B-Instruct~\citep{qwen3vl} that emits the edit
plan, and the multimodal context encoder inside the video DiT is a frozen
Qwen3.5-4B~\citep{qwen35} that turns the plan and source video into
cross-attention tokens. The two share no weights and play distinct roles
throughout the paper.
We write $\pi_\phi$ for the VLM agent, where $\phi$ collects the trainable
parameters (the LoRA adapter on top of Qwen3-VL-8B-Instruct described in
Sec.~\ref{sec:method:conditioning}). Given a source video
$V_{\mathrm{src}}\in\mathbb{R}^{F\times H\times W\times 3}$, a raw user
instruction $y$, and an optional user reference-image set
$R=\{r_1,\ldots,r_K\}$, the target edited video $V_{\mathrm{edit}}$
should realize the requested change while preserving the identity,
motion, and background structure of $V_{\mathrm{src}}$ wherever the
instruction does not ask for a change. The method has three parts. The
VLM agent first maps $(V_{\mathrm{src}},y,R)$ to a complete edit plan and
constructs concrete conditions from optional tool calls. The resulting
conditions are then assembled into a unified conditioning tuple used by
the video DiT. Finally, the video DiT is trained with flow matching,
while the VLM agent is trained with supervised planning and preference
alignment.

\begin{figure}[t]
  \centering
  \includegraphics[width=\linewidth]{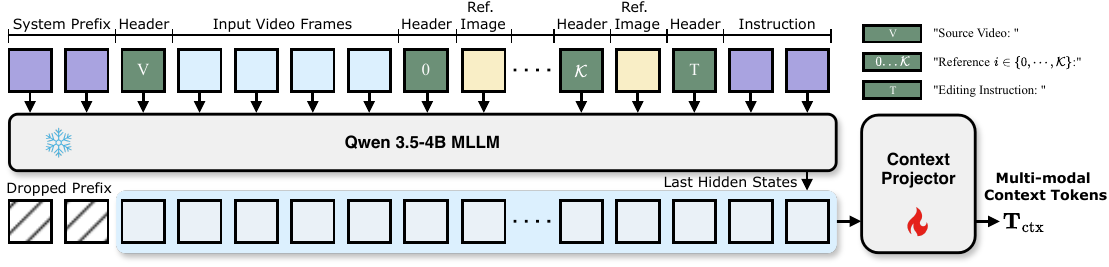}
  \caption{\textbf{Multimodal context encoder.} A frozen Qwen3.5-4B MLLM encodes the instruction, sampled source frames, and reference images into a shared multimodal context that feeds DiT cross-attention.}
  \label{fig:mmencoder}
\end{figure}

\subsection{Agent planning and condition construction}
\label{sec:method:conditioning}

The VLM agent first predicts a typed edit plan:
\begin{equation}
  (y',c,q,m)=\pi_\phi(V_{\mathrm{src}},y,R),
  \label{eq:agent_plan}
\end{equation}
where $y'$ is the rewritten instruction used by the video DiT,
$c\in\mathcal{C}$ is a task label,
$q\in\Sigma^\ast\cup\{\emptyset\}$ is an optional image-search query, and
$m\in\Sigma^\ast\cup\{\emptyset\}$ is an optional mask phrase. We
constrain the output space so that user-provided references suppress web
image search, and a non-empty search query is allowed only for
search-eligible tasks such as addition, replacement, and background change.
The plan fields separate the two forms of underspecification introduced
in Section~\ref{sec:intro}. The rewritten instruction $y'$ standardizes
the free-form user request into a model-ready form that stays consistent
with the task label $c$. The image-search query $q$ retrieves a reference
image that specifies the visual characteristics of a requested real-world
entity or visual pattern, and the mask phrase $m$ localizes spatial
regions described in the instruction.

Only $q$ and $m$ trigger external tools. We use $\mathcal{T}$ for these
tool operators. Both produce reference-image assets of the same type as
the user references in $R$, so the downstream video DiT does not need a
separate mask branch. $\mathcal{T}_{\mathrm{search}}$ denotes web image
retrieval (via the Serper API~\citep{serper}) followed by
reference-image selection, returning a single selected image.
$\mathcal{T}_{\mathrm{mask}}$ denotes grounded segmentation followed by
mask compositing onto $V_{\mathrm{src}}$, returning a masked image of the
same resolution. A tool-execution stage converts the non-empty plan
fields into these reference assets and merges them with any user
references:
\begin{equation}
  R^+ =
  R
  \cup \mathcal{T}_{\mathrm{search}}(q)
  \cup \mathcal{T}_{\mathrm{mask}}(m,V_{\mathrm{src}}).
  \label{eq:condition_construction}
\end{equation}
An empty plan field denotes no tool call: we adopt the convention
$\mathcal{T}_{\mathrm{search}}(\emptyset)=\emptyset$ and
$\mathcal{T}_{\mathrm{mask}}(\emptyset,\cdot)=\emptyset$, so $q=\emptyset$
or $m=\emptyset$ contributes nothing to $R^+$. After condition
construction, the inputs that the video DiT consumes are gathered into a
conditioning tuple:
\begin{equation}
  \mathbf{x}=(y',V_{\mathrm{src}},R^+).
  \label{eq:conditioning_tuple}
\end{equation}
We use $\mathbf{x}$ for this unified conditioning tuple throughout the
method. The task label $c$ is not fed to the video DiT. 

\begin{wrapfigure}{r}{0.53\columnwidth}
  \vspace{-1.1\baselineskip}
  \centering
  \includegraphics[width=0.9\linewidth]{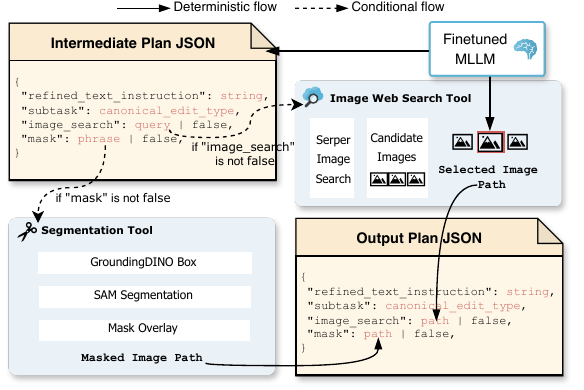}
  \vspace{-1mm}
  \caption{\textbf{Protocol for tool-augmented condition construction. }} %
  \label{fig:agent_tool_protocol}
  \vspace{-1.4\baselineskip}
\end{wrapfigure}
Figure~\ref{fig:agent_tool_protocol} summarizes this protocol, which
turns an intermediate plan into complete conditions. When image search
is triggered (the value \texttt{image\_search} is not empty), a candidate
set is retrieved via the Serper API~\citep{serper} and a reference-image
selection step keeps a single image. When mask generation is triggered
(the value \texttt{mask} is not empty), GroundingDINO~\citep{groundingdino}
and Segment Anything~\citep{sam2} localize the target and composite a
masked image onto $V_{\mathrm{src}}$. Both outputs are inserted into
$R^+$ as reference images. The VLM agent architecture itself is
deliberately simple: Qwen3-VL-8B-Instruct with a lightweight LoRA
adapter~\citep{qwen3vl, hu2022lora}, generating the
four-field JSON output directly. The key design choice is the fixed
four-field output format, and VLM agent training does not update the
downstream video DiT.

\subsection{Unified conditioning in the video DiT}
\label{sec:video_dit}

\begin{figure}[t]
  \centering
  \includegraphics[width=\linewidth]{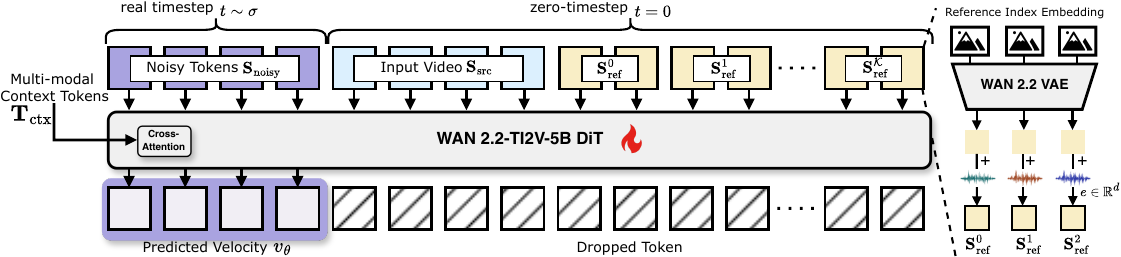}
  \caption{\textbf{\method{} video DiT.} The video DiT predicts denoising velocity from noisy video tokens while attending to source and reference tokens that are concatenated along the token-sequence dimension under zero-timestep modulation.}
  \label{fig:dit}
\end{figure}

Built on Wan2.2-TI2V-5B~\citep{wan}, the \method{} video DiT consumes the
unified conditioning tuple $\mathbf{x}$ along two conditioning paths. A
multimodal context sequence carries the rewritten instruction together with
sampled source frames and reference images into cross-attention, while a
latent token sequence carries noisy video tokens together with source and
reference tokens into self-attention. Figures~\ref{fig:mmencoder}
and~\ref{fig:dit} summarize these two paths.

\textbf{Multimodal context encoder.}
We use a frozen Qwen3.5-4B multimodal encoder~\citep{qwen35} to encode the
rewritten instruction $y'$, sampled source frames from $V_{\mathrm{src}}$,
and reference images in $R^+$. The final hidden states are mapped to the
DiT cross-attention dimension with a zero-initialized linear projector:
\begin{equation}
  \mathbf{T}_{\mathrm{ctx}}
  =
  \operatorname{Linear}\!\left(
    \operatorname{MLLM}(y',V_{\mathrm{src}},R^+)
  \right).
  \label{eq:mllm_context}
\end{equation}
The multimodal context $\mathbf{T}_{\mathrm{ctx}}$ enters DiT
cross-attention, so the instruction, source frames, and selected
references share one multimodal context. This Qwen3.5-4B encoder is
distinct from the VLM agent backbone introduced in
Sec.~\ref{sec:method:conditioning}; only an additional linear projector is
trained, while the encoder weights stay frozen.

\textbf{Latent token sequence and zero-timestep modulation.}
Let $\mathcal{E}$ denote the frozen Wan VAE encoder~\citep{wan},
$\operatorname{Patchify}(\cdot)$ the DiT patch embedding, and
$\mathbf{z}_t$ the noisy video latent at flow-matching timestep $t$
(defined in Eq.~\eqref{eq:flow_matching}). To preserve unedited content
and the identity of reference images, these visual conditions are
concatenated along the token-sequence dimension via the operator
$[\,\cdot\,;\,\cdot\,]$:
\begin{equation}
  \begin{aligned}
    \mathbf{S}_{\mathrm{noisy}} = \operatorname{Patchify}(\mathbf{z}_t),
    \mathbf{S}_{\mathrm{src}} &= \operatorname{Patchify}(\mathcal{E}(V_{\mathrm{src}})),
    \mathbf{S}_{\mathrm{ref}} =\bigl[\,\operatorname{Patchify}(\mathcal{E}(r_k))+e_k\,\bigr]_{r_k\in R^+},\\
    \mathbf{S} &= [\mathbf{S}_{\mathrm{noisy}};\mathbf{S}_{\mathrm{src}};\mathbf{S}_{\mathrm{ref}}],
  \end{aligned}
  \label{eq:latent_sequence}
\end{equation}
where $e_k$ is a learnable reference-index embedding following~\citep{wu2025omnigen2}. If
$R^+=\emptyset$, then $\mathbf{S}_{\mathrm{ref}}$ is empty. Each
reference image is VAE-encoded at its own aspect ratio, and only the
noisy video latent is decoded at output time. Mask conditioning has no
separate branch: a binary mask is first composited as a masked image, then
inserted into $R^+$ like any other reference image.

Noisy tokens, source tokens, and reference tokens play different roles:
noisy tokens move along the flow-matching trajectory, while source and
reference tokens remain fixed across timesteps and provide conditioning.
Wan2.2 DiT blocks receive
timestep embeddings through AdaLN modulation~\citep{wan}. We construct a
real-timestep modulation $a(t)$ and a zero-timestep modulation $a(0)$,
then assign each token a selector $\eta_i$:
\begin{equation}
  a_i =
  (1-\eta_i)a(t)+\eta_i a(0),
  \qquad
  \eta_i =
  \begin{cases}
    0, & i\in \text{noisy tokens},\\
    1, & i\in \text{source or reference tokens}.
  \end{cases}
  \label{eq:zero_timestep}
\end{equation}
Conditioning tokens therefore do not change their modulation with the
denoising noise level, while noisy tokens follow the normal flow-matching
trajectory.

\subsection{Training objectives}
\label{sec:method:training}

\textbf{Video DiT Training.} Given conditioning tuple $\mathbf{x}$, the video DiT predicts a
conditional velocity field $v_\theta(\mathbf{z}_t,t;\mathbf{x})$ over
VAE latents. Let $\mathbf{z}_1=\mathcal{E}(V_{\mathrm{edit}})$ be the latent
of the target video, let $\mathbf{z}_0\sim\mathcal{N}(0,I)$ be Gaussian
noise, and set $\mathbf{z}_t=(1-t)\mathbf{z}_0+t\mathbf{z}_1$. The
video DiT is trained with the standard flow-matching objective
\begin{equation}
  \mathcal{L}_{\mathrm{FM}}(\theta)
  =
  \mathbb{E}_{t,\mathbf{z}_0,\mathbf{z}_1,\mathbf{x}}
  \left[
    \left\lVert
      v_\theta(\mathbf{z}_t,t;\mathbf{x})
      -(\mathbf{z}_1-\mathbf{z}_0)
    \right\rVert_2^2
  \right],
  \label{eq:flow_matching}
\end{equation}
with samples from image editing, instruction-conditioned video editing,
and reference-conditioned video editing mixed under the same objective.
The model therefore sees empty-reference edits, user-reference edits,
and masked-image edits as instances of one conditioning tuple. Prompt
dropout and visual dropout train the unconditional and visual-negative
branches that the inference-time guidance schedule uses
(Appendix~\ref{sec:appendix:inference_cfg}); the training data mixture
is described in Sec.~\ref{sec:data}.

\textbf{VLM Agent Supervised Finetuning (SFT).}
After training the DiT to use unified conditions, we freeze it and train
the VLM agent to construct those conditions. Agent SFT has two
supervised tasks: complete edit planning and reference-image selection.
Let $\mathcal{D}_{\mathrm{plan}}$ contain VLM agent context
$s=(V_{\mathrm{src}},y,R)$ and target plan $p^\star=(y',c,q,m)$. The
planning loss is
\begin{equation}
  \mathcal{L}_{\mathrm{SFT}}(\phi)
  =
  -\mathbb{E}_{(s,p^\star)\sim\mathcal{D}_{\mathrm{plan}}}
  \log \pi_\phi(p^\star\mid s).
  \label{eq:sft_loss}
\end{equation}
The supervision target is not a natural-language rationale. It is the
complete edit plan $p^\star=(y',c,q,m)$ defined in
Eq.~\eqref{eq:agent_plan}; downstream tool execution then turns this
plan into the conditioning tuple in Eq.~\eqref{eq:conditioning_tuple}.
Reference-image selection uses a separate image-choice set and the same
token-level negative log likelihood, but the output is a single candidate
index such as \texttt{Image 3}. In this way, SFT directly supervises the
inference decisions to rewrite text, route the task, generate a search
query, select a reference image, or request a mask.

\textbf{VLM Agent Preference Alignment.}
SFT teaches the VLM agent to produce valid plans, but it does not fully
resolve borderline cases between similar plans. We therefore apply
DPO~\citep{dpo} to plan-level preference pairs.
Given VLM agent context $s$, chosen
plan $p^+$, rejected plan $p^-$, current policy $\pi_\phi$, and frozen
reference policy $\pi_{\mathrm{ref}}$, the objective is
\begin{equation}
  \mathcal{L}_{\mathrm{DPO}}(\phi)
  =
  -\mathbb{E}_{(s,p^+,p^-)}
  \log\sigma\!\left[
    \beta
    \log
    \frac{\pi_\phi(p^+\mid s)\pi_{\mathrm{ref}}(p^-\mid s)}
         {\pi_{\mathrm{ref}}(p^+\mid s)\pi_\phi(p^-\mid s)}
  \right],
  \label{eq:dpo_loss}
\end{equation}
where $\sigma$ is the logistic function and $\beta$ is the DPO
temperature. Preference pairs are built around five plan-level boundary
cases: source-entity false triggers (image search called for an entity
already visible in the source video), ambiguous mask phrases (phrase
too vague to ground cleanly), false image-search
triggers (search called for edits that need no real-world reference,
such as style transfer), constraint-losing rewrites (rewrite drops
user-specified constraints from the original prompt), and task-routing
confusions (a similar but wrong edit category is chosen). Because the
video DiT is frozen throughout VLM agent training, the preference stage
sharpens the VLM agent's decisions without mixing video sampling
variance into the learning objective.

\section{Data Construction}
\label{sec:data}

We construct two interdependent datasets. The first trains the unified video DiT to use source, text, reference, and masked-image
conditions. The second trains the VLM agent to decide which of those
conditions should be supplied at inference time.

\paragraph{Video editing model data.}
\label{sec:data:video_model_corpus}
The video DiT is trained entirely on open-source datasets. Table~\ref{tab:data_video_model_mix} reports the video DiT training
mixture. Its three subsets correspond to the conditioning signals each
sample teaches: image-edit pairs provide diverse instruction-following
supervision, instruction-based video edits provide temporal source
conditioning, and reference-guided video edits provide identity,
and masked-image conditions. 
A subset of the open-source video-edit sources is filtered and relabeled
with Gemini Flash-Lite~\citep{gemini3flashlite}; identity references are
synthesized with FLUX.2-klein~\citep{flux2klein}; and masked images are curated separately.
Additional source-by-source curation and filtering details appear in
Appendix~\ref{sec:appendix:data_details}.

\input{tables/data_video_model_mix}

\paragraph{VLM agent supervised data.}
\label{sec:data:plan_sft}
The VLM agent is trained with three supervision types, with examples
illustrated in Figures~\ref{fig:aurora_sft}, \ref{fig:aurora_select},
and~\ref{fig:aurora_dpo_a}. Planning SFT starts from accepted editing
pairs and degrades clean instructions into shorter or more ambiguous user
requests, with the four-field plan as supervision. Reference-image
selection uses retrieved candidate sets and supervises which single image
should occupy the reference slot. We use 25K planning examples and 10K
reference-image selection examples; detailed construction appears in
Appendix~\ref{sec:appendix:agent_data}.

\paragraph{Preference alignment.}
\label{sec:data:preference}
Preference pairs target boundary cases that the supervised plans alone
do not separate, covering the same five categories defined in
Sec.~\ref{sec:method:training}: source-entity false triggers, ambiguous
mask phrases, false image-search triggers, constraint-losing rewrites,
and task-routing confusions. Each pair keeps the source and most plan
fields fixed while changing one decision axis, so DPO refines the VLM
agent's decision boundaries. The final set contains 1.8K
chosen/rejected pairs. The appendix provides illustrated examples.

\begin{figure}[!t]
  \centering
  \input{figs/agentedit_qualitative}
  \caption{\textbf{Qualitative comparison on \bench{}.} Rows show IP
  (named real-world entity) addition, IP background change, and localized
  removal. Target objects to edit are highlighted in
  (\benchTarget{bold blue}). Reference images come from \method{}'s web
  image search tool and are not provided to the baselines in this
  figure, since this is their default deployment when the user request
  is underspecified. The complementary setup, baselines paired with the
  same \method{} VLM agent, is reported quantitatively in the
  $\checkmark$ rows of Table~\ref{tab:agentedit_bench} and qualitatively
  in Figures~\ref{fig:agent_ablation_univideo}
  and~\ref{fig:agent_ablation_kiwi}.}
  \label{fig:agentedit_qualitative}
\end{figure}
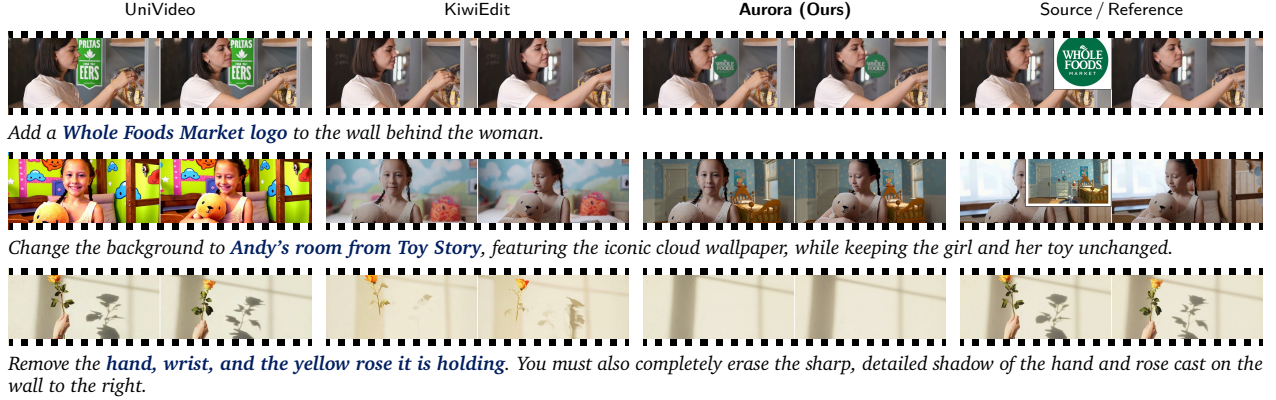

\section{Experiments}
\label{sec:experiments}

We organize the experiments around three questions: whether \method{}
improves editing under underspecified user requests, which part of the
framework is responsible for that gain, and how far the same VLM agent
transfers across video editing models. The first question drives our
main evaluation, in which \bench{} is the primary benchmark for
underspecified requests, while EditVerse-Bench~\citep{editverse} and
OpenVE-Bench~\citep{openve3m} act as sanity checks that the framework
remains competitive when the instruction is already well specified. Ablation studies then address the other two questions,
isolating the source of the gain and measuring its transfer across
video editing models.

\subsection{Experiment setup}

\paragraph{Benchmarks.}
EditVerse-Bench~\citep{editverse} is built from 100 source videos and
200 instruction-video pairs spanning 20 edit categories, with
horizontal and vertical formats evenly balanced. Following the standard
source-video protocol used in current open-source comparisons, we report
results on the 120-case subset, which removes
tasks that require extra controls such as depth, pose, or sketch
inputs. OpenVE-Bench~\citep{openve3m} contains 431 video-edit pairs across
eight categories. Because our current training data does not
cover the creative-edit or camera-edit categories, the table reports
results on the remaining six categories.

\paragraph{Implementation details.} \method{} video DiT uses a two-stage
training recipe: a low-resolution warmup at 399{,}360 pixels
followed by high-resolution finetuning at 921{,}600 pixels, both with
81 frames. Unless noted otherwise, all reported \method{} numbers use the
Stage~2 checkpoint with 50 denoising steps.
Benchmarking scores are judged by Gemini~2.5~Pro~\citep{gemini25pro} at
\texttt{temperature=0}, while video frame handling and
scoring prompts follow the corresponding benchmark-specific evaluation protocol. Appendix~\ref{sec:appendix:exp_details}
provides more details.

\subsection{\bench{}}

\paragraph{Design goal.}
Existing editing benchmarks mostly measure prompt following under
model-ready instructions. \bench{} instead targets requests whose
conditions are incomplete: it contains 150 cases spanning five task
types, each one built around a request that is underspecified along the
textual or the visual axis. It is a VLM-as-judge benchmark:
Gemini~2.5~Pro~\citep{gemini25pro} scores each case with a rubric that
measures instruction following, edit region localization, source
preservation, visual quality, and temporal consistency. The benchmark
maps the two underspecification axes to five task types: IP replacement,
IP addition, IP background change, localized removal, and reasoning edits.
Here, IP refers to a named real-world entity, such as a brand, product,
character, or artwork. For the three IP tasks, the rubric also
scores IP presence and IP identity match. Appendix~\ref{sec:appendix:agentedit_bench} provides the
composition, full scoring rubric, judge prompts, and examples.

\input{tables/agentedit_bench}

\paragraph{Quantitative results.}
Table~\ref{tab:agentedit_bench} compares \method{} against
UniVideo~\citep{univideo} and Kiwi-Edit~\citep{kiwi}, and also measures
each model with and without the \method{} VLM agent on identical source
clips. Even with the agent disabled, the \method{} video DiT alone
reaches $74.7$ Overall and already outperforms UniVideo ($67.0$) and
Kiwi-Edit ($69.7$) under the same raw-prompt input, so the gain on
\bench{} is not driven by the agent stage alone. Adding the VLM agent
on top then raises \method{} from $74.7$ to $87.9$, and the gains are
largest on the IP tasks, especially IP replacement and IP addition,
where the raw request names an entity but does not provide the reference
image the video DiT needs. The VLM agent closes that gap by selecting a
concrete reference image before generation instead of leaving the model
to infer the target appearance from text alone.

\paragraph{Qualitative comparison.}
Figure~\ref{fig:agentedit_qualitative} shows that when the request omits
the reference image needed for a requested real-world entity or visual
pattern, or the spatial grounding needed for a local edit, raw-prompt
baselines often synthesize a generic substitute or edit the wrong
region, whereas \method{} resolves the missing condition before
sampling.
Appendix~\ref{sec:appendix:agentedit_qualitative} extends this
comparison with additional \bench{} cases.

\subsection{Results on existing video editing benchmarks}
\input{tables/standard_bench_overall}
We next evaluate \method{} on existing video editing benchmarks whose
inputs are already sufficiently specified. This section checks whether
the agentic design that helps on \bench{} also preserves strong general
editing quality outside the underspecified setting.
Table~\ref{tab:standard_bench_overall} reports Overall scores judged by
Gemini~2.5~Pro~\citep{gemini25pro} on the 120-case source-video subset
of EditVerse-Bench~\citep{editverse} (0--10 scale) and on
OpenVE-Bench~\citep{openve3m} (1--5 scale), whose task distribution
differs from EditVerse-Bench and includes both scene-level and localized
edits. On both benchmarks, the prompts and any required reference images
are already supplied by the benchmark itself. We disable the VLM agent's web
image search and mask tools throughout this section; the VLM agent only
rewrites the instruction.
\method{} is the strongest open-source method on both benchmarks. The
closed-source Runway Aleph~\citep{runwayaleph} baseline reaches $3.51$
Overall on OpenVE-Bench, $0.13$ above \method{}, so \method{} closes
most but not all of the open-vs.-closed gap on that benchmark. On
OpenVE-Bench, our training data does not include subtitle removal, so
the Subtitle column is reported as zero-shot generalization; even under
this stricter setting, \method{} is the best in that column, and the
3.38 Overall is therefore a conservative estimate. For a like-for-like
comparison against Kiwi-Edit~\citep{kiwi}, which does not natively
support subtitle removal, Tab.~\ref{tab:agent_transfer}(a)(b) reports a
five-category average ($\dagger$) that excludes Subtitle on the same
source clips.
Per-category breakdowns and additional qualitative comparisons appear
in
Tables~\ref{tab:appendix:editverse_bench}--\ref{tab:appendix:openve_bench}
and Appendices~\ref{sec:appendix:qualitative}--\ref{sec:appendix:openve_qualitative}.

\subsection{Ablation study}\input{tables/agent_dpo_ablation}
After establishing \method{}'s performance on both underspecified
requests and existing benchmark instructions, we turn to ablations that
isolate where the gain comes from and how far the VLM agent transfers
across video DiTs.

\paragraph{VLM agent training stages.}
Table~\ref{tab:agent_dpo_ablation} ablates the VLM agent on \method{}'s
own video DiT. Without an agent, the video DiT scores $74.7$ on
\bench{}; supervised planning raises this to $85.0$, and DPO adds a
further $2.9$ points to $87.9$. Supervised planning therefore
carries most of the gain by teaching the VLM agent which conditioning
channels each request needs, while DPO sharpens the
boundary cases defined in Sec.~\ref{sec:method:training} that SFT
leaves ambiguous, such as false image-search triggers, ambiguous
mask phrases, and task-routing confusions.

\input{tables/agent_transfer}

\paragraph{VLM agent transfer across video editing models.}

Table~\ref{tab:agent_transfer} tests whether the \method{} VLM agent
still helps after we keep it fixed and swap only the video editing
model. Each pair compares the same model with and without the VLM
agent's outputs. On OpenVE-Bench, adding the VLM agent improves
Kiwi-Edit~\citep{kiwi} from 3.02 to 3.29 and improves our video DiT from
3.31 to 3.46. The two OpenVE-Bench panels ($\dagger$) report a
five-category average that excludes Subtitle removal, so each model is
evaluated on a category set it natively supports. On EditVerse-Bench,
the same setup improves UniVideo from 6.12 to 6.48 and improves our
video DiT from 7.25 to 7.61. Together with the transfer rows in
Table~\ref{tab:agentedit_bench}, these results show that the VLM agent
remains useful when paired with frozen downstream video editing models.
Both benchmarks already supply prompts and any required reference images
or masks, so the VLM agent's image search and mask tools are disabled
here and its only job is to rewrite the instruction. The gains in
Table~\ref{tab:agent_transfer} therefore show two things directly: the
\method{} VLM agent is effective even as a pure instruction-rewriting
module for unified video editing, and its benefit is not tied to a
single video editing model but carries over to the other unified video
editing models we test. This model-agnostic evaluation is made possible
by the open-source releases of UniVideo~\citep{univideo} and
Kiwi-Edit~\citep{kiwi}, which we thank.
Appendix~\ref{sec:appendix:agent_ablation_baselines} extends this
transfer study with qualitative illustrations in
Figures~\ref{fig:agent_ablation_univideo} and~\ref{fig:agent_ablation_kiwi}.

\section{Limitations}
\label{sec:limitations}

\method{} has two potential limitations. First, we align the VLM agent
only with offline DPO (Sec.~\ref{sec:data:preference}); online RL such
as GRPO~\citep{deepseekmath} over the full editing pipeline, where the
reward would come from the actually edited clip rather than from the
plan, is impractical under our compute budget. Few-step distillation of
the video editing DiT, combined with joint online RL training~\citep{cai2025z} of the
VLM agent and the distilled DiT, is a natural way to close this gap.
Second, the 5B video DiT handles removal, stylization, and
insertion of objects with limited motion reliably, yet adding a
new subject that undergoes large and physically plausible motion
remains difficult. We attribute this to both the
motion-modeling capacity of a 5B backbone and the coverage of
large-motion insertion cases in our editing data. We plan to
address this in future work using a larger backbone and a broader
training mixture.

\paragraph{Broader Impacts.} Like other video editing models, \method{} can be misused for non-consensual identity manipulation, and the image-search tool adds copyright and likeness risks from retrieved references. \method{} should therefore be treated as a creative tool, not an evidence-producing one. 

\section{Conclusion}
\label{sec:conclusion}

This paper presents \method{}, an agentic video editing framework for user requests that leave out information the video model needs. \method{} places a tool-augmented VLM agent in front of a unified video editing model. The VLM agent rewrites the instruction, assigns a task label, and, when needed, retrieves a reference image or grounds a mask before generation. We also introduce \bench{}, a benchmark for textual and visual underspecification that measures the full framework and isolates the contribution of the VLM agent stage on the same source clips. Experiments show that adding the VLM agent raises the overall \bench{} score from 74.7 to 87.9 with the same \method{} video DiT. On EditVerse-Bench and OpenVE-Bench, where the inputs are already sufficiently specified, \method{} remains strong and the VLM agent still helps through instruction rewriting alone. The transfer results further show that this benefit is not tied to a single video editing model and carries over to other unified video editing models. Progress in video editing depends not only on stronger video DiTs, but also on reliably turning raw user requests into model-ready conditions.

\medskip
\bibliographystyle{plainnat}
\bibliography{aurora}

\newpage
\appendix

\begin{figure}[t]
  \centering
  \input{supp/figs/agentedit_qualitative_portrait_b}
  \caption{Extended qualitative comparison on \bench{}. Reference images come from the \method{} VLM agent's web image search tool and are not provided to the baselines, which see only the original underspecified request.}
  \label{fig:agentedit_qualitative_portrait_b}
\end{figure}
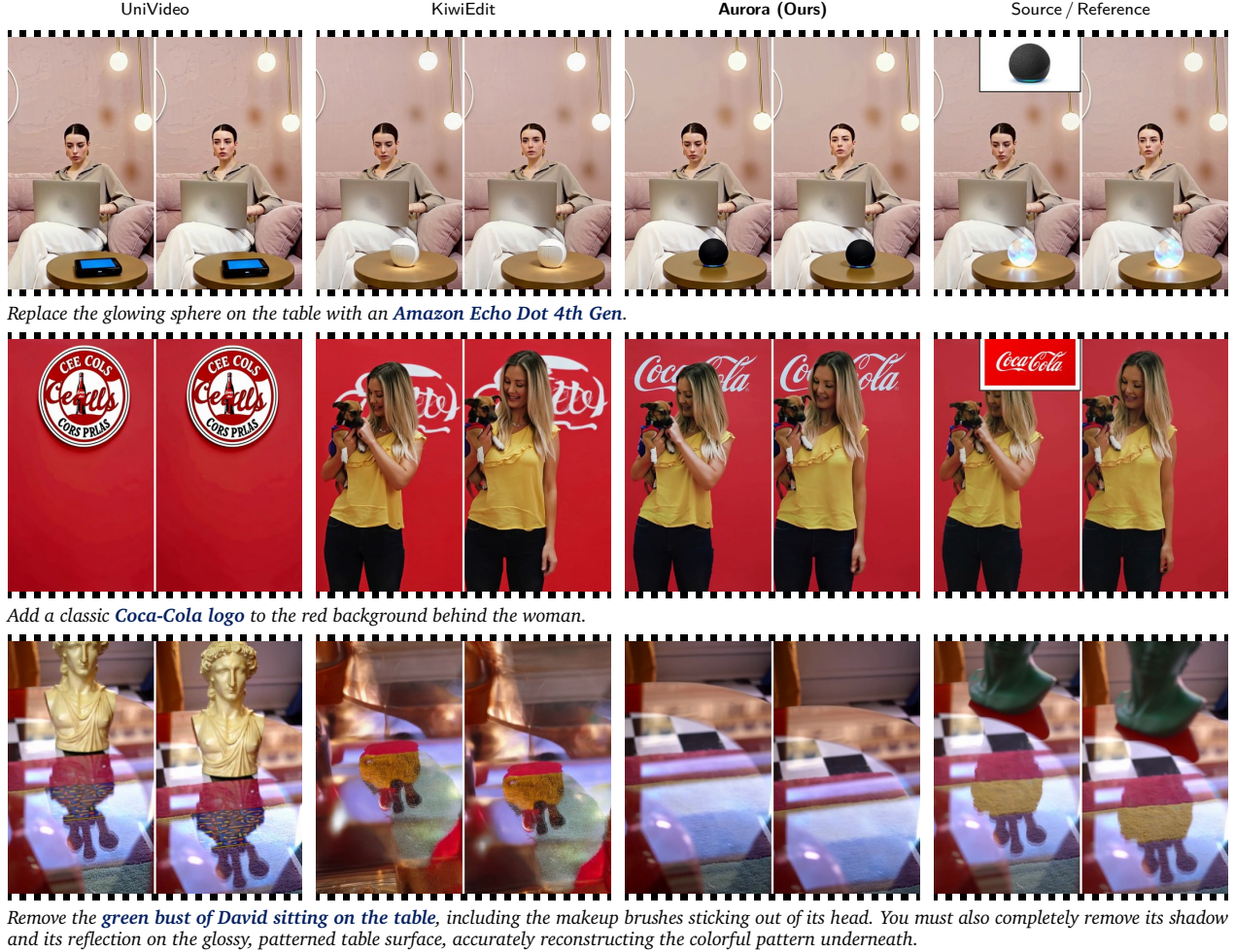

\section{Extended qualitative comparison on \bench{}}
\label{sec:appendix:agentedit_qualitative}

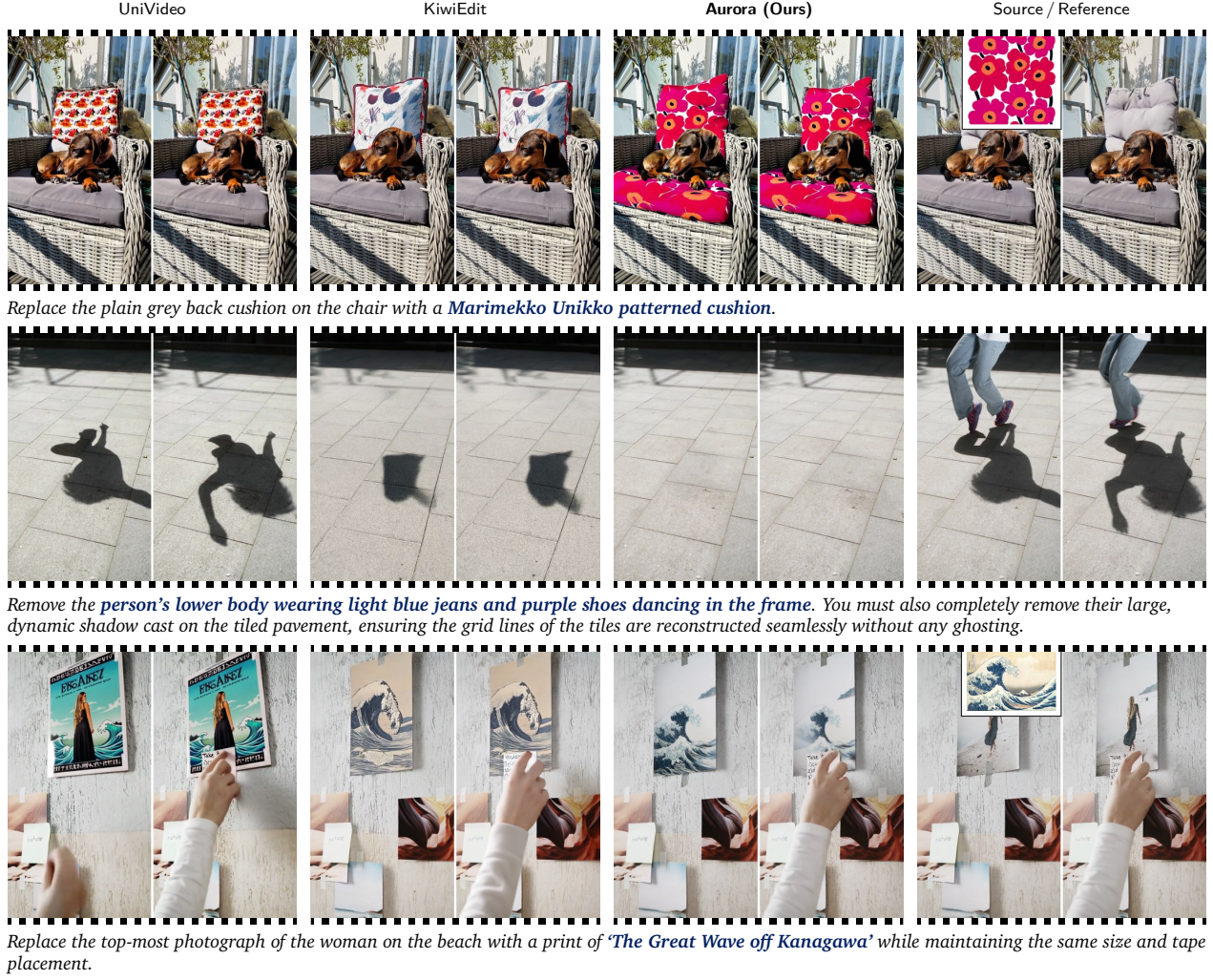
\begin{figure}[!t]
  \centering
  \input{supp/figs/agentedit_qualitative_portrait_a}
  \caption{Extended qualitative comparison on \bench{}.}
  \label{fig:agentedit_qualitative_portrait_a}
\end{figure}

\begin{figure}[!t]
  \centering
  \input{supp/figs/agentedit_qualitative_landscape}
  \caption{Extended qualitative comparison on \bench{}.}
  \label{fig:agentedit_qualitative_landscape}
\end{figure}
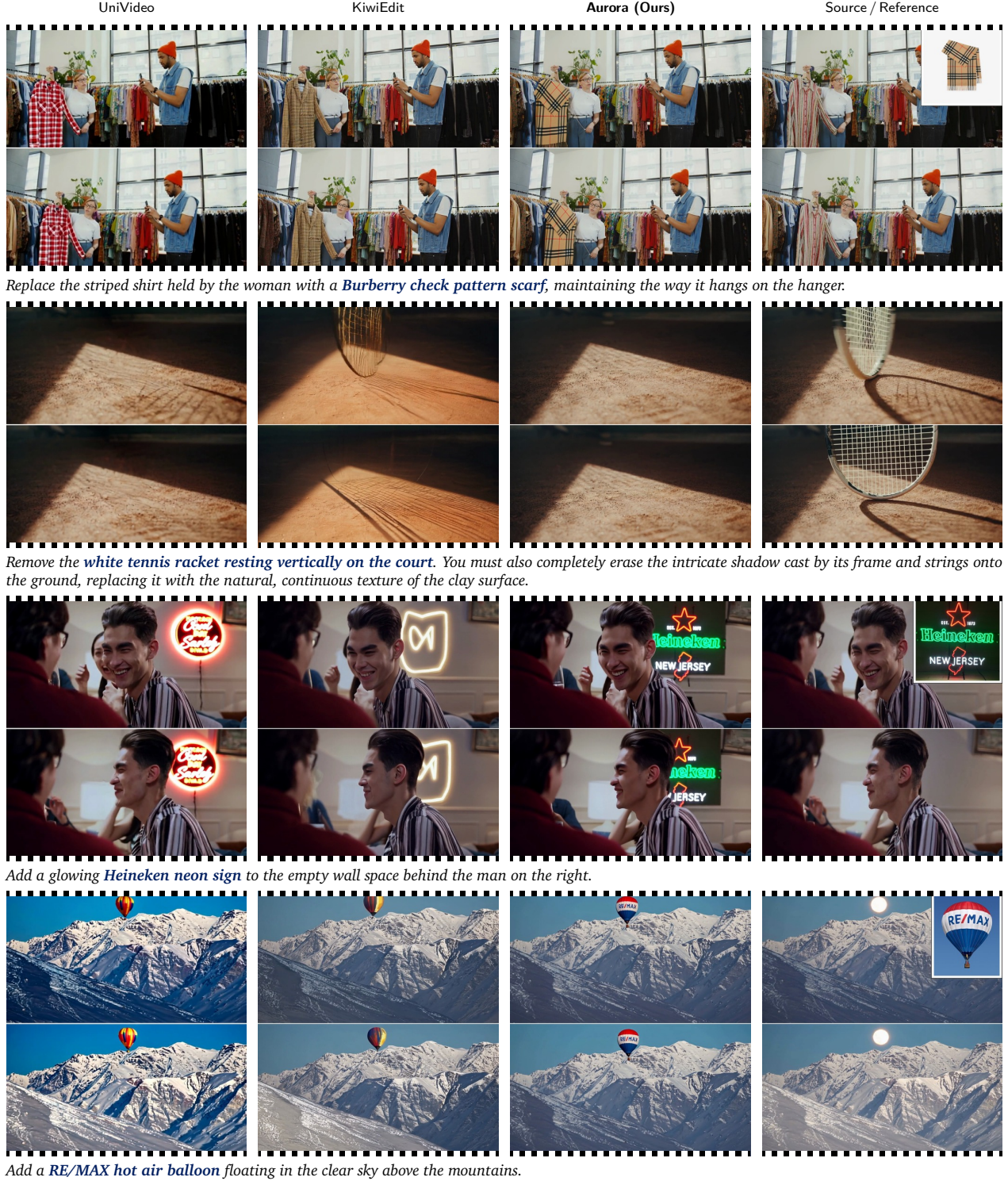

Figures~\ref{fig:agentedit_qualitative_portrait_b},
\ref{fig:agentedit_qualitative_portrait_a}, and
\ref{fig:agentedit_qualitative_landscape} extend
Figure~\ref{fig:agentedit_qualitative} with additional
\bench{} cases against the same two baselines (UniVideo~\citep{univideo} and
Kiwi-Edit~\citep{kiwi}) under their default deployment, in which they use the
raw user prompt without a VLM agent. For IP cases, \method{} uses the
retrieved reference image from our fine-tuned VLM agent, which is shown
in the figure as a small inset overlaid on the first source frame. The
cases are grouped by clip
orientation: Figure~\ref{fig:agentedit_qualitative_landscape}
collects landscape clips covering replacement, removal, and
addition; Figures~\ref{fig:agentedit_qualitative_portrait_b} and
\ref{fig:agentedit_qualitative_portrait_a} collect portrait clips
covering the same five edit types. Across these cases, the two baselines either drop the named entity, substitute a generic one, or fail to remove the target cleanly. In contrast, \method{} more faithfully fulfills the raw user request via the VLM agent's rewritten instruction and, when needed, tool calls that retrieve a reference image. The gains are clearest on edits that require preserving the identity of a real-world entity.

\section{Qualitative comparison on EditVerse-Bench}
\label{sec:appendix:qualitative}

\input{supp/tables/editverse_bench}

\begin{figure}[!t]
  \centering
  \input{supp/figs/qualitative_appendix}
  \caption{Qualitative comparison on
  EditVerse-Bench~\citep{editverse}.}
  \label{fig:qualitative_appendix}
\end{figure}
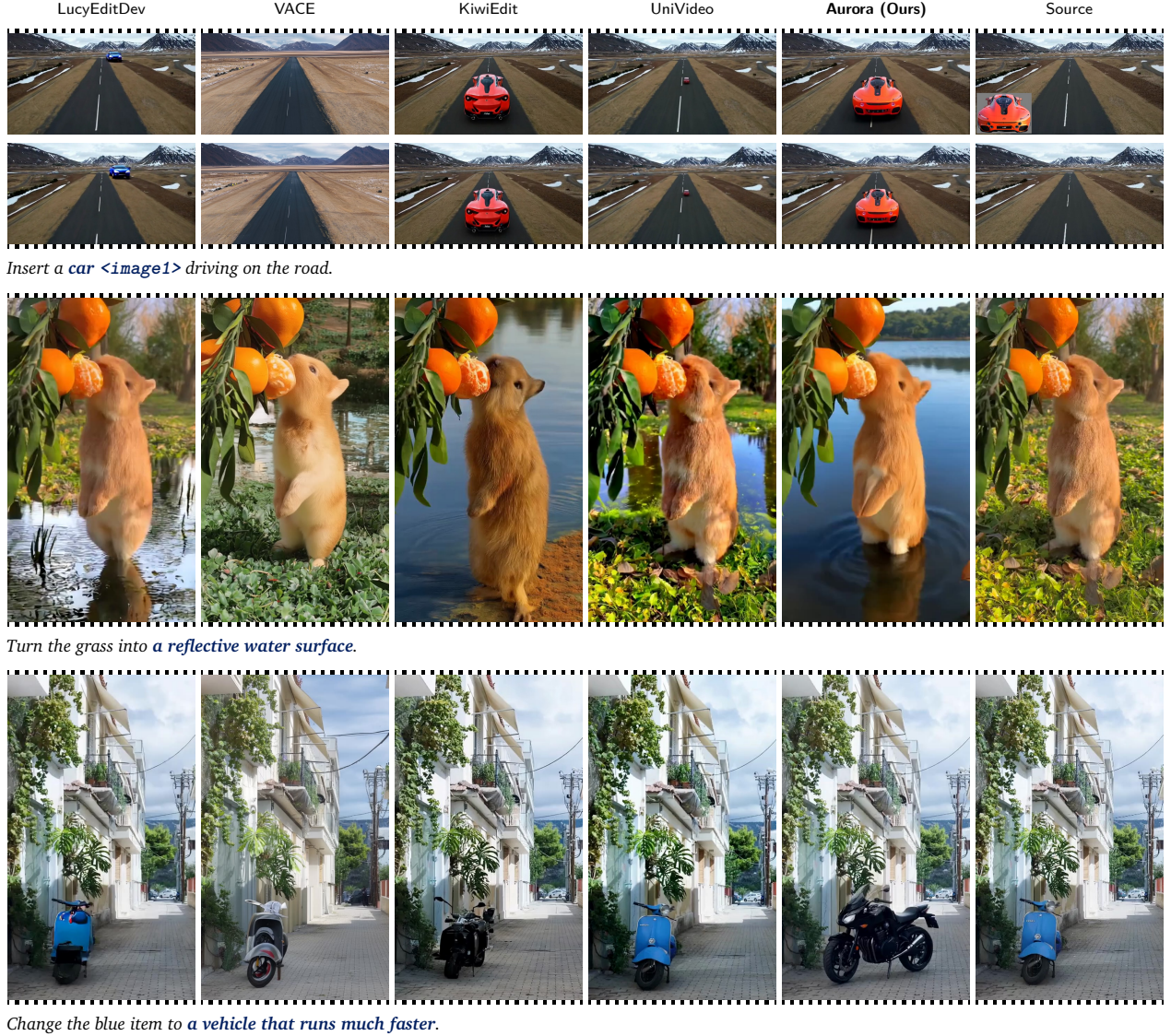

Table~\ref{tab:appendix:editverse_bench} expands the EditVerse-Bench
Overall column from Table~\ref{tab:standard_bench_overall} into the full
per-category breakdown.
Figure~\ref{fig:qualitative_appendix} compares \method{} against four
open-source baselines (LucyEditDev~\citep{lucyedit}, VACE~\citep{vace}, Kiwi-Edit~\citep{kiwi}, UniVideo~\citep{univideo}) on
three EditVerse-Bench~\citep{editverse} cases that together cover reference-conditioned
insertion, background change (i.e., Bkg.), and reasoning. The first case is a
landscape clip shown as the two distant frames sampled at 20\%
and 80\% of the clip to check temporal consistency.

\section{Qualitative comparison on OpenVE-Bench}
\label{sec:appendix:openve_qualitative}

\input{supp/tables/openve_bench}

\begin{figure}[!t]
  \centering
  \input{supp/figs/openve_qualitative}
  \caption{Qualitative comparison on
  OpenVE-Bench~\citep{openve3m}.}
  \label{fig:openve_qualitative}
\end{figure}
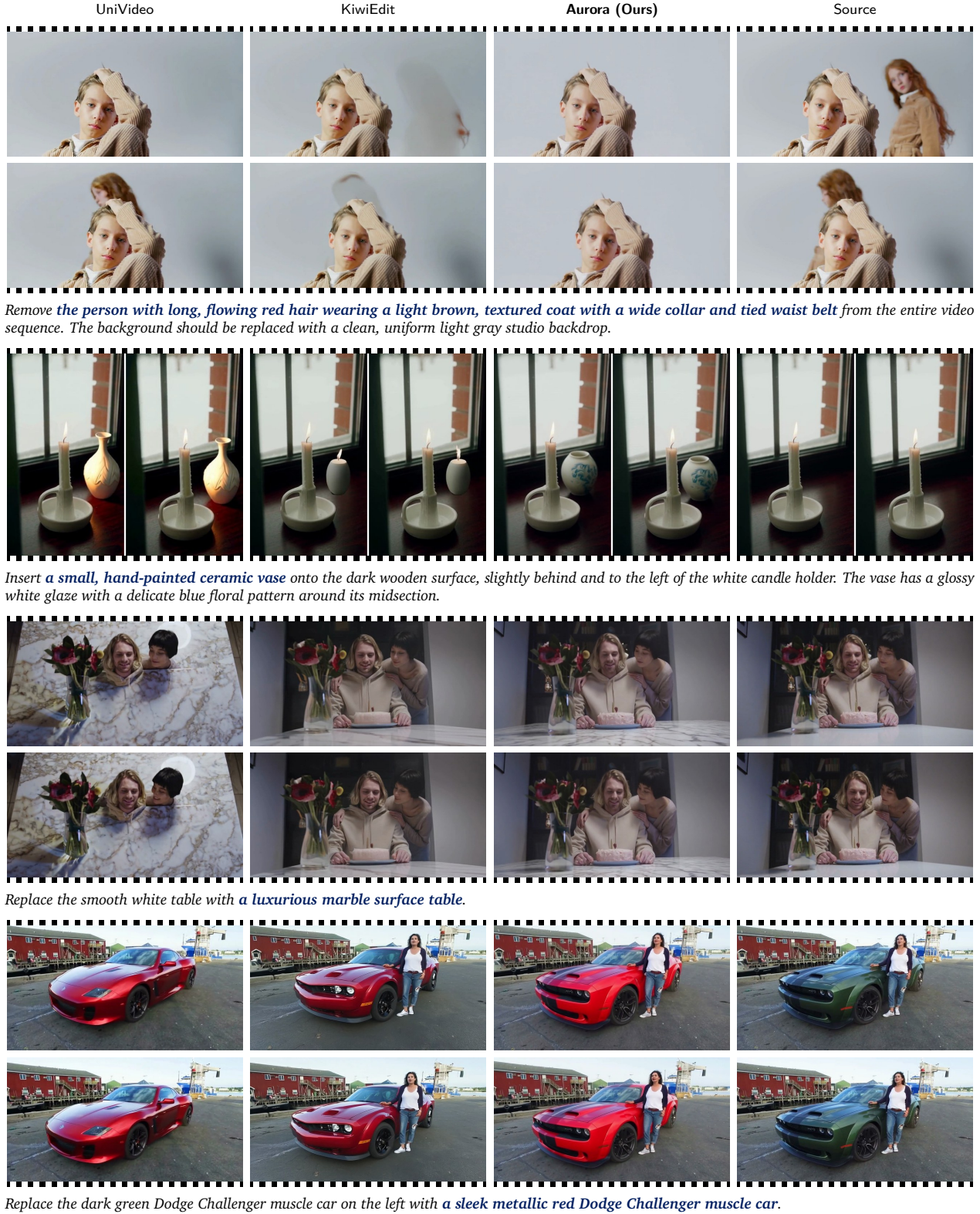

Table~\ref{tab:appendix:openve_bench} expands the OpenVE-Bench Overall
column from Table~\ref{tab:standard_bench_overall} into the full
per-category breakdown.
Figure~\ref{fig:openve_qualitative} compares \method{} against UniVideo
and Kiwi-Edit on four OpenVE-Bench~\citep{openve3m} cases covering local removal,
local addition, and local change. These cases are shown as
the two distant frames (sampled at 25\% and 75\% of the clip) to check temporal consistency.

\section{Qualitative VLM agent transfer across video DiTs}
\label{sec:appendix:agent_ablation_baselines}

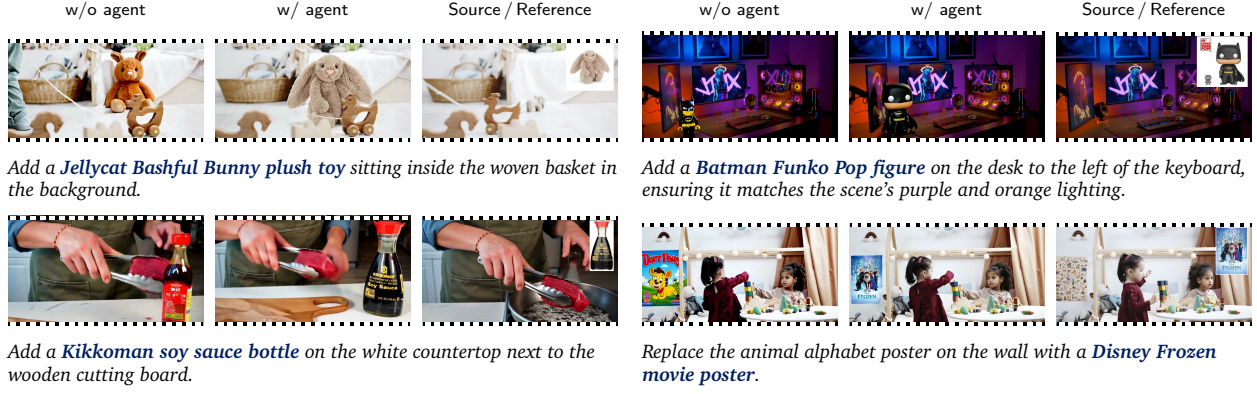
\begin{figure}[!t]
  \centering
  \input{supp/figs/agent_ablation_univideo}
  \caption{Qualitative VLM agent transfer on
  UniVideo~\citep{univideo}.}
  \label{fig:agent_ablation_univideo}
\end{figure}

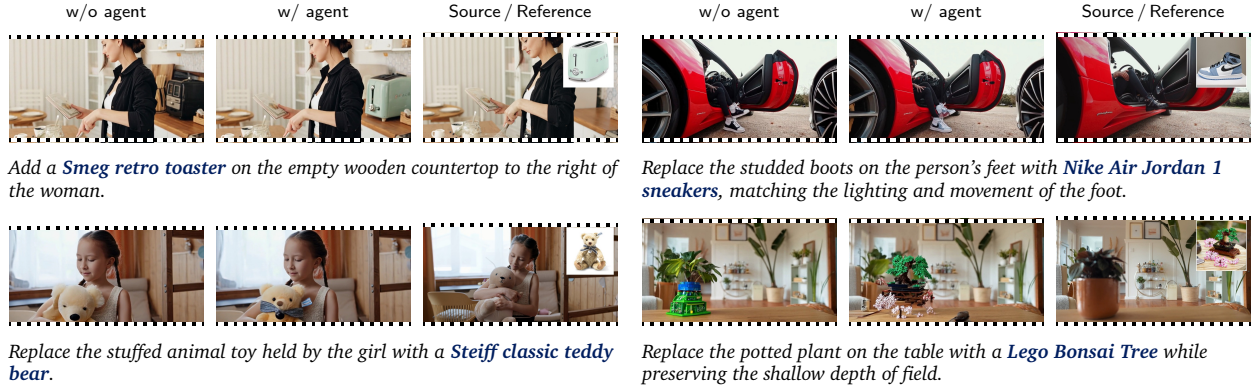
\begin{figure}[!t]
  \centering
  \input{supp/figs/agent_ablation_kiwi}
  \caption{Qualitative VLM agent transfer on
  Kiwi-Edit~\citep{kiwi}.}
  \label{fig:agent_ablation_kiwi}
\end{figure}

Figures~\ref{fig:agent_ablation_univideo}
and~\ref{fig:agent_ablation_kiwi} isolate the VLM agent's contribution
by holding the receiving video DiT fixed and toggling the VLM
agent stage on and off. UniVideo~\citep{univideo} and
Kiwi-Edit~\citep{kiwi} are run twice on the same source clip and the
same user request: once on the raw user prompt (their default
deployment, labeled \emph{w/o agent}), and once on the VLM agent's
rewritten prompt together with the VLM agent's retrieved reference
image (labeled \emph{w/ agent}). In each case, the third cell shows the
first source frame with the retrieved reference image overlaid as a
small inset. Across these examples, the \emph{w/o agent} runs usually
drop the named entity or replace it with a generic substitute, whereas
the \emph{w/ agent} runs follow the rewritten instruction and use the
retrieved reference image to better preserve the identity of the
requested real-world entity.

\section{Experimental details}
\label{sec:appendix:exp_details}

Tables~\ref{tab:appendix:video_model_and_conditioning}--\ref{tab:appendix:video_training_details}
summarize the \method{} video model and training recipe used for the
reported results. Table~\ref{tab:appendix:inference_settings}
summarizes the per-benchmark inference settings and the VLM agent
tool toggles for each evaluation.

\textbf{Output frames per benchmark.} The video DiT always denoises 81
frames internally, matching Wan's $4k{+}1$ temporal constraint, but the
number of frames written to disk differs across benchmarks so that each
saved clip matches the protocol of its evaluator. On \bench{}, the
benchmark consumes the full 81-frame output, so we save all 81 frames.
On EditVerse-Bench, the official evaluator expects 64-frame clips, so we
keep the first 64 frames of the 81-frame output and discard the rest.
On OpenVE-Bench, source clips have variable frame counts, so we
temporal-resize each source to 81 frames before denoising and then
temporal-resize the 81-frame output back to the source's original frame
count when saving, so that the saved edited clip has the same length as
the input.

\input{supp/tables/video_model_and_conditioning}
\input{supp/tables/video_training_details}
\input{supp/tables/inference_settings}

\textbf{Inference-time guidance.}\label{sec:appendix:inference_cfg} At inference time, we evaluate three predictions from the same velocity
field $v_\theta$ trained in Sec.~\ref{sec:method:training}: $v_+$ with
full text and visual conditions, $v_{\mathrm{vneg}}$ with empty or
negative text and visual conditions, and $v_\emptyset$ with no text and
no visual conditions. The guided velocity is
\begin{equation}
  \hat{v}
  =
  v_\emptyset
  +\lambda_{\mathrm{img}}
    (v_{\mathrm{vneg}}-v_\emptyset)
  +\lambda_{\mathrm{txt}}
    (v_+-v_{\mathrm{vneg}}).
  \label{eq:three_pass_cfg}
\end{equation}
When $\lambda_{\mathrm{img}}=1$, Eq.~\eqref{eq:three_pass_cfg}
algebraically reduces to the standard two-pass text CFG path. The
unconditional and visual-negative branches that this formula evaluates
are exactly the two dropout branches trained by the prompt dropout and
visual dropout in Sec.~\ref{sec:method:training}.

\section{Additional VLM agent data details}
\label{sec:appendix:agent_data}

\subsection{Data-construction details}
\label{sec:appendix:data_details}

\paragraph{Video editing model data.}
The mixture of training data in Table~\ref{tab:data_video_model_mix} is built
so that instructions, source videos, and reference images all
appear under the same conditioning format. Large open-source video-edit
sources such as ReCo~\citep{reco}, Ditto~\citep{ditto},
EgoEdit~\citep{egoedit}, and OpenVE~\citep{openve3m} are judged by
Gemini Flash-Lite~\citep{gemini3flashlite} on source quality, target
quality, edit localization, identity preservation, motion consistency,
edit authenticity, and prompt alignment. Accepted samples keep the
original instruction when it is faithful; otherwise the judge-provided
rewrite is used.
The ``Ditto combined-task data'' row of Table~\ref{tab:data_video_model_mix}
is mined from same-source pairs in Ditto~\citep{ditto}. For each source
video $S$ with two edited variants $T_i$ and $T_j$ produced by
instructions $A$ and $B$, the model is trained on the pair whose input is
$T_i$ and whose target is $T_j$, so that $S$ is never seen and the new
instruction must describe the change from $T_i$ to $T_j$ using only what
is currently visible. Three rule-based filters first discard candidates
in which $A$ and $B$ are both style transfers, both global scene
changes, or both manipulate the same object, since those collapse into a
single edit. Gemini Flash-Lite~\citep{gemini3flashlite} then inspects the
three frames and accepts a candidate only when the two edits are
visually distinct and the change from input to target can be written as
one short instruction describable from the input alone. The final subset contains 64{,}782
pairs.

\paragraph{Planning SFT and reference-image selection.}
Planning SFT starts from accepted editing data with a clean
instruction. We do not ask the teacher to write an explanation; instead,
we ask what a real user would most likely omit and how the missing
information should be restored into the four-field format. The clean
instruction is degraded into a shorter, more ambiguous, more colloquial,
or identity-hiding raw request, while the supervision target remains the
plan defined in Eq.~\eqref{eq:agent_plan}. Sampling is organized
by task control rather than raw frequency. Reference-image selection uses another
10K supervised examples adapted from~\citep{gensearcher}. Each example
contains a retrieved entity, web context, and several candidate images;
the target is a discrete candidate index such as \texttt{Image 3}, so
the VLM agent learns which image should populate the reference slot after
retrieval.

Figure~\ref{fig:aurora_sft} shows the additional planning SFT examples. Each cell pairs two source video
frames with the colloquial user request and the four-field edit plan
that the VLM agent is supervised to produce. The fifteen cells together
illustrate when each plan field carries information versus when it
remains \texttt{false}: an identity-bearing target or a named
real-world entity triggers \texttt{image\_search}; a localized removal
triggers \texttt{mask}; a global style change and a combined edit leave
both tool fields empty.

\input{supp/figs/aurora_sft}

Figure~\ref{fig:aurora_select} shows the additional reference-image selection SFT
subset summarized in Section~\ref{sec:data:plan_sft}. Once the planner has
emitted a non-empty \texttt{image\_search} query, web image search returns
multiple candidates. The VLM agent's task here is simple: pick the single
image that is visually grounded and factually consistent with the
named-entity constraints in the search prompt.

\input{supp/figs/aurora_select}

\paragraph{Preference alignment.}
Preference data targets cases that SFT does not isolate well. Each
chosen/rejected pair keeps the source, raw request, and most plan fields
fixed while varying one decision axis. The 1.8K-pair set covers the
same five plan-level boundary cases defined in
Sec.~\ref{sec:method:training}: source-entity false triggers, ambiguous
mask phrases, false image-search triggers, constraint-losing rewrites,
and task-routing confusions.

\definecolor{dpoGoodEdge}{HTML}{2E7D32}
\definecolor{dpoGoodHi}{HTML}{D7ECC8}
\definecolor{dpoBadEdge}{HTML}{C62828}
\definecolor{dpoBadHi}{HTML}{F4C9C2}

Figures~\ref{fig:aurora_dpo_a} and~\ref{fig:aurora_dpo_b} show the
preference alignment data. Each panel keeps the user prompt and
shared plan fields fixed and varies one preference axis at a time. The
chosen and rejected plan JSONs are shown in full so the reader can see
exactly which field flips between the two; that field is marked with a
\textcolor{dpoGoodEdge}{\rule[-0.05em]{0.30em}{0.85em}} green bar and
bold green text in the chosen panel and a
\textcolor{dpoBadEdge}{\rule[-0.05em]{0.30em}{0.85em}} red bar and bold
red text in the rejected panel.

\input{supp/figs/aurora_dpo}

\section{\bench{}}
\label{sec:appendix:agentedit_bench}

\begin{figure}[!t]
  \centering
  \begin{subfigure}[c]{0.42\linewidth}
    \centering
    \includegraphics[height=5.0cm,keepaspectratio]{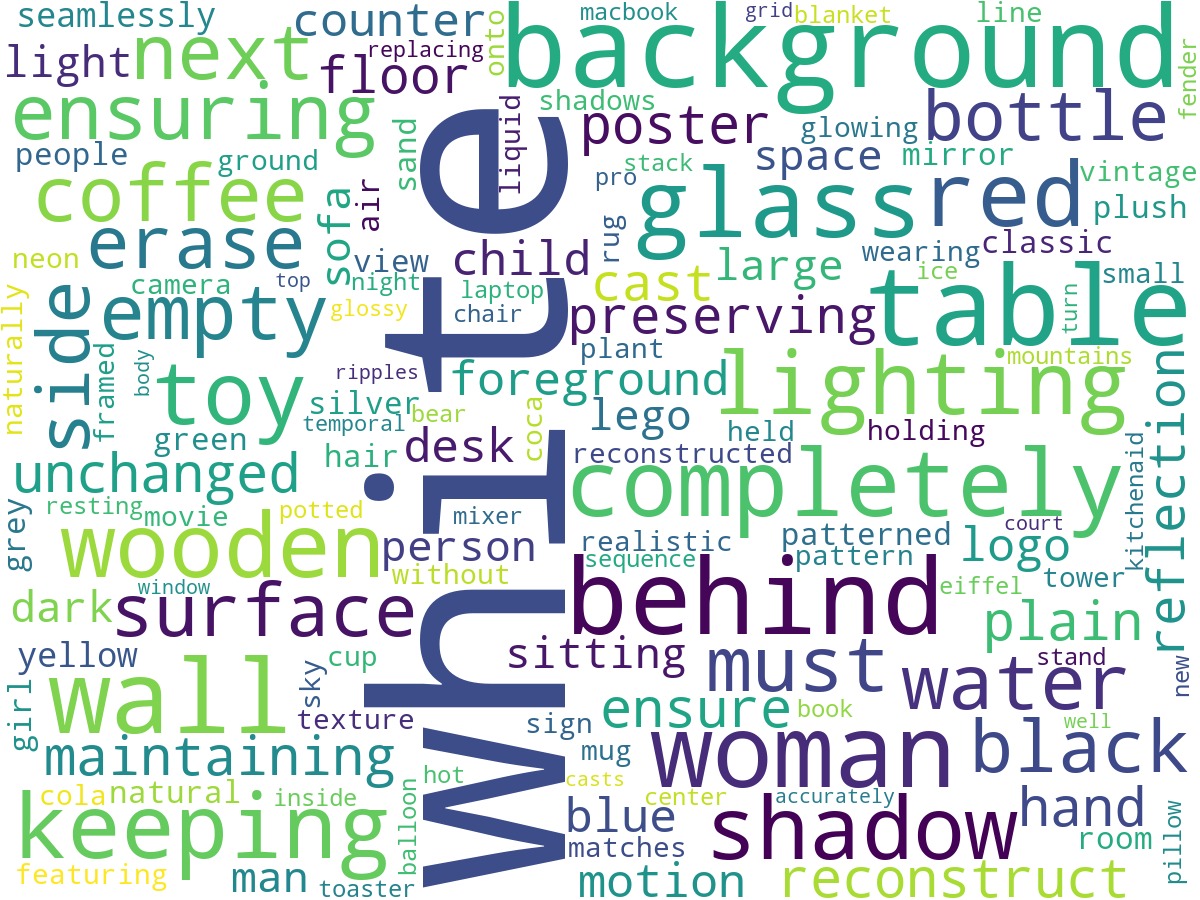}
    \subcaption{Instruction word cloud, generic edit verbs and stop words removed.}
    \label{fig:appendix:agentedit_bench:wordcloud}
  \end{subfigure}\hfill
  \begin{subfigure}[c]{0.55\linewidth}
    \centering
    \includegraphics[height=5.0cm,keepaspectratio]{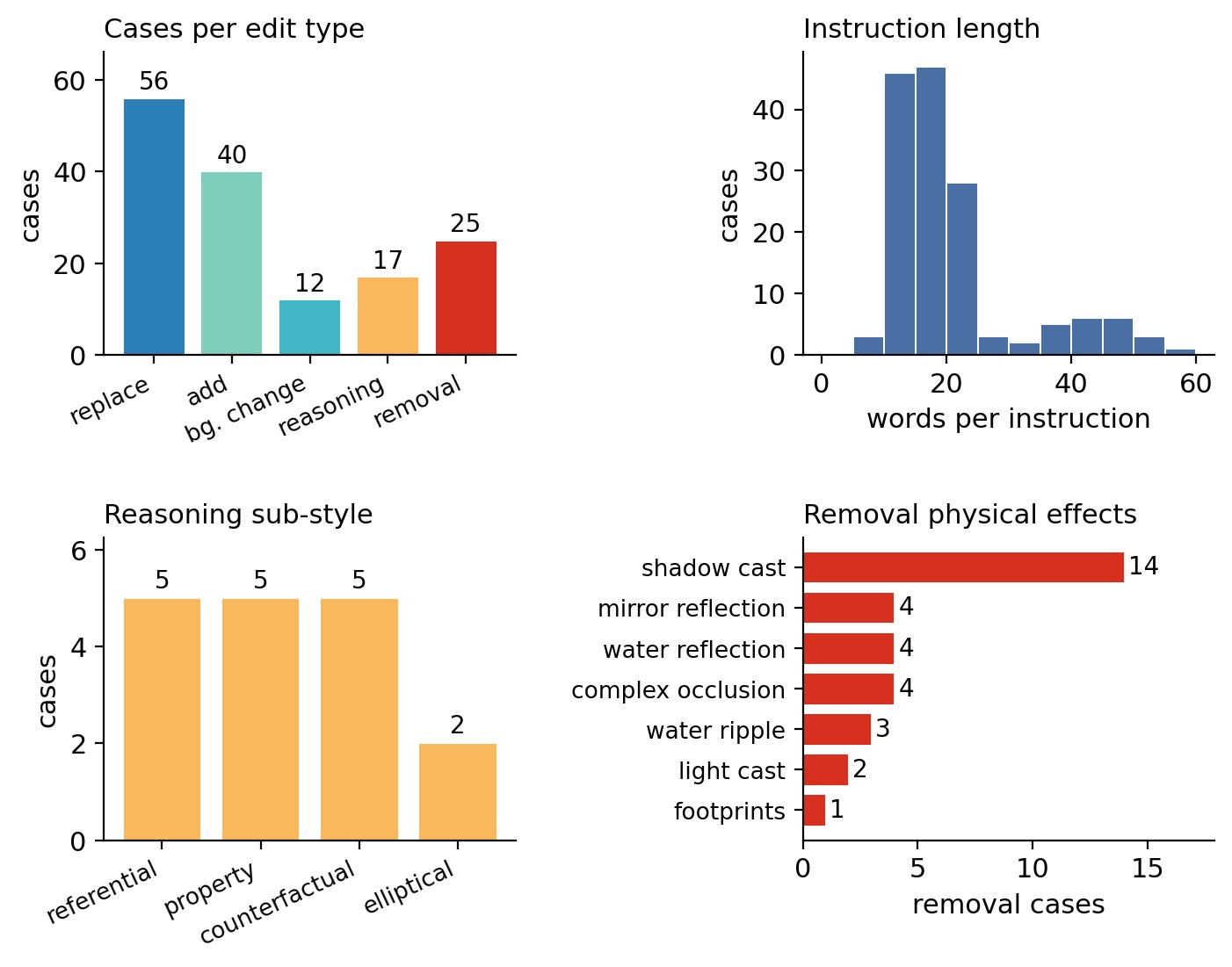}
    \subcaption{Cases per edit type, instruction length, reasoning sub-style, and removal physical effects.}
    \label{fig:appendix:agentedit_bench:stats}
  \end{subfigure}
  \caption{\bench{} distribution. The vocabulary in
  (\subref{fig:appendix:agentedit_bench:wordcloud}) is dominated by
  scene and entity nouns, and the four count plots in
  (\subref{fig:appendix:agentedit_bench:stats}) show the case mix and
  the substructure inside the reasoning and removal subsets.}
  \label{fig:appendix:agentedit_bench:distribution}
\end{figure}

\bench{} is a 150-case video editing benchmark over five edit types. All source videos of benchmark are drawn from the Pexels
stock library.
Figure~\ref{fig:appendix:agentedit_bench:examples} shows nine cases
covering all five edit types, with the named target or counterfactual
cue and the edit region or physical effect highlighted in the
instruction. Figure~\ref{fig:appendix:agentedit_bench:distribution}
reports the instruction word cloud and four count distributions: cases
per edit type, instruction length, reasoning sub-style, and the
physical effects that removal targets carry.

\begin{figure}[!t]
  \centering
  \input{supp/figs/agentedit_bench_examples}
  \caption{\bench{} examples.
  Each case is one source clip plus one user request; we
  highlight the named target or counterfactual cue
  (\benchTarget{bold blue}) and the edit region or physical effect
  (\benchRegion{italic green}).}
  \label{fig:appendix:agentedit_bench:examples}
\end{figure}

\subsection{Scoring rubric and judge prompt}
\label{sec:appendix:agentedit_bench:rubric}

Each case is scored on the seven axes in
Table~\ref{tab:appendix:agentedit_bench:rubric}; IP edits use all
seven, and \texttt{reasoning} and \texttt{removal} use only axes 1--5.
The removal subset adds an anti-replacement clause to axis~1: an
output that substitutes a new object for the target is capped at $1$,
which prevents hallucinated replacements from scoring as clean
inpainting. Separating \emph{IP Presence} from \emph{IP Identity
Match} distinguishes an edit that places \emph{something} in the right
region from one that misses the region entirely. Per-edit-type numbers
use the native axis count; the cross-type aggregate normalizes every
case to a $0$--$21$ scale and we report it as a percentage.

\begin{table}[t]
  \centering
  \caption{\bench{} scoring rubric. IP edits use all seven axes
  (max 21); reasoning and removal use axes 1--5 only (max 15).}
  \label{tab:appendix:agentedit_bench:rubric}
  \input{supp/tables/agentedit_bench_rubric}
\end{table}

The judge runs with \texttt{temperature}=0, and
scores each case on three frames sampled at indices $0$, $n/3$, and
$2n/3$. Per-axis scores are averaged across the three frames and the
raw judge text is persisted next to the parsed scores. 
Figure~\ref{fig:appendix:agentedit_bench:prompts} shows the verbatim
prompts.

\clearpage
\begin{center}
\input{supp/figs/agentedit_bench_judge_prompts}
\end{center}
\vspace{-2pt}
\captionof{figure}{VLM-as-judge prompts. The two variants
share the body and differ in axis count and the removal-only
anti-replacement clause.}
\label{fig:appendix:agentedit_bench:prompts}
\vspace{4pt}

\end{document}

%% file: math_commands.tex
\usepackage{amsmath,amsfonts,amssymb,bm,dsfont}

\def\eqref#1{(\ref{#1})}

\DeclareMathAlphabet{\mathsfit}{\encodingdefault}{\sfdefault}{m}{sl}
\SetMathAlphabet{\mathsfit}{bold}{\encodingdefault}{\sfdefault}{bx}{n}

\def\1{{\mathds{1}}}

%% file: figs/filmstrip.tex
\newif\iffilmstripdraft
\filmstripdraftfalse

\providecommand{\fsdraw}[4]{%
  \iffilmstripdraft
    \includegraphics[width=#1]{#2}%
  \else
  \begin{tikzpicture}[inner sep=0pt, outer sep=0pt]
    \node[inner sep=0pt] (fsImg) {\includegraphics[width=#1]{#2}};
    \pgfmathsetlengthmacro{\fsW}{#1}%
    \pgfmathsetlengthmacro{\fsBar}{0.020*\fsW}%
    \pgfmathsetlengthmacro{\fsHoleW}{0.026*\fsW}%
    \pgfmathsetlengthmacro{\fsHoleH}{0.020*\fsW}%
    \pgfmathsetlengthmacro{\fsPitch}{0.050*\fsW}%
    \pgfmathsetlengthmacro{\fsTopY}{-0.5*\fsBar-0.5*\fsHoleH}%
    \pgfmathsetlengthmacro{\fsTopYY}{-0.5*\fsBar+0.5*\fsHoleH}%
    \pgfmathsetlengthmacro{\fsBotY}{0.5*\fsBar-0.5*\fsHoleH}%
    \pgfmathsetlengthmacro{\fsBotYY}{0.5*\fsBar+0.5*\fsHoleH}%
    \pgfmathsetlengthmacro{\fsHoleOff}{0.5*(\fsPitch-\fsHoleW)}%
    \pgfmathtruncatemacro{\fsN}{\fsW/\fsPitch}%
    \if#3T%
      \fill[black]
        (fsImg.north west) rectangle ([yshift=-\fsBar]fsImg.north east);
    \fi
    \if#4T%
      \fill[black]
        ([yshift=\fsBar]fsImg.south west) rectangle (fsImg.south east);
    \fi
    \foreach \fsI in {0,...,\fsN}{%
      \pgfmathsetlengthmacro{\fsX}{\fsI*\fsPitch+\fsHoleOff}%
      \pgfmathsetlengthmacro{\fsXR}{\fsX+\fsHoleW}%
      \if#3T%
        \fill[white]
          ([xshift=\fsX, yshift=\fsTopY]fsImg.north west)
          rectangle ([xshift=\fsXR, yshift=\fsTopYY]fsImg.north west);
      \fi
      \if#4T%
        \fill[white]
          ([xshift=\fsX, yshift=\fsBotY]fsImg.south west)
          rectangle ([xshift=\fsXR, yshift=\fsBotYY]fsImg.south west);
      \fi
    }
  \end{tikzpicture}%
  \fi
}

\providecommand{\filmstrip}[2][\linewidth]{\fsdraw{#1}{#2}{T}{T}}
\providecommand{\filmstripT}[2][\linewidth]{\fsdraw{#1}{#2}{T}{F}}
\providecommand{\filmstripM}[2][\linewidth]{\fsdraw{#1}{#2}{F}{F}}
\providecommand{\filmstripB}[2][\linewidth]{\fsdraw{#1}{#2}{F}{T}}

%% file: tables/data_video_model_mix.tex
\begin{table}[H]
  \centering
  \footnotesize
  \setlength{\tabcolsep}{10pt}
  \renewcommand{\arraystretch}{1.08}
  \caption{\textbf{Video DiT training data.} Counts report accepted
  training pairs after curation. \textdagger{} marks filtered sources,
  \textsuperscript{cap} marks re-captioned sources, and
  \textsuperscript{syn} marks reference synthesis, captioning, or
  masked-image generation.}
  \label{tab:data_video_model_mix}
  \begin{tabular}{L{0.24\linewidth}L{0.57\linewidth}r}
    \toprule
    \rowcolor{benchHeader}
    \benchgroup{Training subset} & \benchgroup{Sources} & \benchgroup{Count} \\
    \midrule
    Image editing
      & CrispEdit-2M~\citep{editmgt}, UltraEdit~\citep{ultraedit}, TextEdit~\citep{internvlu}
      & 2,388,440 \\
    Instruction video editing
      & ReCo~\citep{reco}\textdagger{}, Ditto~\citep{ditto}\textdagger{}, OpenVE~\citep{openve3m}\textdagger{}, EgoEdit~\citep{egoedit}\textdagger{}, ROSE~\citep{rose}\textsuperscript{cap}, EffectErase~\citep{effecterase}\textsuperscript{cap}, Ditto combined-task data~\citep{ditto}\textdagger{}
      & 1,672,008 \\
    Reference-guided video editing
      & OpenS2V~\citep{opens2v}\textdagger{}\textsuperscript{syn}, RefVIE~\citep{kiwi}, SpatialVID~\citep{spatialvid}\textdagger{}, ROSE~\citep{rose}\textsuperscript{syn}, EffectErase~\citep{effecterase}\textsuperscript{syn}, HuMoSet~\citep{humo}\textdagger{}
      & 609,988 \\
    \bottomrule
  \end{tabular}
\end{table}

%% file: figs/agentedit_qualitative.tex
\begingroup
\setlength{\tabcolsep}{0pt}
\renewcommand{\arraystretch}{1.0}

\providecommand{\AECStrip}[1]{\filmstrip[\linewidth]{#1}}
\providecommand{\AECHead}[1]{{\fontsize{6.6}{7.6}\selectfont\sffamily\centering #1\par}}
\providecommand{\AECHeadOurs}[1]{{\fontsize{6.6}{7.6}\selectfont\sffamily\bfseries\centering #1\par}}
\providecommand{\AECPrompt}[1]{{\fontsize{7.4}{8.6}\selectfont\itshape\raggedright #1\par}}

\providecommand{\AECHeaderRow}{%
  \noindent\begin{tabular}{@{}p{0.241\textwidth}@{\hspace{0.012\textwidth}}p{0.241\textwidth}@{\hspace{0.012\textwidth}}p{0.241\textwidth}@{\hspace{0.012\textwidth}}p{0.241\textwidth}@{}}
    \AECHead{UniVideo} & \AECHead{KiwiEdit} & \AECHeadOurs{\method{} (Ours)} & \AECHead{Source\,/\,Reference} \\
  \end{tabular}\par
  \vspace{0.30ex}
}

\providecommand{\AECRow}[5]{%
  \noindent\begin{tabular}{@{}p{0.241\textwidth}@{\hspace{0.012\textwidth}}p{0.241\textwidth}@{\hspace{0.012\textwidth}}p{0.241\textwidth}@{\hspace{0.012\textwidth}}p{0.241\textwidth}@{}}
    \AECStrip{#1} & \AECStrip{#2} & \AECStrip{#3} & \AECStrip{#4} \\
  \end{tabular}\par
  \vspace{-0.2ex}\AECPrompt{#5}\vspace{0.55ex}
}

\AECHeaderRow

\AECRow
  {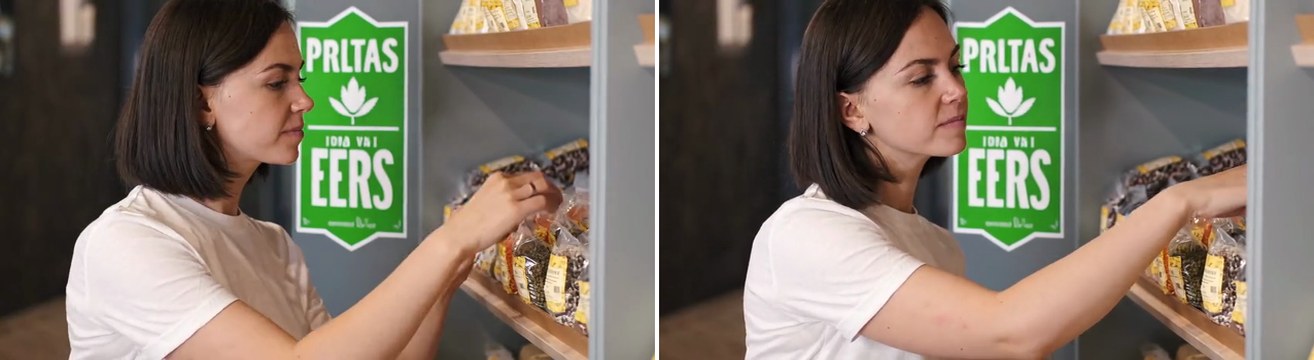}
  {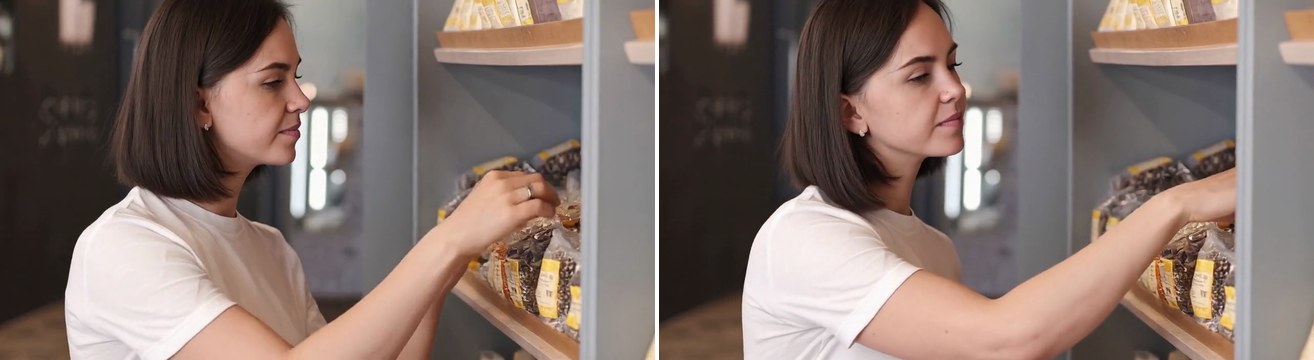}
  {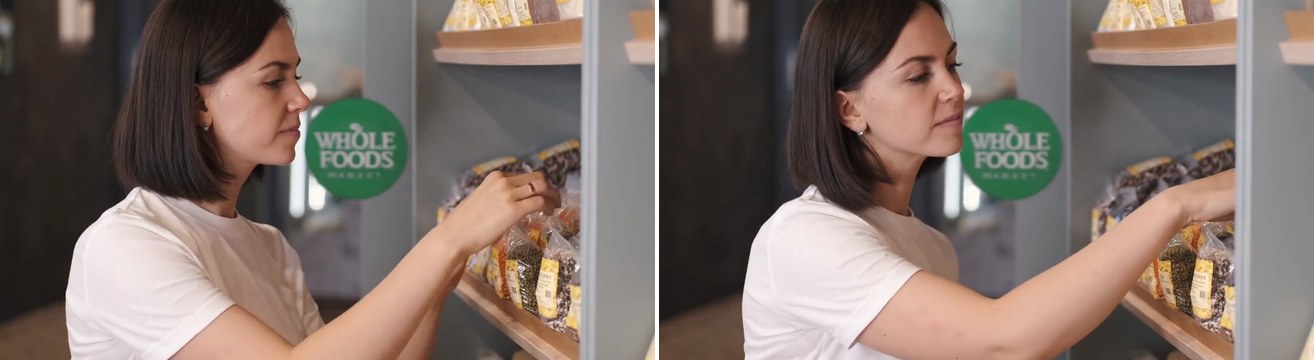}
  {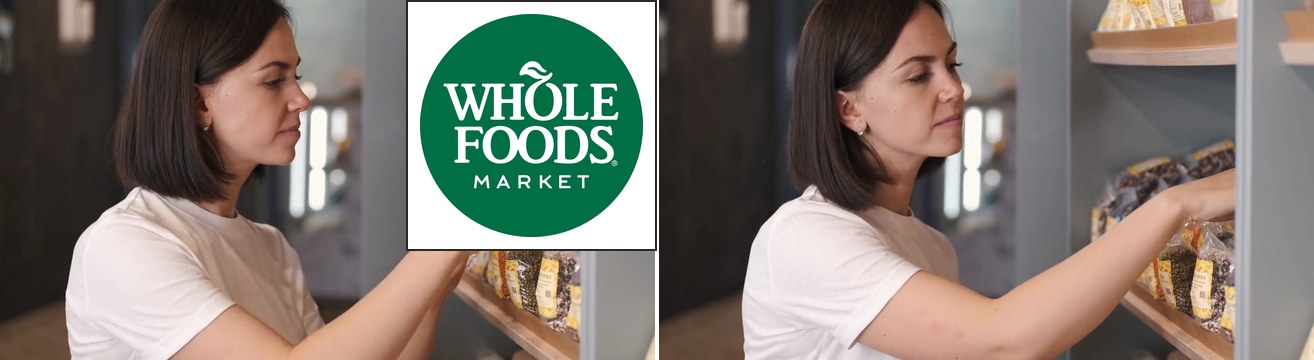}
  {Add a \benchTarget{Whole Foods Market logo} to the wall behind the woman.}

\AECRow
  {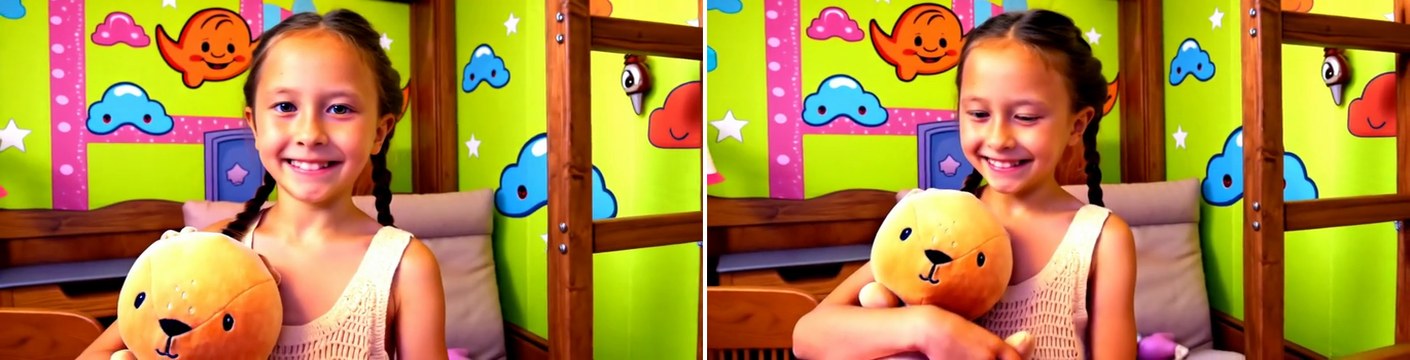}
  {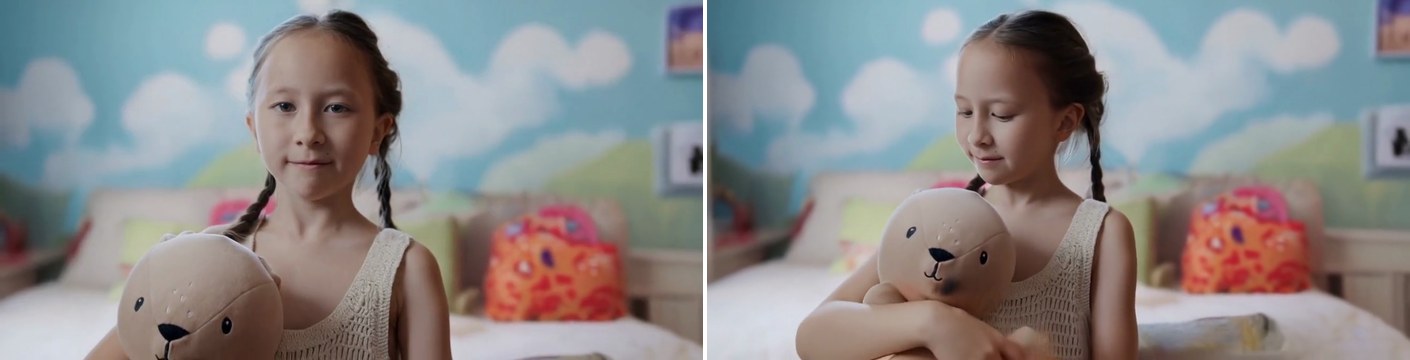}
  {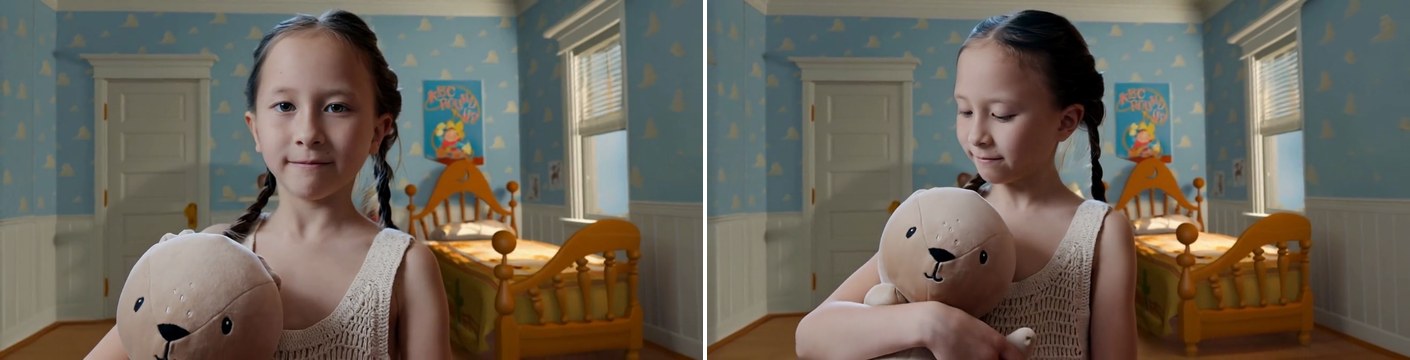}
  {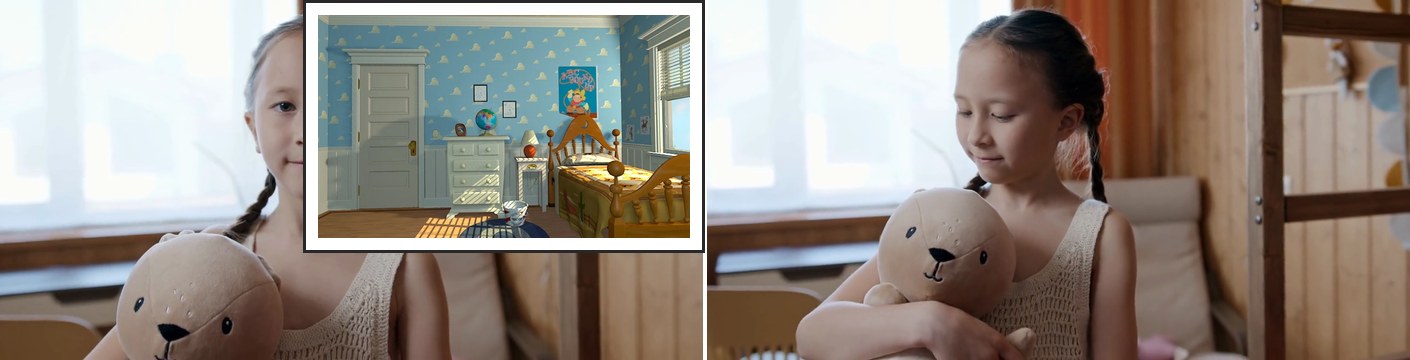}
  {Change the background to \benchTarget{Andy's room from Toy Story}, featuring the iconic cloud wallpaper, while keeping the girl and her toy unchanged.}

\AECRow
  {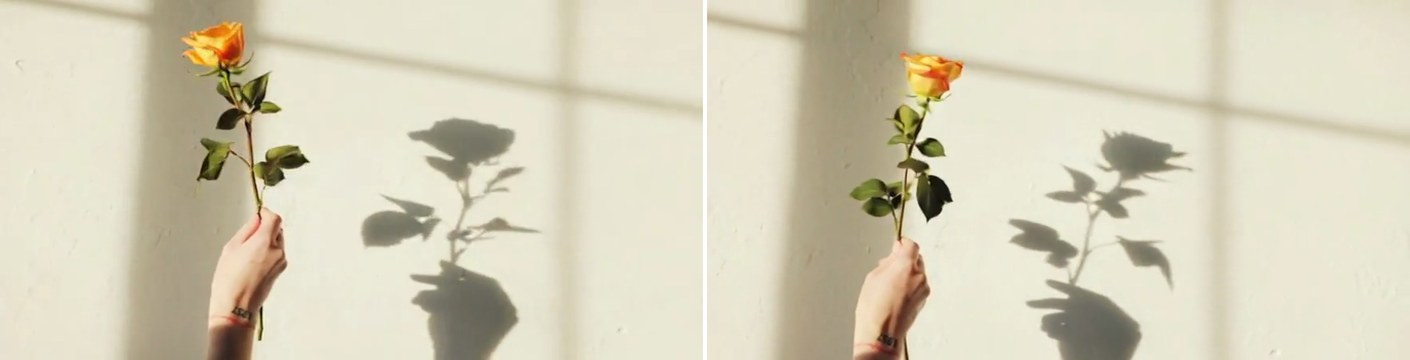}
  {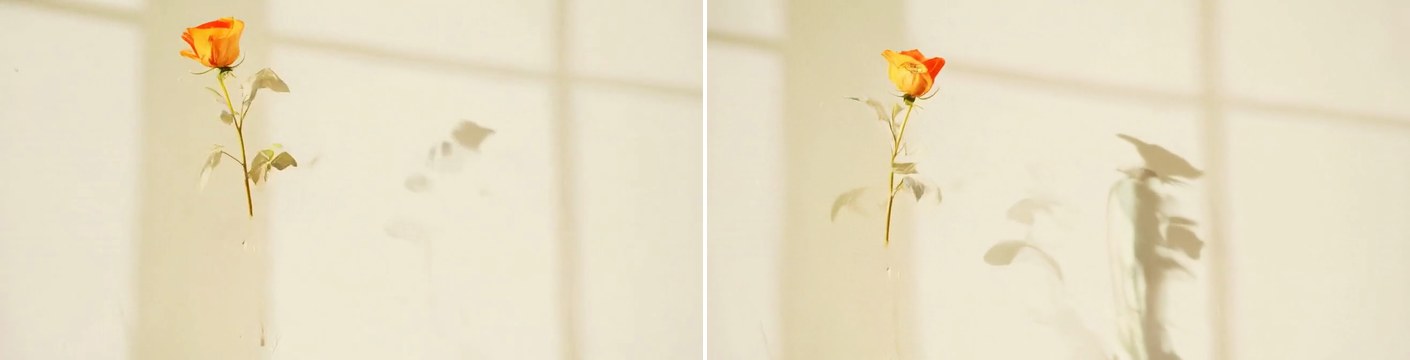}
  {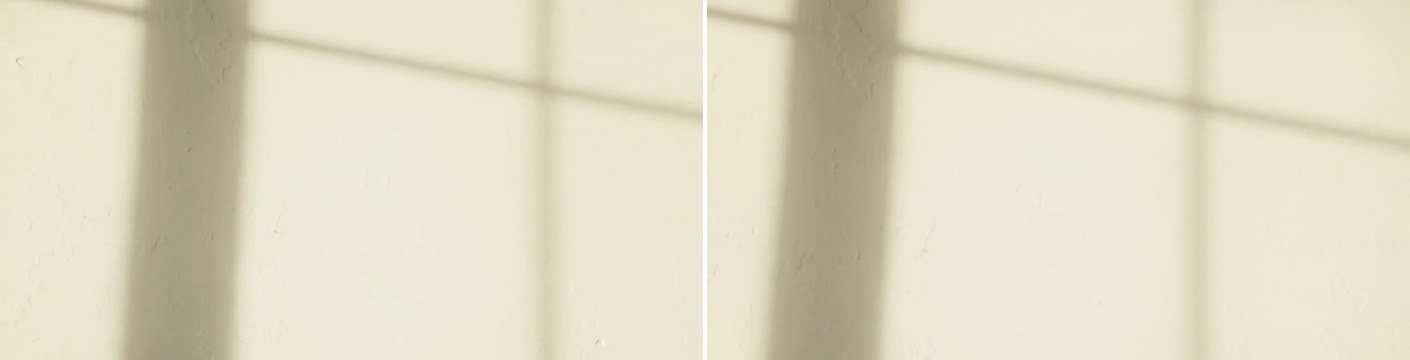}
  {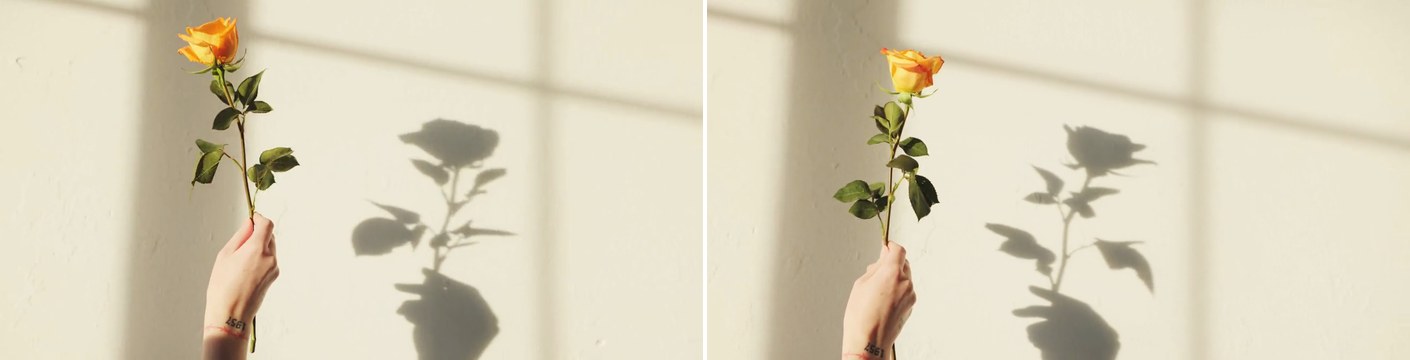}
  {Remove the \benchTarget{hand, wrist, and the yellow rose it is holding}. You must also completely erase the sharp, detailed shadow of the hand and rose cast on the wall to the right.}

\endgroup

%% file: tables/agentedit_bench.tex
\begin{table}[!t]
  \caption{\textbf{Main results on \bench{}}. Each case is scored by Gemini~2.5~Pro~\citep{gemini25pro} against a rubric, and we report the score as a percentage of that rubric's maximum (higher is better). Bold marks the best score per column.}
  \label{tab:agentedit_bench}
  \centering
  \footnotesize
  \setlength{\tabcolsep}{4.3pt}
  \renewcommand{\arraystretch}{1.06}
  \resizebox{\textwidth}{!}{%
  \begin{tabular}{lcccccc!{\vrule width 0.6pt}c}
    \toprule
    \rowcolor{benchHeader}
    \benchgroup{Method}
    & \benchgroup{\method{} Agent}
    & \benchgroup{IP Replace}
    & \benchgroup{IP Add}
    & \benchgroup{IP Background Change}
    & \benchgroup{Reasoning}
    & \benchgroup{Removal}
    & \benchgroup{Overall} \\
    \midrule
    UniVideo~\citep{univideo}
    & $\times$
    & 67.1
    & 63.4
    & 65.6
    & 77.3
    & 66.5
    & 67.0 \\
    UniVideo~\citep{univideo}
    & \checkmark
    & 80.3
    & 81.3
    & 66.8
    & 74.6
    & 68.1
    & 76.8 \\
    \addlinespace[0.3ex]
    Kiwi-Edit~\citep{kiwi}
    & $\times$
    & 66.7
    & 65.1
    & 86.6
    & 79.5
    & 69.0
    & 69.7 \\
    Kiwi-Edit~\citep{kiwi}
    & \checkmark
    & 71.6
    & 73.1
    & 78.3
    & 84.3
    & 57.8
    & 71.7 \\
    \addlinespace[0.3ex]
    \rowcolor{benchOurs}
    \textbf{\method{} (Ours)}
    & $\times$
    & 73.0
    & 68.0
    & 71.2
    & 86.4
    & 83.0
    & 74.7 \\
    \rowcolor{benchOurs}
    \textbf{\method{} (Ours)}
    & \checkmark
    & \bestscore{89.6}
    & \bestscore{85.3}
    & \bestscore{90.1}
    & \bestscore{88.1}
    & \bestscore{86.8}
    & \bestscore{87.9} \\
    \bottomrule
  \end{tabular}
  }
\end{table}

%% file: tables/standard_bench_overall.tex
\begin{wraptable}{r}{0.44\columnwidth}
  \vspace{.0\baselineskip}
  \caption{\textbf{Overall scores on existing video editing benchmarks.} Gray
  rows denote closed-source baselines; ``--'' means the score is not reported.}
  \label{tab:standard_bench_overall}
  \centering
  \footnotesize
  \setlength{\tabcolsep}{4pt}
  \renewcommand{\arraystretch}{1.05}
  \begin{tabular}{lcc}
    \toprule
    \rowcolor{benchHeader}
    \benchgroup{Method}
    & \benchgroup{EditVerse-Bench}
    & \benchgroup{OpenVE-Bench} \\
    \midrule
    \rowcolor{benchReference}
    \refscore{Runway~Aleph}~\citep{runwayaleph}
    & \refscore{7.17}
    & \refscore{3.51} \\
    \rowcolor{benchReference}
    \refscore{EditVerse}~\citep{editverse}
    & \refscore{7.52}
    & \refscore{--} \\
    \midrule
    VACE~\citep{vace}
    & 4.60
    & 1.55 \\
    InsViE~\citep{insvie}
    & --
    & 1.58 \\
    LucyEditDev~\citep{lucyedit}
    & 4.65
    & 2.12 \\
    Ditto~\citep{ditto}
    & --
    & 2.25 \\
    OpenVE-Edit~\citep{openve3m}
    & --
    & 2.57 \\
    UniVideo~\citep{univideo}
    & 6.12
    & 3.10 \\
    Senorita-2M~\citep{senorita}
    & 6.54
    & -- \\
    Kiwi-Edit~\citep{kiwi}
    & 7.00
    & -- \\
    \rowcolor{benchOurs}
    \textbf{\method{} (Ours)}
    & \bestscore{7.61}
    & \bestscore{3.38} \\
    \bottomrule
  \end{tabular}
  \vspace{-1.0\baselineskip}
\end{wraptable}

%% file: tables/agent_dpo_ablation.tex
\begin{wraptable}{r}{0.28\columnwidth}
  \vspace{-1.2\baselineskip}
  \centering
  \footnotesize
  \setlength{\tabcolsep}{8pt}
  \renewcommand{\arraystretch}{1.05}
  \caption{\textbf{Ablation of agent training stages} on \bench{}.}
  \label{tab:agent_dpo_ablation}
  \begin{tabular}{lc}
    \toprule
    \rowcolor{benchHeader}
    \benchgroup{Configuration} & \benchgroup{Overall} \\
    \midrule
    \method{}, no agent    & 74.7 \\
    $+$ agent, SFT only    & 85.0 \\
    \rowcolor{benchOurs}
    $+$ agent, SFT $+$ DPO & \bestscore{87.9} \\
    \bottomrule
  \end{tabular}
  \vspace{-1.0\baselineskip}
\end{wraptable}

%% file: tables/agent_transfer.tex
\begin{table}[!t]
  \caption{\textbf{Agent transfer ablation} across video editing models on
  EditVerse-Bench~\citep{editverse} and
  OpenVE-Bench~\citep{openve3m}. 
  $\dagger$ denotes the five-category OpenVE-Bench average excluding Subtitle removal.}
  \label{tab:agent_transfer}
  \centering
  \footnotesize
  \setlength{\tabcolsep}{7pt}
  \renewcommand{\arraystretch}{1.0}
  \begin{minipage}[t]{0.245\textwidth}
    \centering
    \subcaption*{(a) OpenVE-Bench$\dagger$}
    \begin{tabular}{lc}
      \toprule
      \rowcolor{benchHeader}
      \benchgroup{Setting} & \benchgroup{Overall} \\
      \midrule
      Kiwi-Edit & 3.02 \\
      \rowcolor{benchOurs}
      $+$ \method{} agent & \bestscore{3.29} \\
      \midrule
      $\Delta$ & \textbf{$+0.27$} \\
      \bottomrule
    \end{tabular}
  \end{minipage}\hfill
  \begin{minipage}[t]{0.245\textwidth}
    \centering
    \subcaption*{(b) OpenVE-Bench$\dagger$}
    \begin{tabular}{lc}
      \toprule
      \rowcolor{benchHeader}
      \benchgroup{Setting} & \benchgroup{Overall} \\
      \midrule
      \method{} (Ours) & 3.31 \\
      \rowcolor{benchOurs}
      $+$ \method{} agent & \bestscore{3.46} \\
      \midrule
      $\Delta$ & \textbf{$+0.15$} \\
      \bottomrule
    \end{tabular}
  \end{minipage}\hfill
  \begin{minipage}[t]{0.245\textwidth}
    \centering
    \subcaption*{(c) EditVerse-Bench}
    \begin{tabular}{lc}
      \toprule
      \rowcolor{benchHeader}
      \benchgroup{Setting} & \benchgroup{Overall} \\
      \midrule
      UniVideo & 6.12 \\
      \rowcolor{benchOurs}
      $+$ \method{} agent & \bestscore{6.48} \\
      \midrule
      $\Delta$ & \textbf{$+0.36$} \\
      \bottomrule
    \end{tabular}
  \end{minipage}\hfill
  \begin{minipage}[t]{0.245\textwidth}
    \centering
    \subcaption*{(d) EditVerse-Bench}
    \begin{tabular}{lc}
      \toprule
      \rowcolor{benchHeader}
      \benchgroup{Setting} & \benchgroup{Overall} \\
      \midrule
      \method{} (Ours) & 7.25 \\
      \rowcolor{benchOurs}
      $+$ \method{} agent & \bestscore{7.61} \\
      \midrule
      $\Delta$ & \textbf{$+0.36$} \\
      \bottomrule
    \end{tabular}
  \end{minipage}
\end{table}

%% file: supp/figs/agentedit_qualitative_portrait_b.tex
\begingroup
\setlength{\tabcolsep}{0pt}
\renewcommand{\arraystretch}{1.0}

\providecommand{\AEPbStrip}[1]{\filmstrip[\linewidth]{#1}}
\providecommand{\AEPbHead}[1]{{\fontsize{6.6}{7.6}\selectfont\sffamily\centering #1\par}}
\providecommand{\AEPbHeadOurs}[1]{{\fontsize{6.6}{7.6}\selectfont\sffamily\bfseries\centering #1\par}}
\providecommand{\AEPbPrompt}[1]{{\fontsize{7.4}{8.6}\selectfont\itshape\raggedright #1\par}}

\providecommand{\AEPbHeaderRow}{%
  \noindent\begin{tabular}{@{}p{0.241\textwidth}@{\hspace{0.012\textwidth}}p{0.241\textwidth}@{\hspace{0.012\textwidth}}p{0.241\textwidth}@{\hspace{0.012\textwidth}}p{0.241\textwidth}@{}}
    \AEPbHead{UniVideo} & \AEPbHead{KiwiEdit} & \AEPbHeadOurs{\method{} (Ours)} & \AEPbHead{Source\,/\,Reference} \\
  \end{tabular}\par
  \vspace{0.30ex}
}

\providecommand{\AEPbRow}[5]{%
  \noindent\begin{tabular}{@{}p{0.241\textwidth}@{\hspace{0.012\textwidth}}p{0.241\textwidth}@{\hspace{0.012\textwidth}}p{0.241\textwidth}@{\hspace{0.012\textwidth}}p{0.241\textwidth}@{}}
    \AEPbStrip{#1} & \AEPbStrip{#2} & \AEPbStrip{#3} & \AEPbStrip{#4} \\
  \end{tabular}\par
  \vspace{-0.2ex}\AEPbPrompt{#5}\vspace{0.55ex}
}

\AEPbHeaderRow

\AEPbRow
  {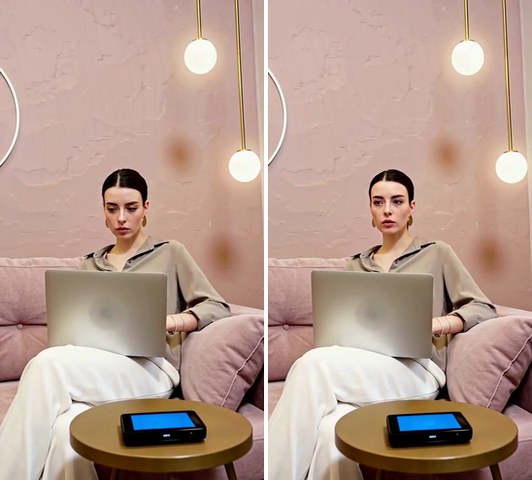}
  {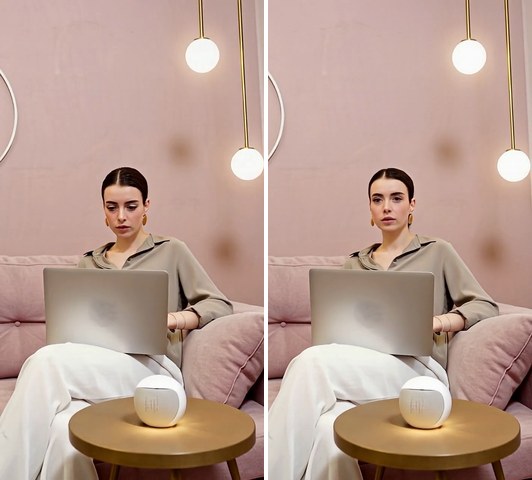}
  {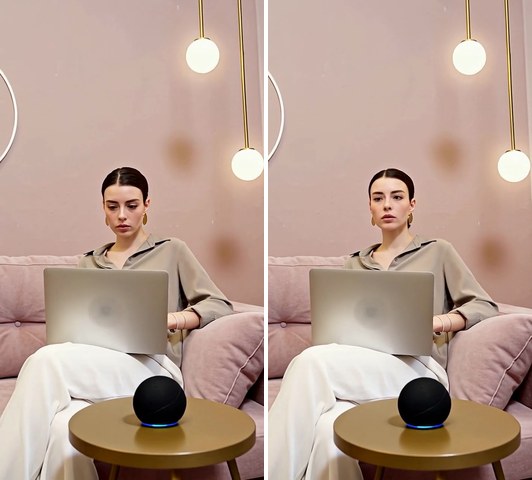}
  {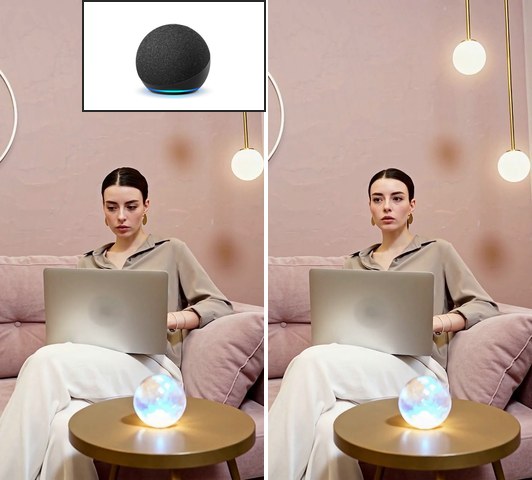}
  {Replace the glowing sphere on the table with an \benchTarget{Amazon Echo Dot 4th Gen}.}

\AEPbRow
  {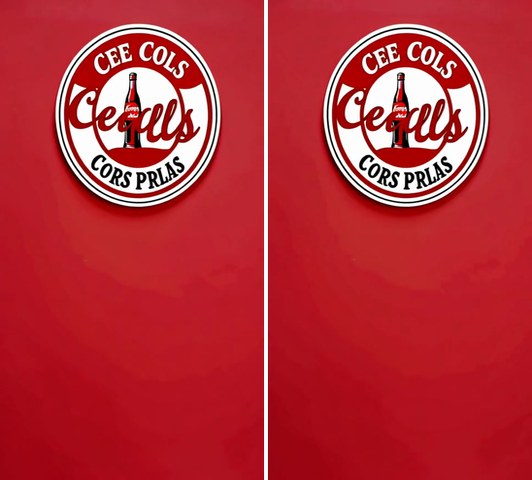}
  {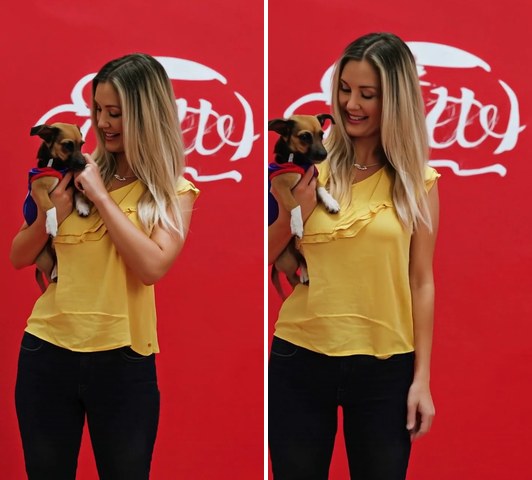}
  {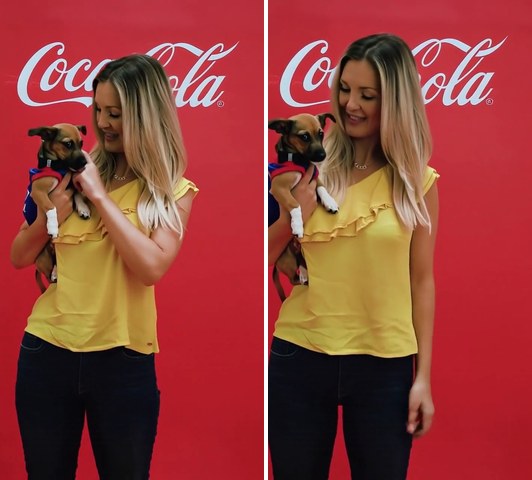}
  {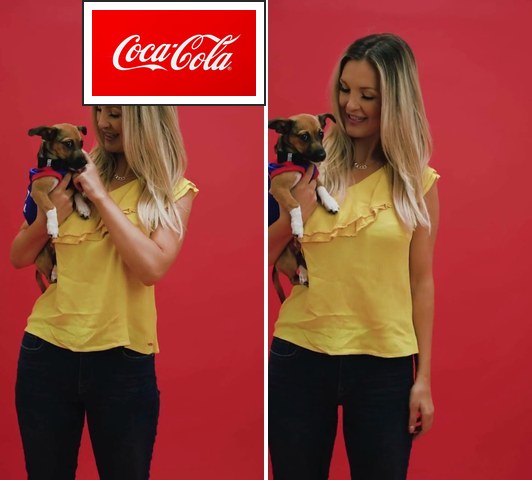}
  {Add a classic \benchTarget{Coca-Cola logo} to the red background behind the woman.}

\AEPbRow
  {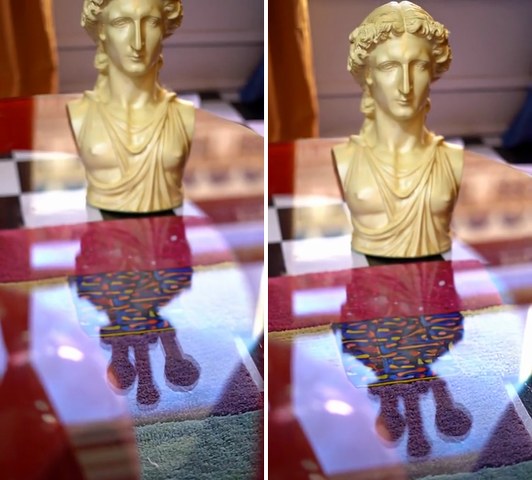}
  {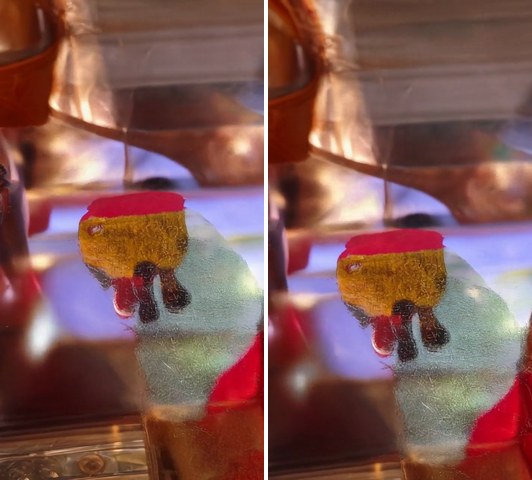}
  {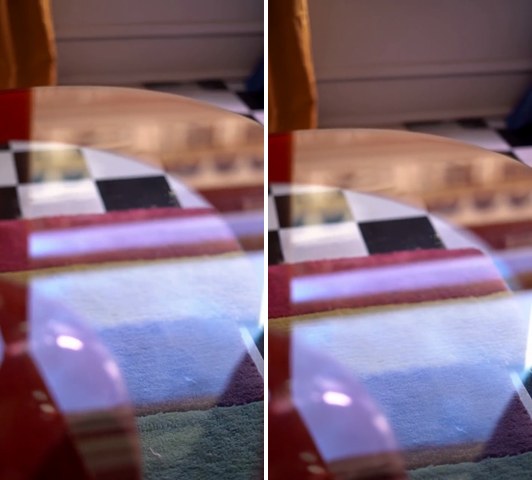}
  {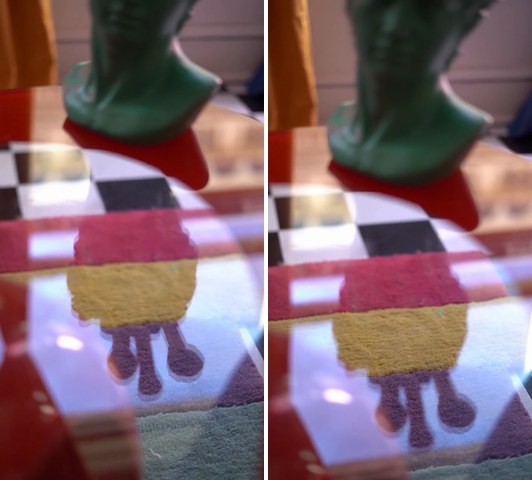}
  {Remove the \benchTarget{green bust of David sitting on the table}, including the makeup brushes sticking out of its head. You must also completely remove its shadow and its reflection on the glossy, patterned table surface, accurately reconstructing the colorful pattern underneath.}

\endgroup

%% file: supp/figs/agentedit_qualitative_portrait_a.tex
\begingroup
\setlength{\tabcolsep}{0pt}
\renewcommand{\arraystretch}{1.0}

\providecommand{\AEPaStrip}[1]{\filmstrip[\linewidth]{#1}}
\providecommand{\AEPaHead}[1]{{\fontsize{6.6}{7.6}\selectfont\sffamily\centering #1\par}}
\providecommand{\AEPaHeadOurs}[1]{{\fontsize{6.6}{7.6}\selectfont\sffamily\bfseries\centering #1\par}}
\providecommand{\AEPaPrompt}[1]{{\fontsize{7.4}{8.6}\selectfont\itshape\raggedright #1\par}}

\providecommand{\AEPaHeaderRow}{%
  \noindent\begin{tabular}{@{}p{0.241\textwidth}@{\hspace{0.012\textwidth}}p{0.241\textwidth}@{\hspace{0.012\textwidth}}p{0.241\textwidth}@{\hspace{0.012\textwidth}}p{0.241\textwidth}@{}}
    \AEPaHead{UniVideo} & \AEPaHead{KiwiEdit} & \AEPaHeadOurs{\method{} (Ours)} & \AEPaHead{Source\,/\,Reference} \\
  \end{tabular}\par
  \vspace{0.30ex}
}

\providecommand{\AEPaRow}[5]{%
  \noindent\begin{tabular}{@{}p{0.241\textwidth}@{\hspace{0.012\textwidth}}p{0.241\textwidth}@{\hspace{0.012\textwidth}}p{0.241\textwidth}@{\hspace{0.012\textwidth}}p{0.241\textwidth}@{}}
    \AEPaStrip{#1} & \AEPaStrip{#2} & \AEPaStrip{#3} & \AEPaStrip{#4} \\
  \end{tabular}\par
  \vspace{-0.2ex}\AEPaPrompt{#5}\vspace{0.55ex}
}

\AEPaHeaderRow

\AEPaRow
  {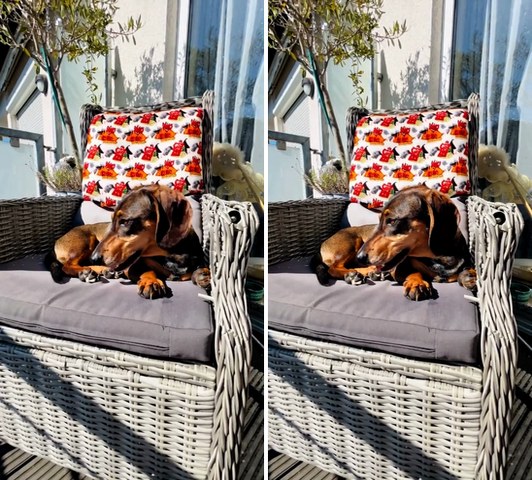}
  {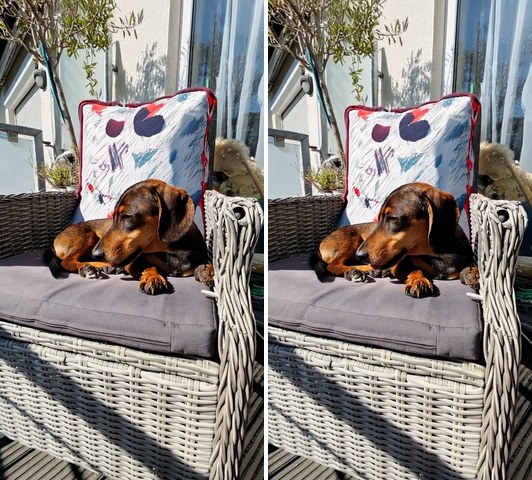}
  {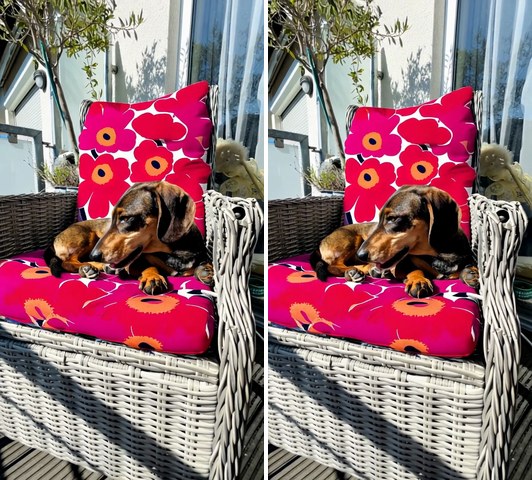}
  {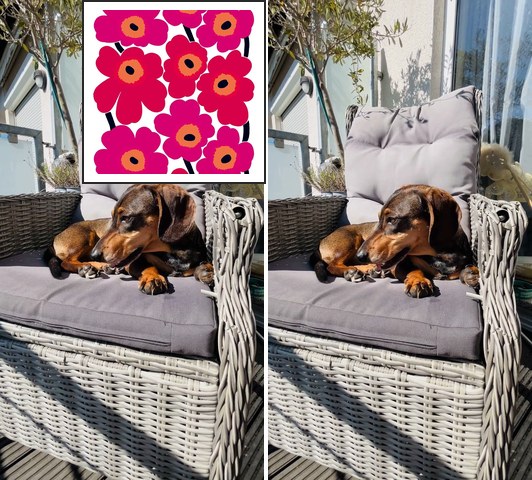}
  {Replace the plain grey back cushion on the chair with a \benchTarget{Marimekko Unikko patterned cushion}.}

\AEPaRow
  {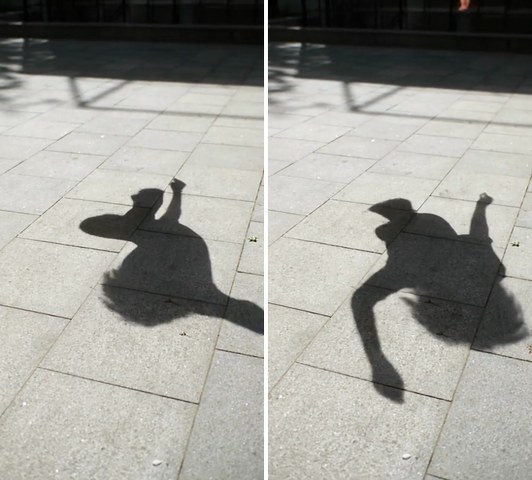}
  {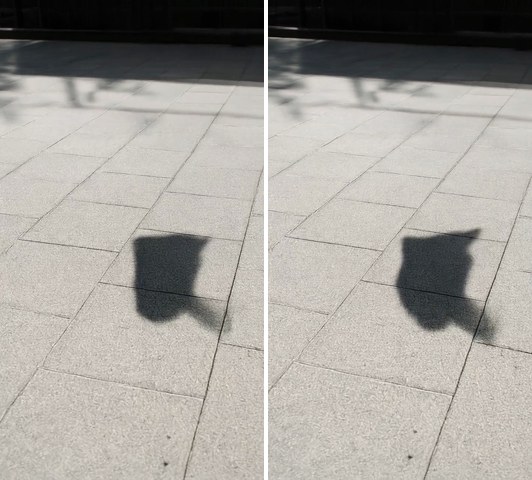}
  {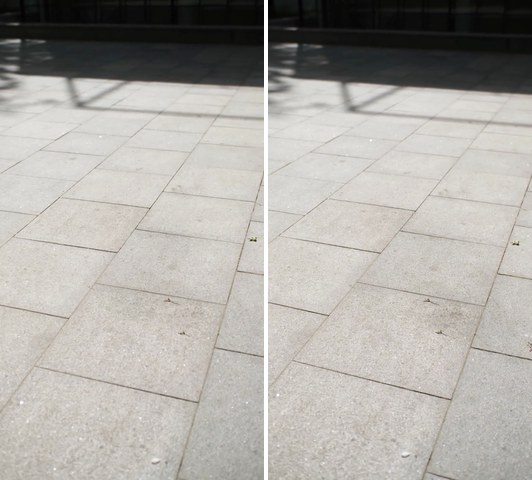}
  {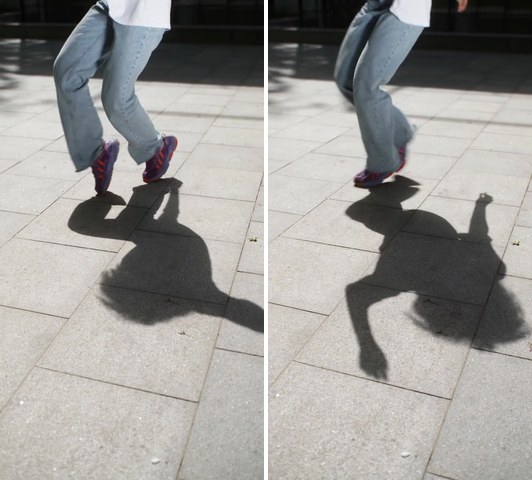}
  {Remove the \benchTarget{person's lower body wearing light blue jeans and purple shoes dancing in the frame}. You must also completely remove their large, dynamic shadow cast on the tiled pavement, ensuring the grid lines of the tiles are reconstructed seamlessly without any ghosting.}

\AEPaRow
  {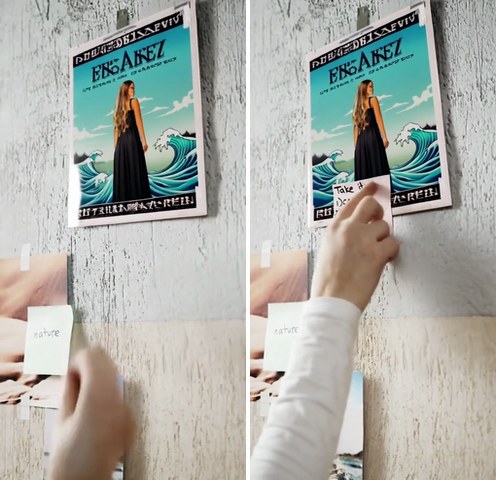}
  {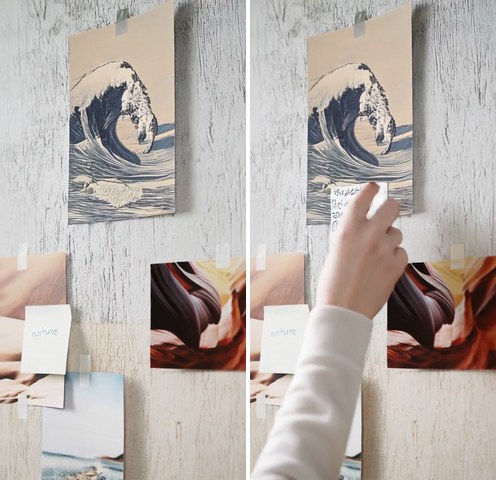}
  {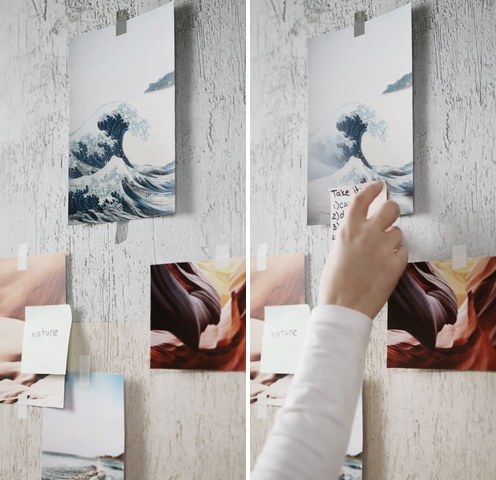}
  {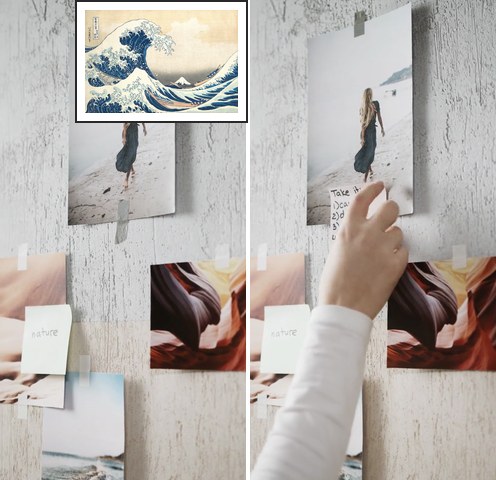}
  {Replace the top-most photograph of the woman on the beach with a print of \benchTarget{`The Great Wave off Kanagawa'} while maintaining the same size and tape placement.}

\endgroup

%% file: supp/figs/agentedit_qualitative_landscape.tex
\begingroup
\setlength{\tabcolsep}{0pt}
\renewcommand{\arraystretch}{1.0}

\providecommand{\AELStrip}[1]{\filmstrip[\linewidth]{#1}}
\providecommand{\AELHead}[1]{{\fontsize{6.6}{7.6}\selectfont\sffamily\centering #1\par}}
\providecommand{\AELHeadOurs}[1]{{\fontsize{6.6}{7.6}\selectfont\sffamily\bfseries\centering #1\par}}
\providecommand{\AELPrompt}[1]{{\fontsize{7.4}{8.6}\selectfont\itshape\raggedright #1\par}}

\providecommand{\AELHeaderRow}{%
  \noindent\begin{tabular}{@{}p{0.241\textwidth}@{\hspace{0.012\textwidth}}p{0.241\textwidth}@{\hspace{0.012\textwidth}}p{0.241\textwidth}@{\hspace{0.012\textwidth}}p{0.241\textwidth}@{}}
    \AELHead{UniVideo} & \AELHead{KiwiEdit} & \AELHeadOurs{\method{} (Ours)} & \AELHead{Source\,/\,Reference} \\
  \end{tabular}\par
  \vspace{0.30ex}
}

\providecommand{\AELRow}[5]{%
  \noindent\begin{tabular}{@{}p{0.241\textwidth}@{\hspace{0.012\textwidth}}p{0.241\textwidth}@{\hspace{0.012\textwidth}}p{0.241\textwidth}@{\hspace{0.012\textwidth}}p{0.241\textwidth}@{}}
    \AELStrip{#1} & \AELStrip{#2} & \AELStrip{#3} & \AELStrip{#4} \\
  \end{tabular}\par
  \vspace{-0.2ex}\AELPrompt{#5}\vspace{0.55ex}
}

\AELHeaderRow

\AELRow
  {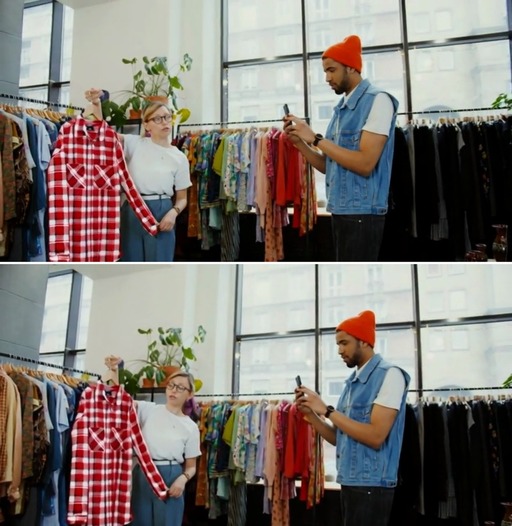}
  {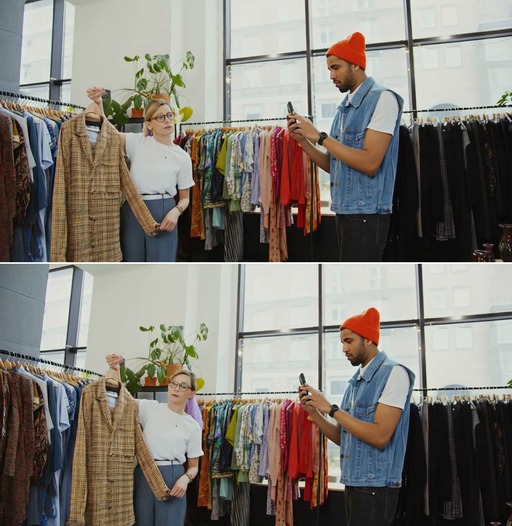}
  {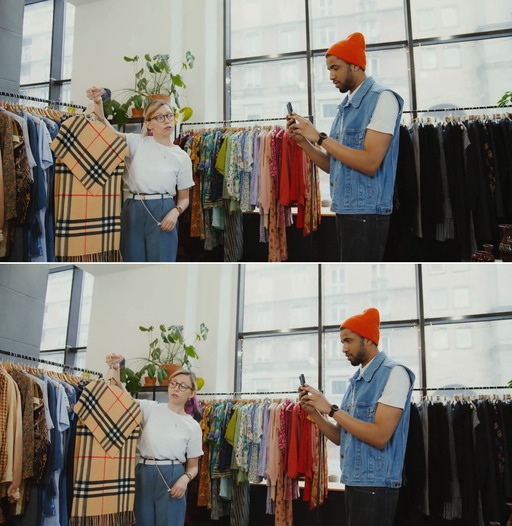}
  {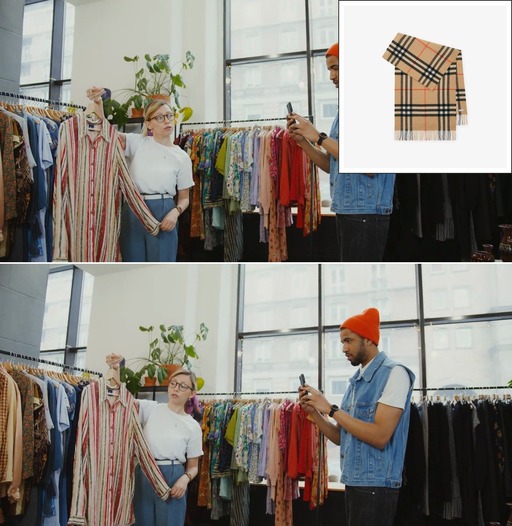}
  {Replace the striped shirt held by the woman with a \benchTarget{Burberry check pattern scarf}, maintaining the way it hangs on the hanger.}

\AELRow
  {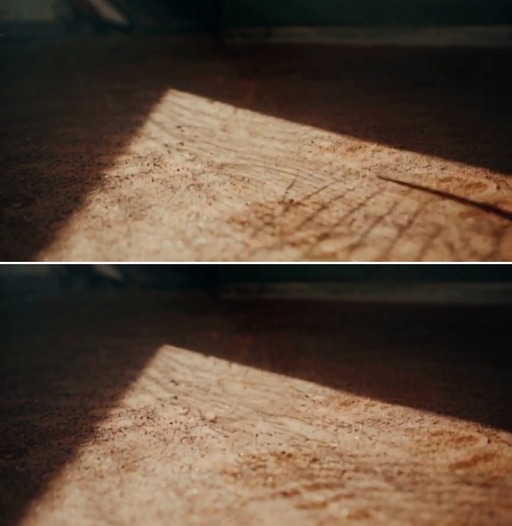}
  {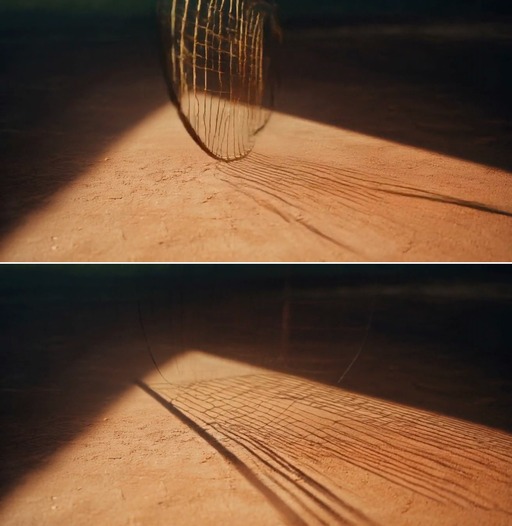}
  {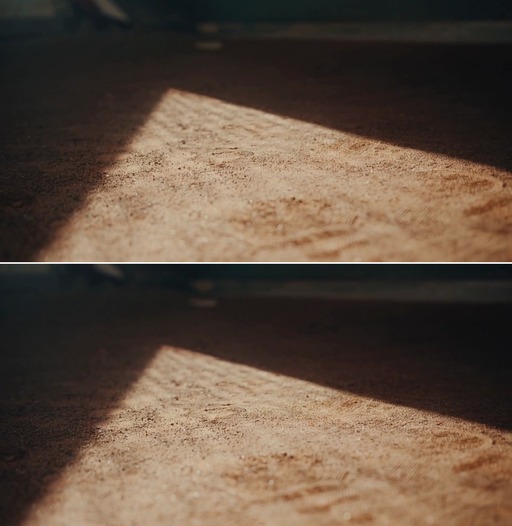}
  {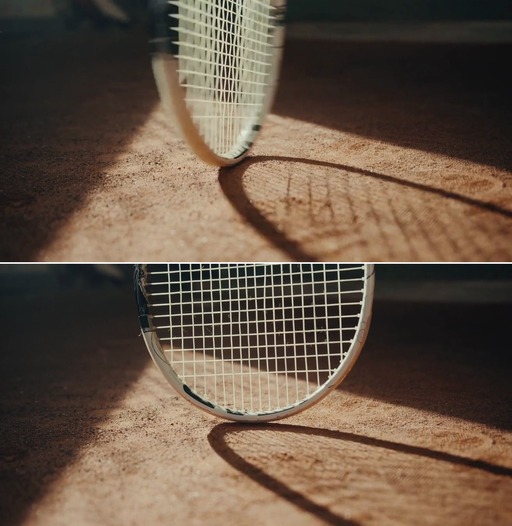}
  {Remove the \benchTarget{white tennis racket resting vertically on the court}. You must also completely erase the intricate shadow cast by its frame and strings onto the ground, replacing it with the natural, continuous texture of the clay surface.}

\AELRow
  {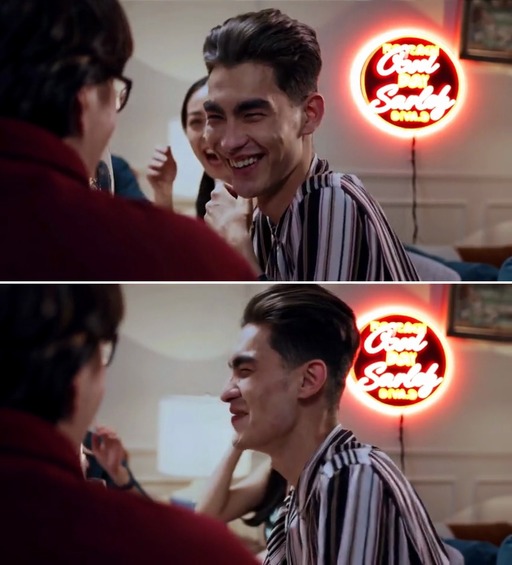}
  {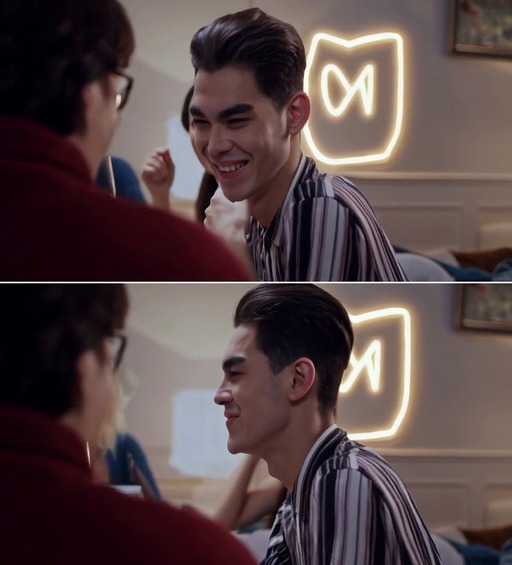}
  {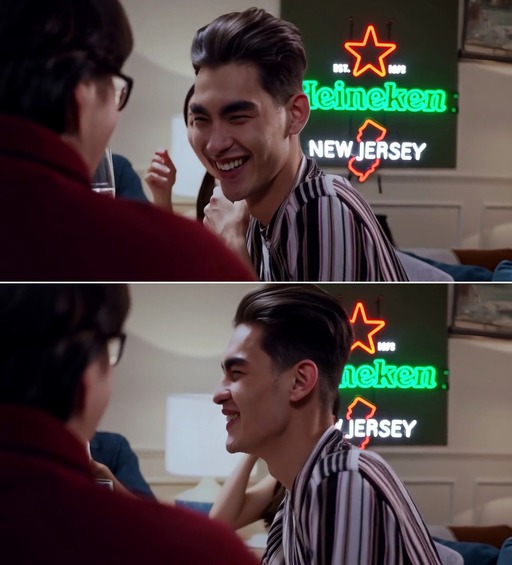}
  {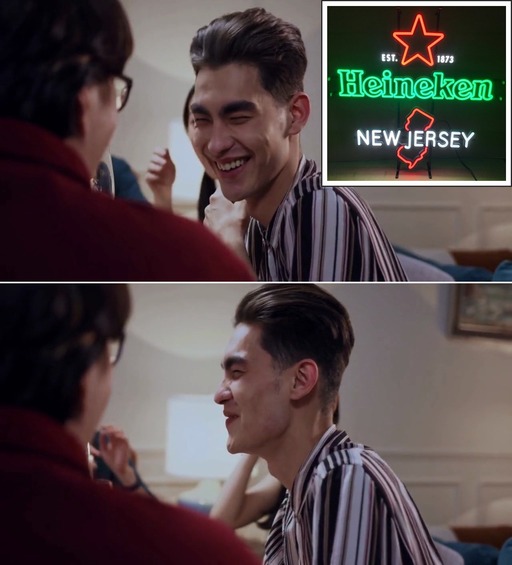}
  {Add a glowing \benchTarget{Heineken neon sign} to the empty wall space behind the man on the right.}

\AELRow
  {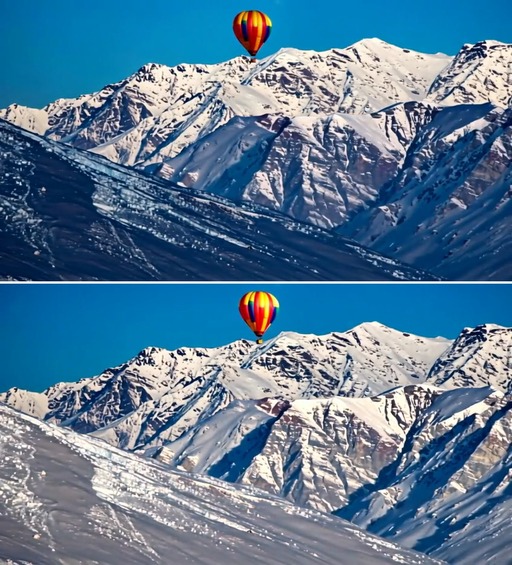}
  {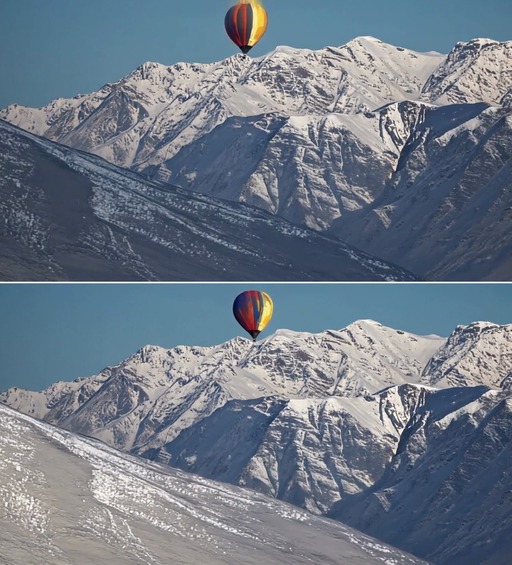}
  {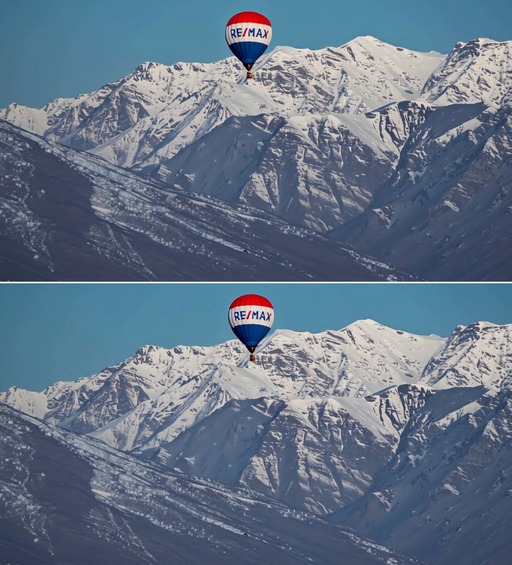}
  {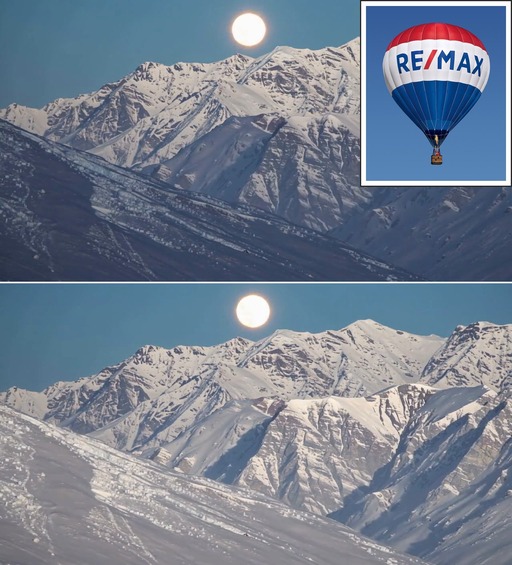}
  {Add a \benchTarget{RE/MAX hot air balloon} floating in the clear sky above the mountains.}

\endgroup

%% file: supp/tables/editverse_bench.tex
\begin{table}[!t]
  \caption{Per-category quantitative comparison on EditVerse-Bench~\citep{editverse}. Scores are
  obtained using Gemini~2.5~Pro~\citep{gemini25pro}. Gray rows denote closed-source reference methods; bold marks the best
  open-source score per column. Overall scores are summarized in
  Table~\ref{tab:standard_bench_overall}.}
  \label{tab:appendix:editverse_bench}
  \centering
  \footnotesize
  \setlength{\tabcolsep}{3.1pt}
  \renewcommand{\arraystretch}{1.05}
  \resizebox{\textwidth}{!}{%
  \begin{tabular}{lcccccccccccc!{\vrule width 0.6pt}c}
    \toprule
    \rowcolor{benchHeader}
    \benchgroup{Method}
    & \benchgroup{Add} & \benchgroup{Remove} & \benchgroup{Change} & \benchgroup{Style} & \benchgroup{ID} & \benchgroup{Reason}
    & \benchgroup{Bkg.} & \benchgroup{Color} & \benchgroup{Material} & \benchgroup{Effect} & \benchgroup{Weather}
    & \benchgroup{Combined}
    & \benchgroup{Overall} \\
    \midrule
    \rowcolor{benchReference}
    \refscore{Runway~Aleph}~\citep{runwayaleph}
    & \refscore{6.68}
    & \refscore{7.40}
    & \refscore{7.70}
    & \refscore{7.80}
    & \refscore{6.45}
    & \refscore{7.85}
    & \refscore{7.93}
    & \refscore{7.95}
    & \refscore{6.70}
    & \refscore{6.03}
    & \refscore{6.40}
    & \refscore{7.13}
    & \refscore{7.17} \\
    \rowcolor{benchReference}
    \refscore{EditVerse}~\citep{editverse}
    & \refscore{6.93}
    & \refscore{8.10}
    & \refscore{7.40}
    & \refscore{8.73}
    & \refscore{7.50}
    & \refscore{7.73}
    & \refscore{7.68}
    & \refscore{7.90}
    & \refscore{7.30}
    & \refscore{6.25}
    & \refscore{7.53}
    & \refscore{7.25}
    & \refscore{7.52} \\
    \midrule
    VACE~\citep{vace}
    & 2.90
    & 2.28
    & 4.85
    & 7.83
    & 4.18
    & 4.98
    & 5.85
    & 6.30
    & 4.00
    & 3.23
    & 4.80
    & 3.98
    & 4.60 \\
    LucyEditDev~\citep{lucyedit}
    & 4.67
    & 3.70
    & 4.00
    & 4.40
    & 5.30
    & 3.27
    & 4.63
    & 7.50
    & 5.27
    & 4.33
    & 4.30
    & 4.47
    & 4.65 \\
    UniVideo~\citep{univideo}
    & 6.00
    & 6.20
    & 5.73
    & 7.80
    & 7.00
    & 5.70
    & 6.97
    & 5.87
    & 4.80
    & 5.30
    & 6.27
    & 5.80
    & 6.12 \\
    Senorita-2M~\citep{senorita}
    & 5.83
    & 7.13
    & 6.53
    & 8.33
    & 6.63
    & 6.63
    & 6.63
    & 7.20
    & 6.07
    & 5.17
    & 6.40
    & 5.93
    & 6.54 \\
    Kiwi-Edit~\citep{kiwi}
    & 6.73
    & 6.85
    & \bestscore{7.53}
    & 8.43
    & 6.75
    & 7.38
    & 6.93
    & \bestscore{8.10}
    & 6.33
    & 6.13
    & 7.18
    & 5.65
    & 7.00 \\
    \rowcolor{benchOurs}
    \textbf{\method{} (Ours)}
    & \bestscore{7.35}
    & \bestscore{8.00}
    & 7.03
    & \bestscore{8.63}
    & \bestscore{7.68}
    & \bestscore{7.70}
    & \bestscore{7.93}
    & 7.73
    & \bestscore{7.15}
    & \bestscore{6.58}
    & \bestscore{7.93}
    & \bestscore{7.55}
    & \bestscore{7.61} \\
    \bottomrule
  \end{tabular}
  }
\end{table}

%% file: supp/figs/qualitative_appendix.tex
\begingroup
\setlength{\tabcolsep}{0pt}
\renewcommand{\arraystretch}{1.0}

\providecommand{\QACell}[1]{\filmstrip[\linewidth]{#1}}
\providecommand{\QACellT}[1]{\filmstripT[\linewidth]{#1}}
\providecommand{\QACellB}[1]{\filmstripB[\linewidth]{#1}}
\providecommand{\QAHead}[1]{{\fontsize{6.4}{7.2}\selectfont\sffamily\centering #1\par}}
\providecommand{\QAHeadOurs}[1]{{\fontsize{6.4}{7.2}\selectfont\sffamily\bfseries\centering #1\par}}
\providecommand{\QAPrompt}[1]{{\fontsize{7.2}{8.4}\selectfont\itshape\raggedright #1\par}}

\providecommand{\QAHeaderRow}{%
  \noindent\begin{tabular}{@{}p{0.162\textwidth}@{\hspace{0.0055\textwidth}}p{0.162\textwidth}@{\hspace{0.0055\textwidth}}p{0.162\textwidth}@{\hspace{0.0055\textwidth}}p{0.162\textwidth}@{\hspace{0.0055\textwidth}}p{0.162\textwidth}@{\hspace{0.0055\textwidth}}p{0.162\textwidth}@{}}
    \QAHead{LucyEditDev} & \QAHead{VACE} & \QAHead{KiwiEdit} & \QAHead{UniVideo} & \QAHeadOurs{\method{} (Ours)} & \QAHead{Source} \\
  \end{tabular}\par
  \vspace{0.30ex}
}

\providecommand{\QARow}[6]{%
  \noindent\begin{tabular}{@{}p{0.162\textwidth}@{\hspace{0.0055\textwidth}}p{0.162\textwidth}@{\hspace{0.0055\textwidth}}p{0.162\textwidth}@{\hspace{0.0055\textwidth}}p{0.162\textwidth}@{\hspace{0.0055\textwidth}}p{0.162\textwidth}@{\hspace{0.0055\textwidth}}p{0.162\textwidth}@{}}
    \QACell{#1} & \QACell{#2} & \QACell{#3} & \QACell{#4} & \QACell{#5} & \QACell{#6} \\
  \end{tabular}\par
  \vspace{0.20ex}
}

\providecommand{\QARowT}[6]{%
  \noindent\begin{tabular}{@{}p{0.162\textwidth}@{\hspace{0.0055\textwidth}}p{0.162\textwidth}@{\hspace{0.0055\textwidth}}p{0.162\textwidth}@{\hspace{0.0055\textwidth}}p{0.162\textwidth}@{\hspace{0.0055\textwidth}}p{0.162\textwidth}@{\hspace{0.0055\textwidth}}p{0.162\textwidth}@{}}
    \QACellT{#1} & \QACellT{#2} & \QACellT{#3} & \QACellT{#4} & \QACellT{#5} & \QACellT{#6} \\
  \end{tabular}\par\nointerlineskip
}
\providecommand{\QARowB}[6]{%
  \noindent\begin{tabular}{@{}p{0.162\textwidth}@{\hspace{0.0055\textwidth}}p{0.162\textwidth}@{\hspace{0.0055\textwidth}}p{0.162\textwidth}@{\hspace{0.0055\textwidth}}p{0.162\textwidth}@{\hspace{0.0055\textwidth}}p{0.162\textwidth}@{\hspace{0.0055\textwidth}}p{0.162\textwidth}@{}}
    \QACellB{#1} & \QACellB{#2} & \QACellB{#3} & \QACellB{#4} & \QACellB{#5} & \QACellB{#6} \\
  \end{tabular}\par
  \vspace{0.20ex}
}

\providecommand{\QACasePrompt}[1]{%
  \vspace{-0.1ex}\QAPrompt{#1}\vspace{1.10ex}
}

\providecommand{\QACase}[2]{%
  \QARowT
    {supp/imgs/qualitative_appendix/#1_lucyedit_f1.jpg}
    {supp/imgs/qualitative_appendix/#1_vace_f1.jpg}
    {supp/imgs/qualitative_appendix/#1_kiwiedit_f1.jpg}
    {supp/imgs/qualitative_appendix/#1_univideo_f1.jpg}
    {supp/imgs/qualitative_appendix/#1_aurora_f1.jpg}
    {supp/imgs/qualitative_appendix/#1_source_f1.jpg}
  \QARowB
    {supp/imgs/qualitative_appendix/#1_lucyedit_f3.jpg}
    {supp/imgs/qualitative_appendix/#1_vace_f3.jpg}
    {supp/imgs/qualitative_appendix/#1_kiwiedit_f3.jpg}
    {supp/imgs/qualitative_appendix/#1_univideo_f3.jpg}
    {supp/imgs/qualitative_appendix/#1_aurora_f3.jpg}
    {supp/imgs/qualitative_appendix/#1_source_f3.jpg}
  \QACasePrompt{#2}
}

\providecommand{\QPCase}[2]{%
  \QARow
    {supp/imgs/qualitative_appendix_portrait/#1_lucyedit.jpg}
    {supp/imgs/qualitative_appendix_portrait/#1_vace.jpg}
    {supp/imgs/qualitative_appendix_portrait/#1_kiwiedit.jpg}
    {supp/imgs/qualitative_appendix_portrait/#1_univideo.jpg}
    {supp/imgs/qualitative_appendix_portrait/#1_aurora.jpg}
    {supp/imgs/qualitative_appendix_portrait/#1_source.jpg}
  \QACasePrompt{#2}
}

\QAHeaderRow

\QACase{c_id_car}{Insert a \benchTarget{car \texttt{<image1>}} driving on the road.}

\QPCase{b_bg_water}{Turn the grass into \benchTarget{a reflective water surface}.}

\QPCase{d_rsn_car}{Change the blue item to \benchTarget{a vehicle that runs much faster}.}

\endgroup

%% file: supp/tables/openve_bench.tex
\begin{table}[!t]
  \caption{Per-category quantitative comparison on
  OpenVE-Bench~\citep{openve3m}, judged by
  Gemini~2.5~Pro~\citep{gemini25pro} on a 1--5 scale. $\dagger$ marks zero-shot
  generalization: \method{} is not trained on subtitle removal data. Overall
  scores are summarized in Table~\ref{tab:standard_bench_overall}.}
  \label{tab:appendix:openve_bench}
  \centering
  \footnotesize
  \setlength{\tabcolsep}{3.2pt}
  \renewcommand{\arraystretch}{1.05}
  \resizebox{\textwidth}{!}{%
  \begin{tabular}{lccccccc!{\vrule width 0.6pt}c}
    \toprule
    \rowcolor{benchHeader}
    \benchgroup{Method} & \benchgroup{\#Params.}
    & \benchgroup{Global Style} & \benchgroup{Bkg. Change}
    & \benchgroup{Local Change} & \benchgroup{Local Remove}
    & \benchgroup{Local Add} & \benchgroup{Subtitle}
    & \benchgroup{Overall} \\
    \midrule
    \rowcolor{benchReference}
    \refscore{Runway~Aleph}~\citep{runwayaleph}
    & \refscore{--}
    & \refscore{3.72} & \refscore{2.62} & \refscore{4.18}
    & \refscore{4.16} & \refscore{2.78} & \refscore{3.62}
    & \refscore{3.51} \\
    \midrule
    VACE~\citep{vace}
    & 14B
    & 1.49 & 1.55 & 2.07 & 1.46 & 1.26 & 1.48
    & 1.55 \\
    InsViE~\citep{insvie}
    & 2B
    & 2.20 & 1.06 & 1.48 & 1.36 & 1.17 & 2.18
    & 1.58 \\
    Lucy-Edit~\citep{lucyedit}
    & 5B
    & 2.27 & 1.57 & 3.20 & 1.75 & 2.30 & 1.61
    & 2.12 \\
    Ditto~\citep{ditto}
    & 14B
    & 4.01 & 1.68 & 2.03 & 1.53 & 1.41 & 2.81
    & 2.25 \\
    OpenVE-Edit~\citep{openve3m}
    & 5B
    & 3.16 & 2.36 & 2.98 & 1.85 & 2.15 & 2.91
    & 2.57 \\
    Kiwi-Edit~\citep{kiwi}
    & 5B
    & 3.64 & 2.64 & 3.83 & 2.63 & 2.36 & --
    & -- \\
    UniVideo~\citep{univideo}
    & 13B
    & 3.75 & 2.38 & 3.78 & 2.93 & \bestscore{2.87} & 2.91
    & 3.10 \\
    \rowcolor{benchOurs}
    \textbf{\method{} (Ours)}
    & 5B
    & \bestscore{4.07} & \bestscore{2.67} & \bestscore{4.03}
    & \bestscore{3.73} & 2.78 & \bestscore{2.97}$^{\dagger}$
    & \bestscore{3.38} \\
    \bottomrule
  \end{tabular}
  }
\end{table}

%% file: supp/figs/openve_qualitative.tex
\begingroup
\setlength{\tabcolsep}{0pt}
\renewcommand{\arraystretch}{1.0}

\providecommand{\OQCell}[1]{\filmstrip[\linewidth]{#1}}
\providecommand{\OQCellT}[1]{\filmstripT[\linewidth]{#1}}
\providecommand{\OQCellB}[1]{\filmstripB[\linewidth]{#1}}
\providecommand{\OQHead}[1]{{\fontsize{7.0}{8.2}\selectfont\sffamily\centering #1\par}}
\providecommand{\OQHeadOurs}[1]{{\fontsize{7.0}{8.2}\selectfont\sffamily\bfseries\centering #1\par}}
\providecommand{\OQPrompt}[1]{{\fontsize{7.4}{8.6}\selectfont\itshape\raggedright #1\par}}

\providecommand{\OQHeaderRow}{%
  \noindent\begin{tabular}{@{}p{0.243\textwidth}@{\hspace{0.0083\textwidth}}p{0.243\textwidth}@{\hspace{0.0083\textwidth}}p{0.243\textwidth}@{\hspace{0.0083\textwidth}}p{0.243\textwidth}@{}}
    \OQHead{UniVideo} & \OQHead{KiwiEdit} & \OQHeadOurs{\method{} (Ours)} & \OQHead{Source} \\
  \end{tabular}\par
  \vspace{0.30ex}
}

\providecommand{\OQRowT}[4]{%
  \noindent\begin{tabular}{@{}p{0.243\textwidth}@{\hspace{0.0083\textwidth}}p{0.243\textwidth}@{\hspace{0.0083\textwidth}}p{0.243\textwidth}@{\hspace{0.0083\textwidth}}p{0.243\textwidth}@{}}
    \OQCellT{#1} & \OQCellT{#2} & \OQCellT{#3} & \OQCellT{#4} \\
  \end{tabular}\par\nointerlineskip
}
\providecommand{\OQRowB}[4]{%
  \noindent\begin{tabular}{@{}p{0.243\textwidth}@{\hspace{0.0083\textwidth}}p{0.243\textwidth}@{\hspace{0.0083\textwidth}}p{0.243\textwidth}@{\hspace{0.0083\textwidth}}p{0.243\textwidth}@{}}
    \OQCellB{#1} & \OQCellB{#2} & \OQCellB{#3} & \OQCellB{#4} \\
  \end{tabular}\par\vspace{0.20ex}
}
\providecommand{\OQRow}[4]{%
  \noindent\begin{tabular}{@{}p{0.243\textwidth}@{\hspace{0.0083\textwidth}}p{0.243\textwidth}@{\hspace{0.0083\textwidth}}p{0.243\textwidth}@{\hspace{0.0083\textwidth}}p{0.243\textwidth}@{}}
    \OQCell{#1} & \OQCell{#2} & \OQCell{#3} & \OQCell{#4} \\
  \end{tabular}\par\vspace{0.20ex}
}

\providecommand{\OQCasePrompt}[1]{%
  \vspace{-0.1ex}\OQPrompt{#1}\vspace{1.10ex}
}

\providecommand{\OQCase}[2]{%
  \OQRowT
    {supp/imgs/openve_qualitative/#1_univideo_f1.jpg}
    {supp/imgs/openve_qualitative/#1_kiwi_f1.jpg}
    {supp/imgs/openve_qualitative/#1_aurora_f1.jpg}
    {supp/imgs/openve_qualitative/#1_source_f1.jpg}
  \OQRowB
    {supp/imgs/openve_qualitative/#1_univideo_f3.jpg}
    {supp/imgs/openve_qualitative/#1_kiwi_f3.jpg}
    {supp/imgs/openve_qualitative/#1_aurora_f3.jpg}
    {supp/imgs/openve_qualitative/#1_source_f3.jpg}
  \OQCasePrompt{#2}
}

\providecommand{\OQCaseNoSrc}[2]{%
  \OQRowT
    {supp/imgs/openve_qualitative/#1_univideo_f1.jpg}
    {supp/imgs/openve_qualitative/#1_kiwi_f1.jpg}
    {supp/imgs/openve_qualitative/#1_aurora_f1.jpg}
    {supp/imgs/openve_qualitative/_no_source_landscape.jpg}
  \OQRowB
    {supp/imgs/openve_qualitative/#1_univideo_f3.jpg}
    {supp/imgs/openve_qualitative/#1_kiwi_f3.jpg}
    {supp/imgs/openve_qualitative/#1_aurora_f3.jpg}
    {supp/imgs/openve_qualitative/_no_source_landscape.jpg}
  \OQCasePrompt{#2}
}

\providecommand{\OQPCase}[2]{%
  \OQRow
    {supp/imgs/openve_qualitative/#1_univideo_strip.jpg}
    {supp/imgs/openve_qualitative/#1_kiwi_strip.jpg}
    {supp/imgs/openve_qualitative/#1_aurora_strip.jpg}
    {supp/imgs/openve_qualitative/#1_source_strip.jpg}
  \OQCasePrompt{#2}
}

\OQHeaderRow

\OQCase{b_remove_redhair}{Remove \benchTarget{the person with long, flowing red hair wearing a light brown, textured coat with a wide collar and tied waist belt} from the entire video sequence. The background should be replaced with a clean, uniform light gray studio backdrop.}

\OQPCase{c_add_vase}{Insert \benchTarget{a small, hand-painted ceramic vase} onto the dark wooden surface, slightly behind and to the left of the white candle holder. The vase has a glossy white glaze with a delicate blue floral pattern around its midsection.}

\OQCase{d_marble_table}{Replace the smooth white table with \benchTarget{a luxurious marble surface table}.}

\OQCase{e_red_challenger}{Replace the dark green Dodge Challenger muscle car on the left with \benchTarget{a sleek metallic red Dodge Challenger muscle car}.}

\endgroup

%% file: supp/figs/agent_ablation_univideo.tex
\begingroup
\setlength{\tabcolsep}{0pt}
\renewcommand{\arraystretch}{1.0}

\providecommand{\AAStrip}[1]{\filmstrip[\linewidth]{#1}}
\providecommand{\AAHead}[1]{{\fontsize{6.6}{7.6}\selectfont\sffamily\centering #1\par}}
\providecommand{\AAPrompt}[1]{{\fontsize{7.4}{8.6}\selectfont\itshape\raggedright #1\par}}

\providecommand{\AAHeaderRow}{%
  \noindent\begin{tabular}{@{}p{0.155\textwidth}@{\hspace{0.010\textwidth}}p{0.155\textwidth}@{\hspace{0.010\textwidth}}p{0.155\textwidth}@{\hspace{0.020\textwidth}}p{0.155\textwidth}@{\hspace{0.010\textwidth}}p{0.155\textwidth}@{\hspace{0.010\textwidth}}p{0.155\textwidth}@{}}
    \AAHead{w/o agent} & \AAHead{w/ agent} & \AAHead{Source\,/\,Reference}
  & \AAHead{w/o agent} & \AAHead{w/ agent} & \AAHead{Source\,/\,Reference} \\
  \end{tabular}\par
  \vspace{0.30ex}
}

\providecommand{\AARow}[6]{%
  \noindent\begin{tabular}{@{}p{0.155\textwidth}@{\hspace{0.010\textwidth}}p{0.155\textwidth}@{\hspace{0.010\textwidth}}p{0.155\textwidth}@{\hspace{0.020\textwidth}}p{0.155\textwidth}@{\hspace{0.010\textwidth}}p{0.155\textwidth}@{\hspace{0.010\textwidth}}p{0.155\textwidth}@{}}
    \AAStrip{#1} & \AAStrip{#2} & \AAStrip{#3} & \AAStrip{#4} & \AAStrip{#5} & \AAStrip{#6} \\
  \end{tabular}\par\vspace{-0.2ex}
}

\providecommand{\AAPromptRow}[2]{%
  \noindent\begin{tabular}{@{}p{0.485\textwidth}@{\hspace{0.020\textwidth}}p{0.485\textwidth}@{}}
    \AAPrompt{#1} & \AAPrompt{#2} \\
  \end{tabular}\par\vspace{0.55ex}
}

\AAHeaderRow

\AARow
  {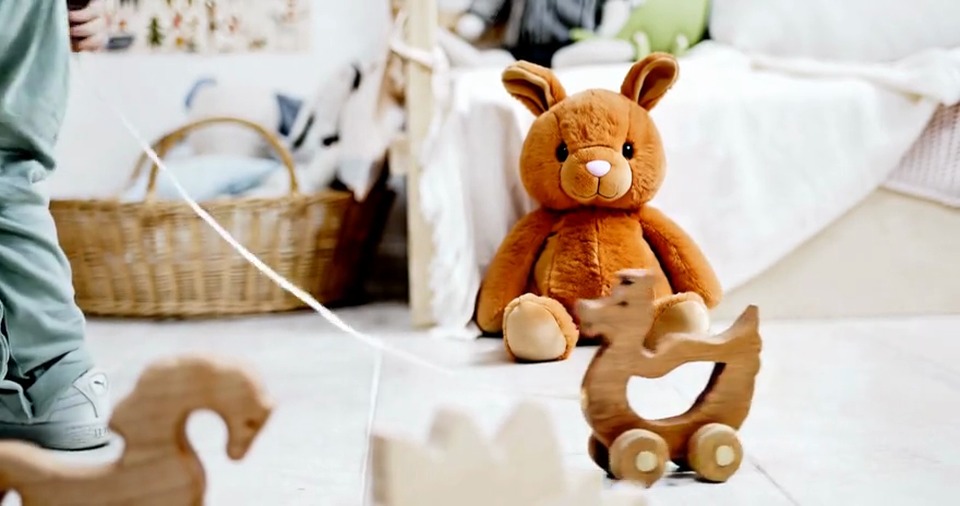}
  {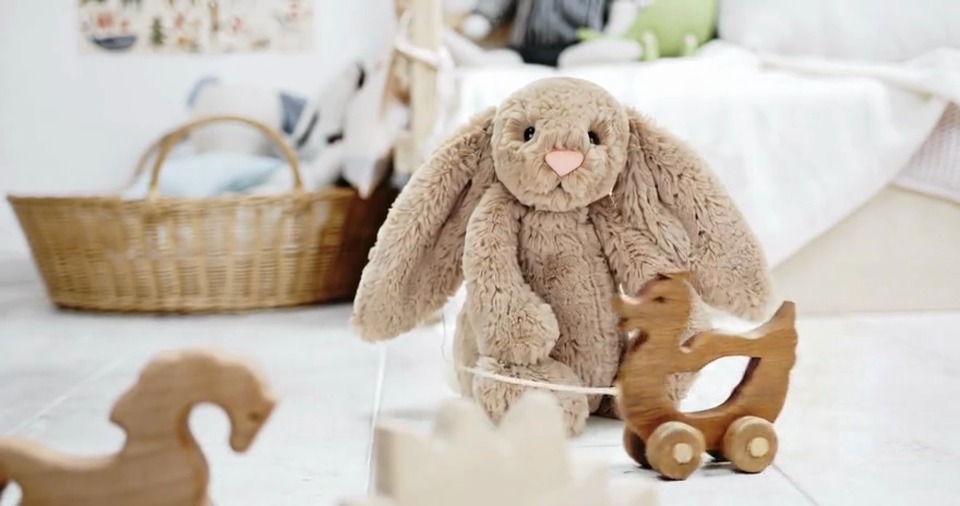}
  {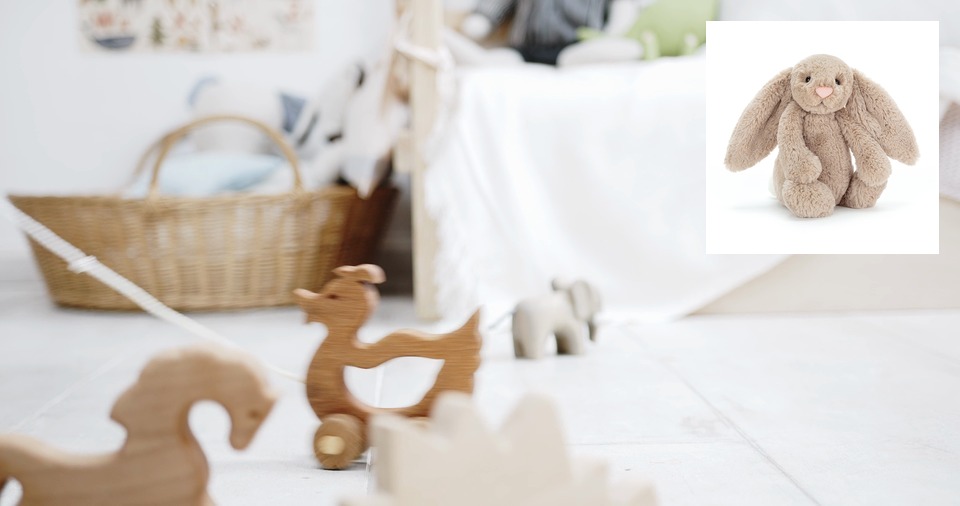}
  {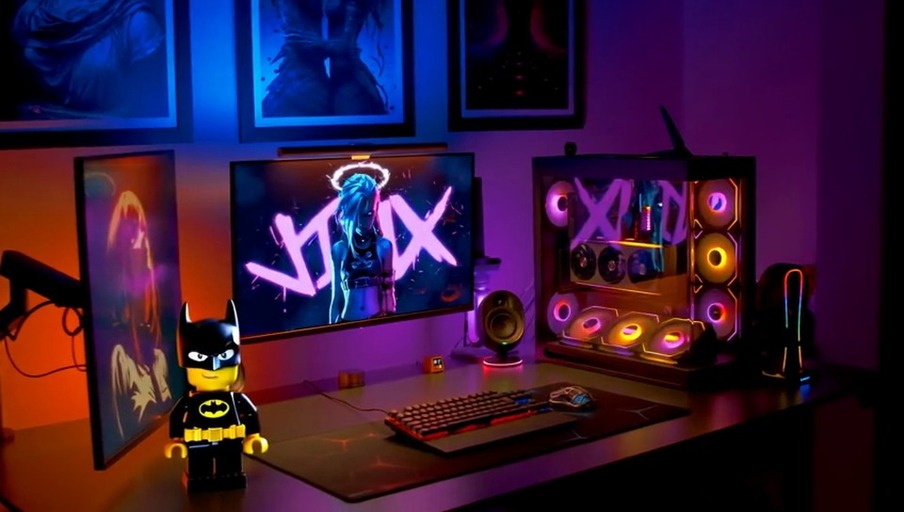}
  {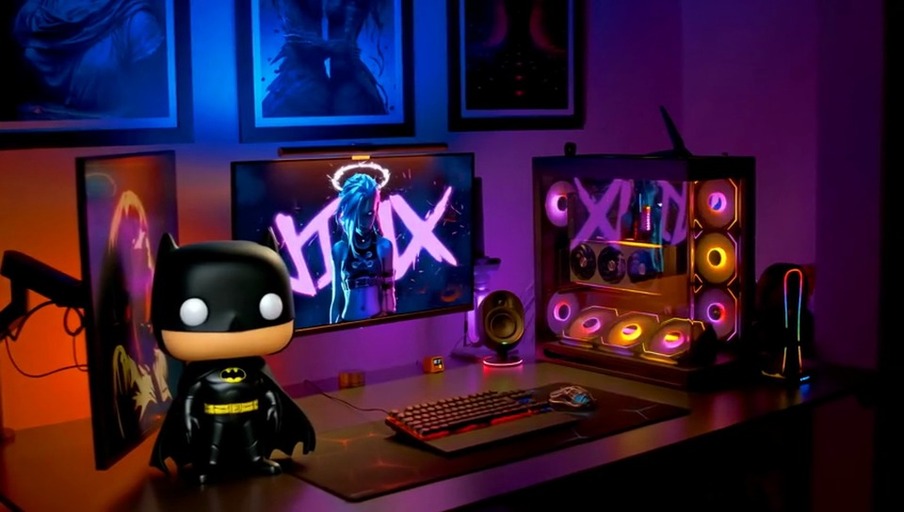}
  {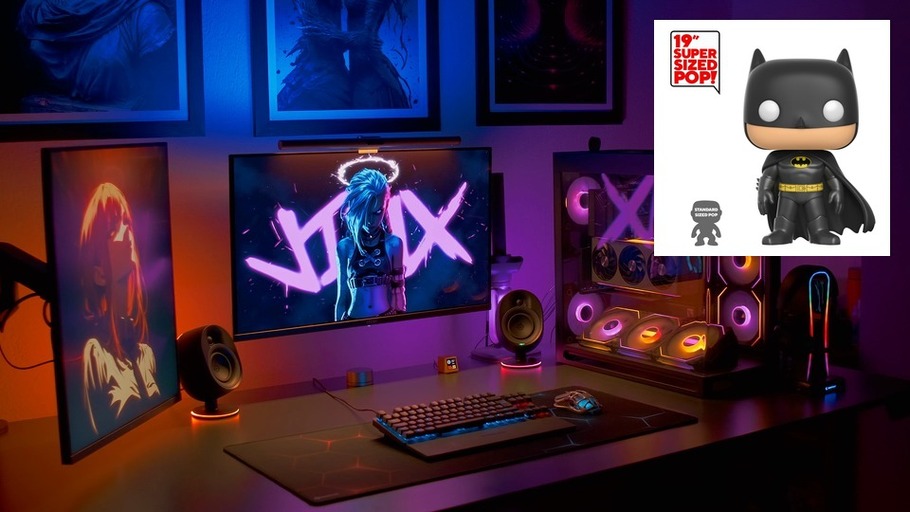}

\AAPromptRow
  {Add a \benchTarget{Jellycat Bashful Bunny plush toy} sitting inside the woven basket in the background.}
  {Add a \benchTarget{Batman Funko Pop figure} on the desk to the left of the keyboard, ensuring it matches the scene's purple and orange lighting.}

\AARow
  {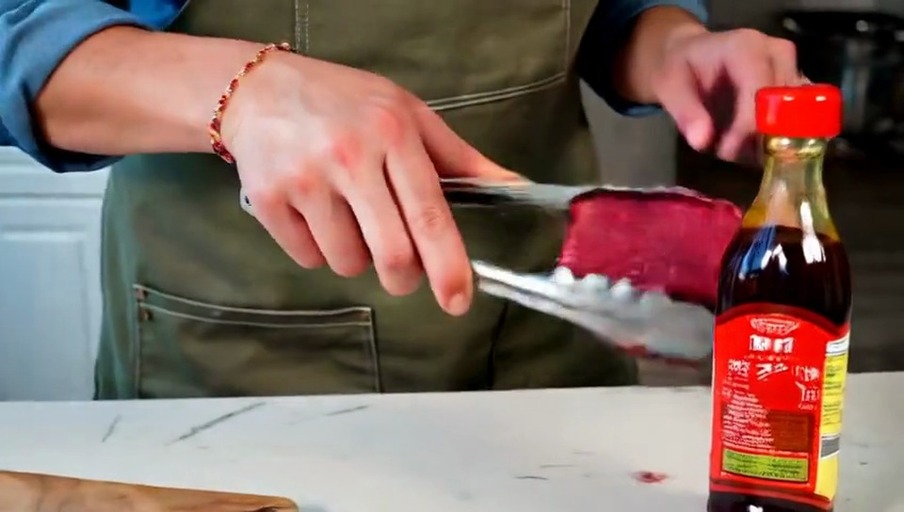}
  {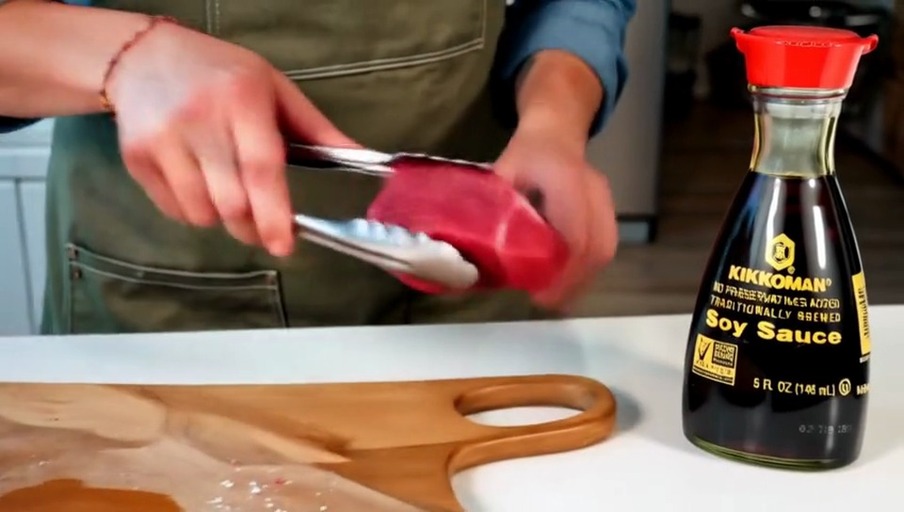}
  {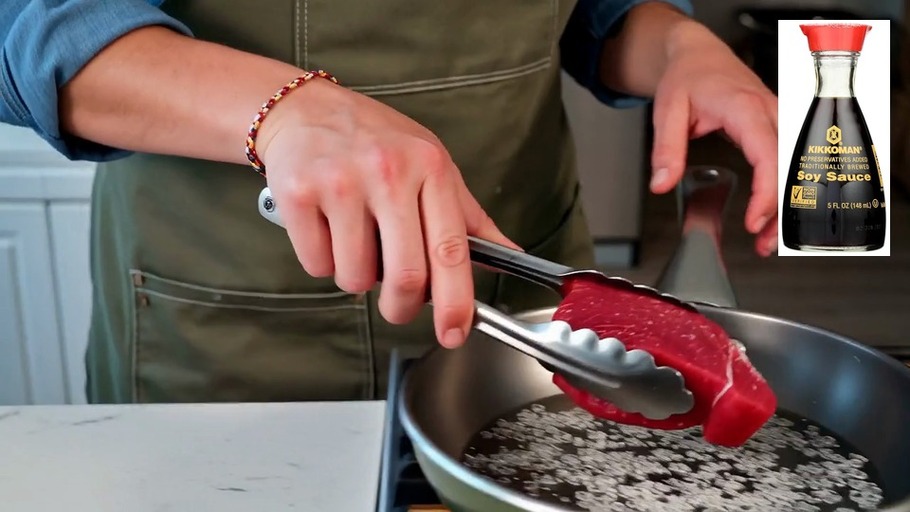}
  {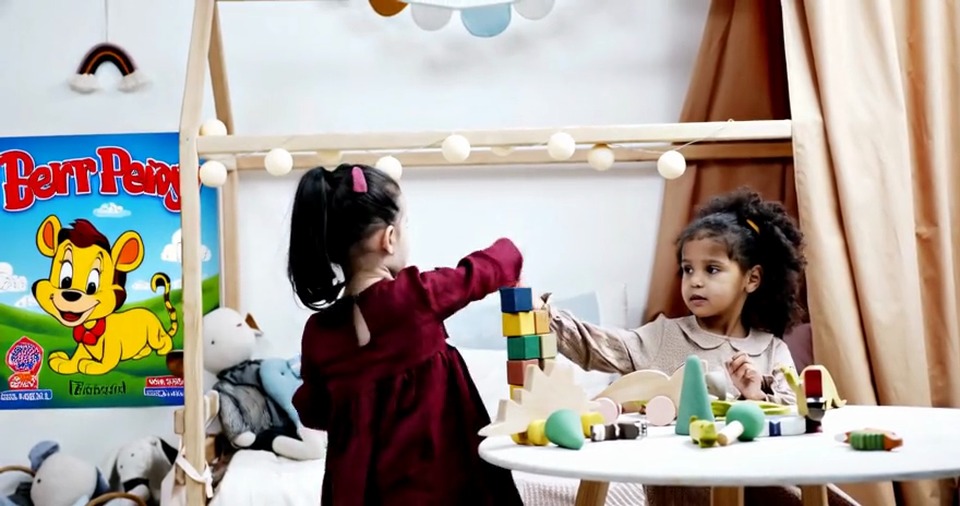}
  {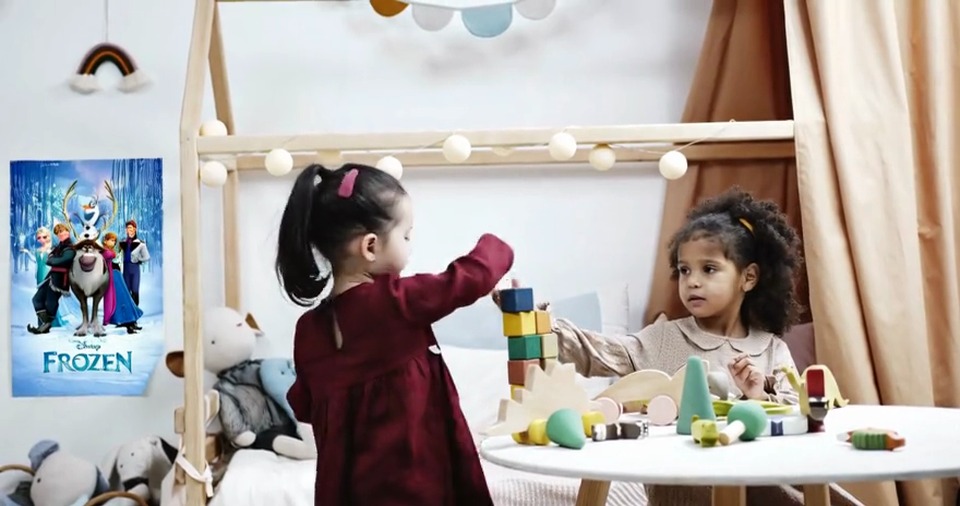}
  {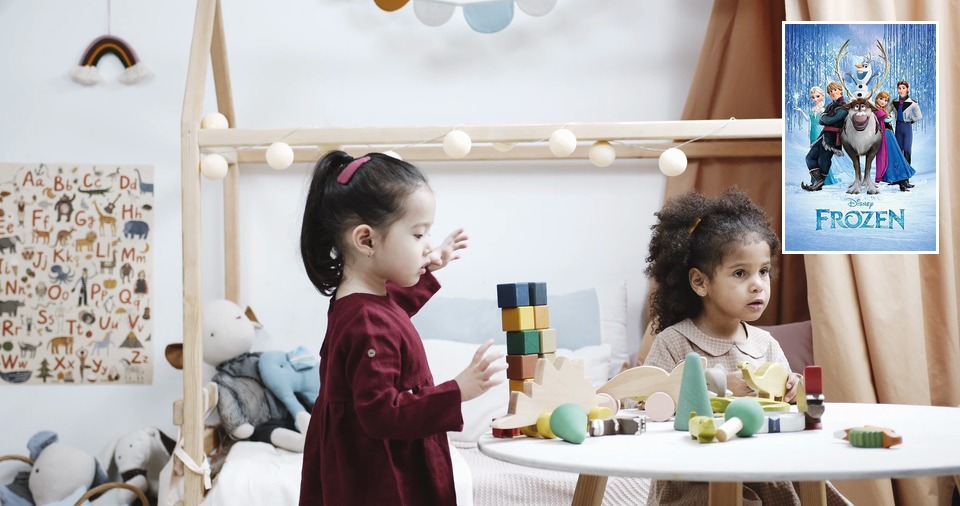}

\AAPromptRow
  {Add a \benchTarget{Kikkoman soy sauce bottle} on the white countertop next to the wooden cutting board.}
  {Replace the animal alphabet poster on the wall with a \benchTarget{Disney Frozen movie poster}.}

\endgroup

%% file: supp/figs/agent_ablation_kiwi.tex
\begingroup
\setlength{\tabcolsep}{0pt}
\renewcommand{\arraystretch}{1.0}

\providecommand{\AAStrip}[1]{\filmstrip[\linewidth]{#1}}
\providecommand{\AAHead}[1]{{\fontsize{6.6}{7.6}\selectfont\sffamily\centering #1\par}}
\providecommand{\AAPrompt}[1]{{\fontsize{7.4}{8.6}\selectfont\itshape\raggedright #1\par}}

\providecommand{\AAHeaderRow}{%
  \noindent\begin{tabular}{@{}p{0.155\textwidth}@{\hspace{0.010\textwidth}}p{0.155\textwidth}@{\hspace{0.010\textwidth}}p{0.155\textwidth}@{\hspace{0.020\textwidth}}p{0.155\textwidth}@{\hspace{0.010\textwidth}}p{0.155\textwidth}@{\hspace{0.010\textwidth}}p{0.155\textwidth}@{}}
    \AAHead{w/o agent} & \AAHead{w/ agent} & \AAHead{Source\,/\,Reference}
  & \AAHead{w/o agent} & \AAHead{w/ agent} & \AAHead{Source\,/\,Reference} \\
  \end{tabular}\par
  \vspace{0.30ex}
}

\providecommand{\AARow}[6]{%
  \noindent\begin{tabular}{@{}p{0.155\textwidth}@{\hspace{0.010\textwidth}}p{0.155\textwidth}@{\hspace{0.010\textwidth}}p{0.155\textwidth}@{\hspace{0.020\textwidth}}p{0.155\textwidth}@{\hspace{0.010\textwidth}}p{0.155\textwidth}@{\hspace{0.010\textwidth}}p{0.155\textwidth}@{}}
    \AAStrip{#1} & \AAStrip{#2} & \AAStrip{#3} & \AAStrip{#4} & \AAStrip{#5} & \AAStrip{#6} \\
  \end{tabular}\par\vspace{-0.2ex}
}

\providecommand{\AAPromptRow}[2]{%
  \noindent\begin{tabular}{@{}p{0.485\textwidth}@{\hspace{0.020\textwidth}}p{0.485\textwidth}@{}}
    \AAPrompt{#1} & \AAPrompt{#2} \\
  \end{tabular}\par\vspace{0.55ex}
}

\AAHeaderRow

\AARow
  {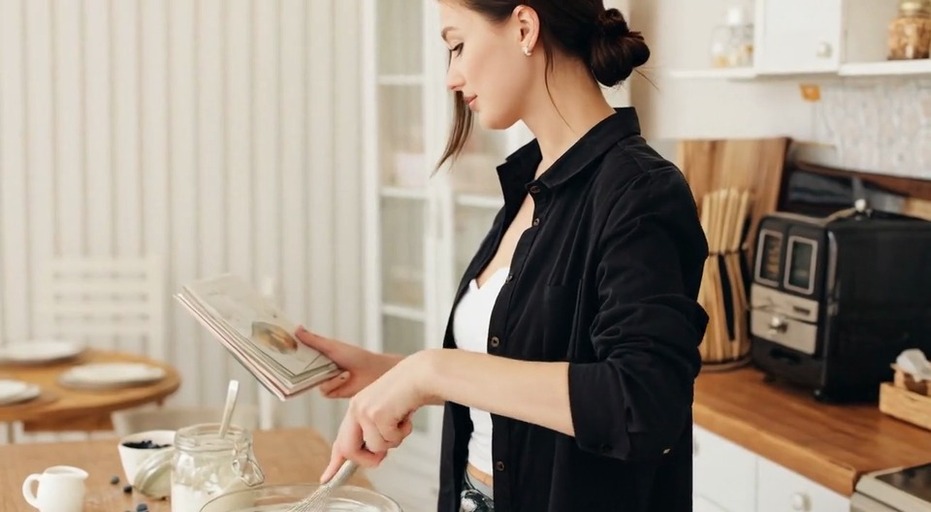}
  {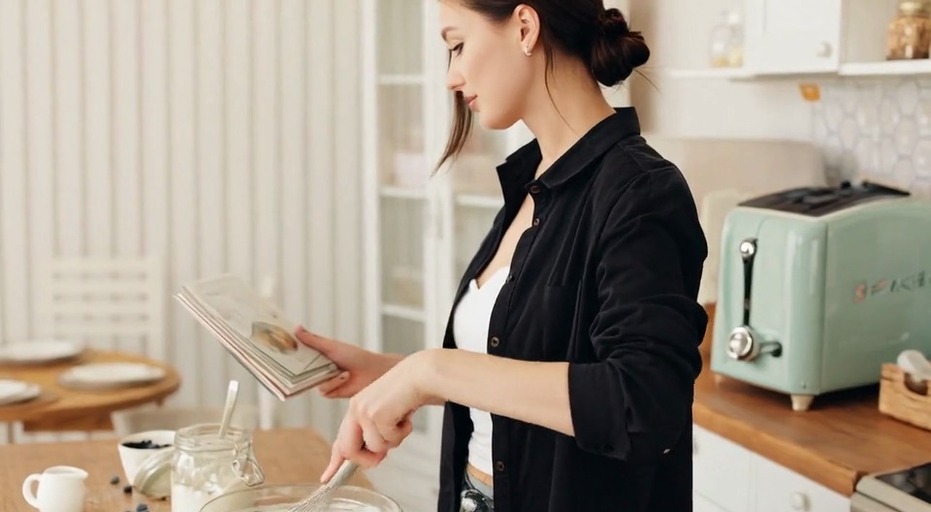}
  {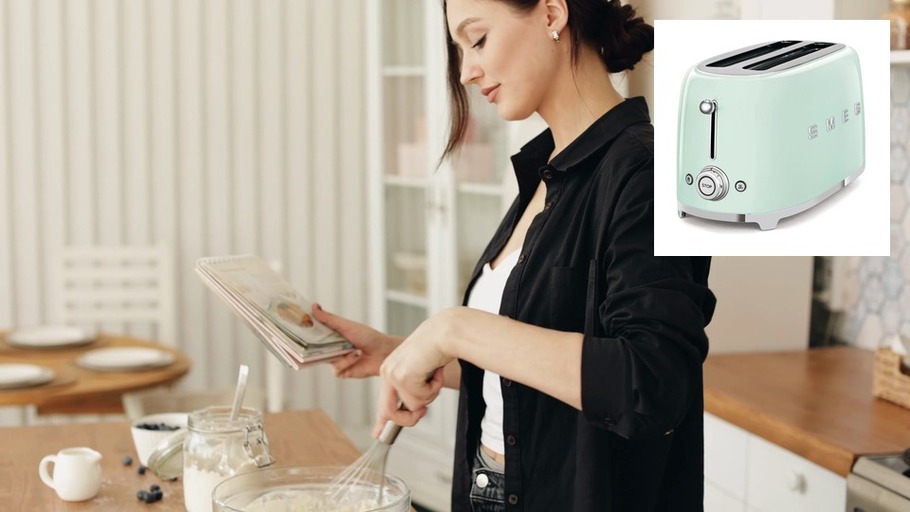}
  {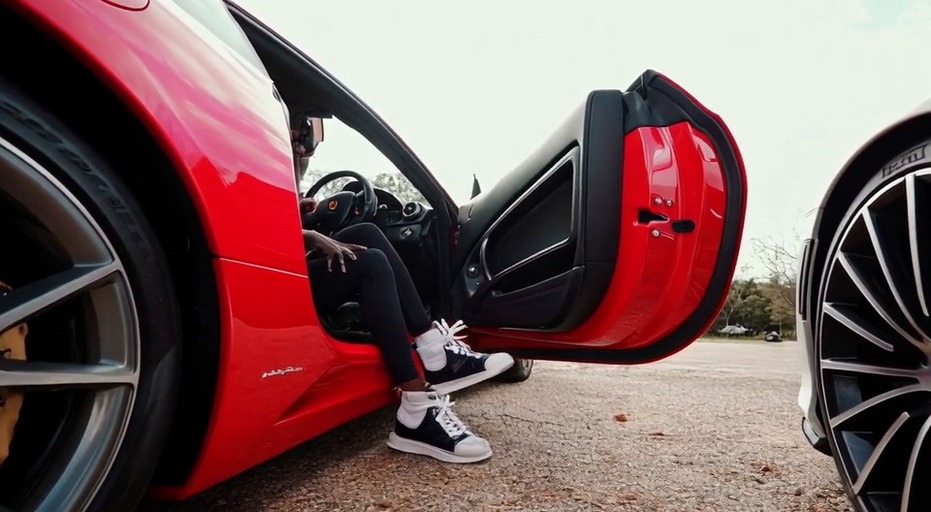}
  {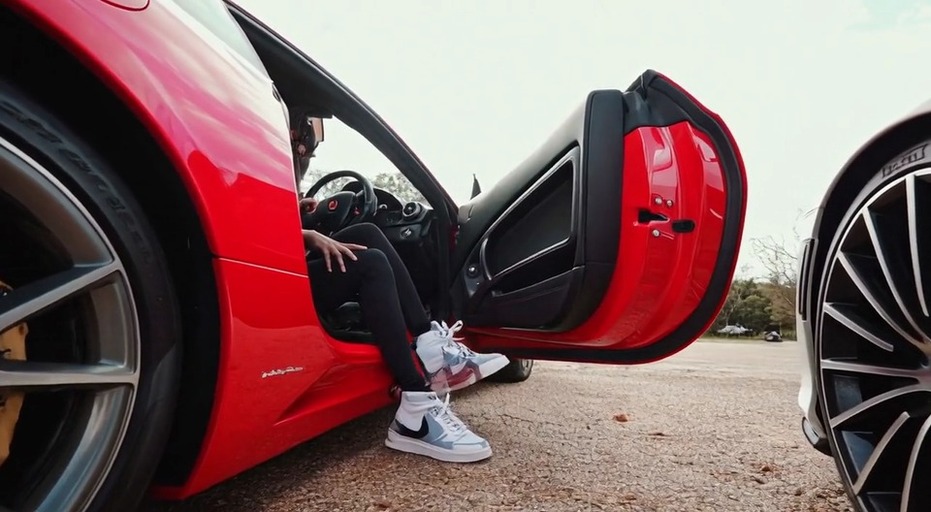}
  {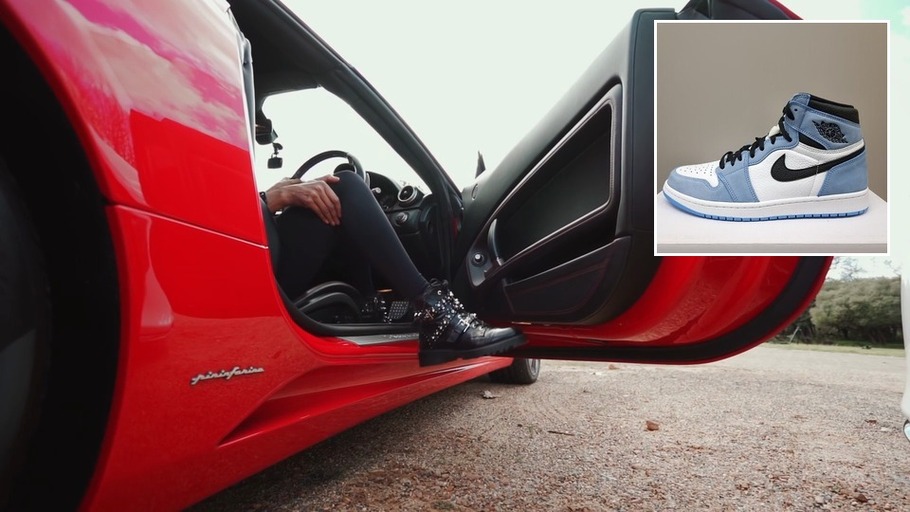}

\AAPromptRow
  {Add a \benchTarget{Smeg retro toaster} on the empty wooden countertop to the right of the woman.}
  {Replace the studded boots on the person's feet with \benchTarget{Nike Air Jordan 1 sneakers}, matching the lighting and movement of the foot.}

\AARow
  {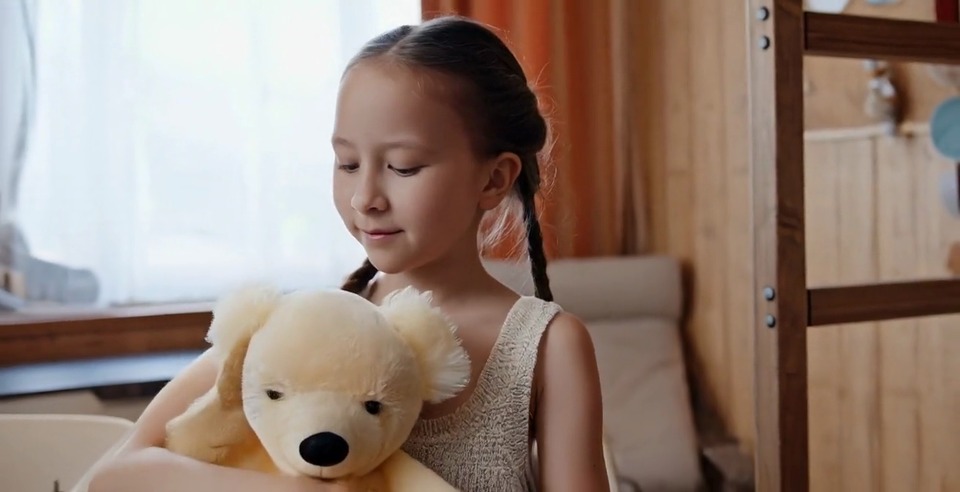}
  {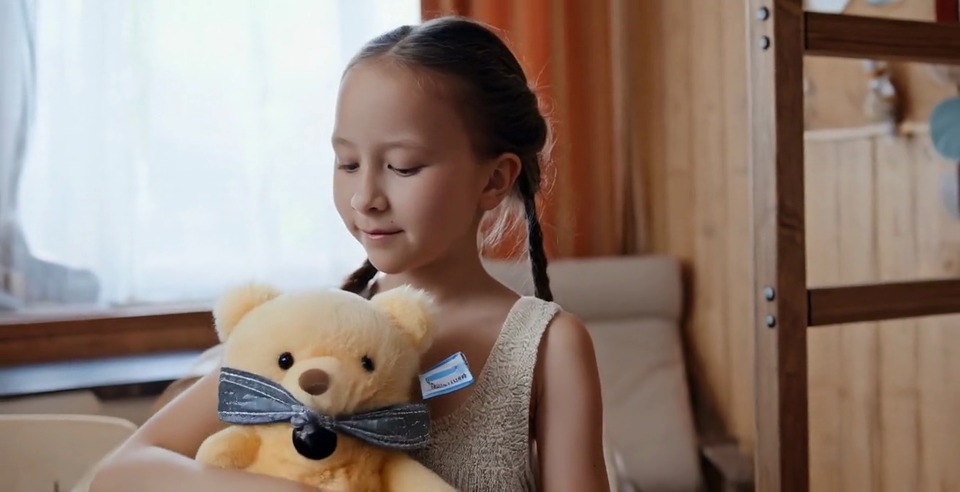}
  {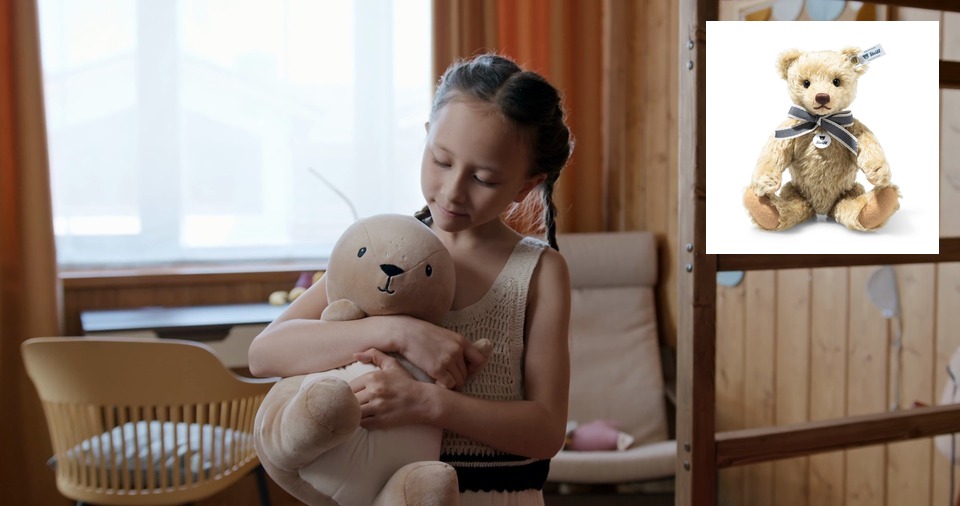}
  {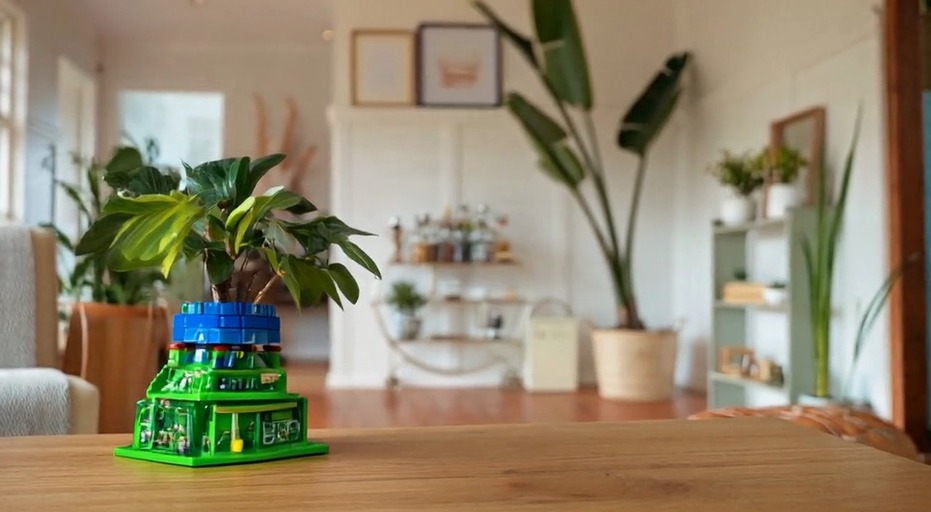}
  {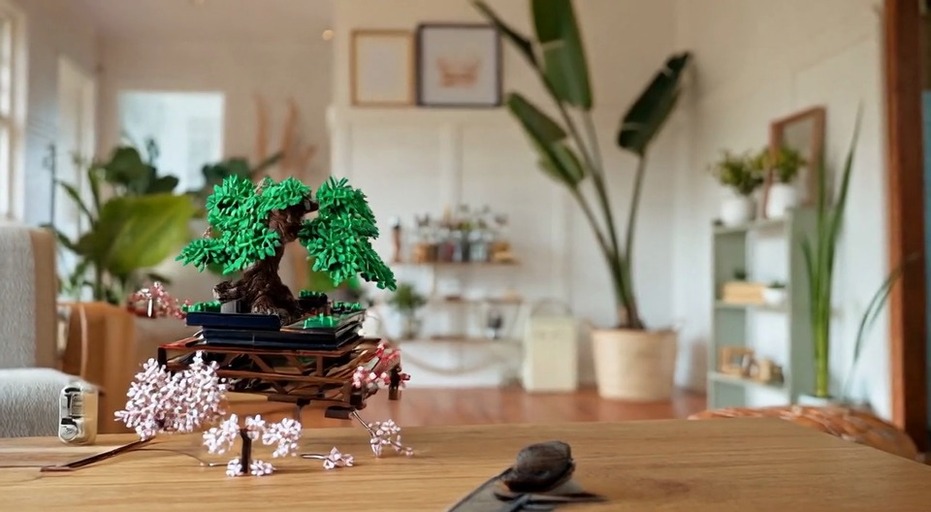}
  {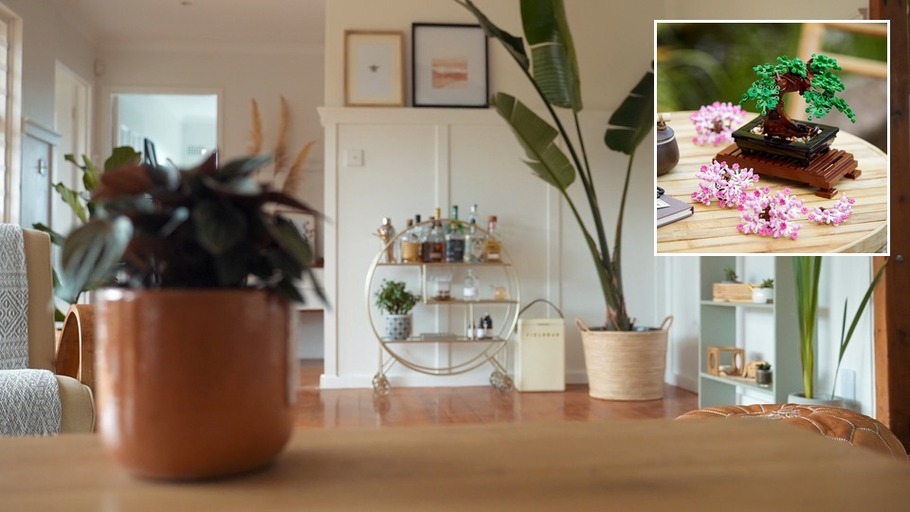}

\AAPromptRow
  {Replace the stuffed animal toy held by the girl with a \benchTarget{Steiff classic teddy bear}.}
  {Replace the potted plant on the table with a \benchTarget{Lego Bonsai Tree} while preserving the shallow depth of field.}

\endgroup

%% file: supp/tables/video_model_and_conditioning.tex
\begin{table*}[!t]
  \caption{\method{} video DiT architecture hyperparameters.}
  \label{tab:appendix:video_model_and_conditioning}
  \centering
  \footnotesize
  \setlength{\tabcolsep}{4pt}
  \renewcommand{\arraystretch}{1.08}
  \begin{tabular}{L{0.18\textwidth}L{0.22\textwidth}L{0.53\textwidth}}
    \toprule
    \rowcolor{benchHeader}
    \benchgroup{Component} & \benchgroup{Item} & \benchgroup{Setting} \\
    \midrule
    Video DiT & Backbone & Wan2.2-TI2V-5B with patch size $(1,2,2)$. \\
    VAE & Model & Wan2.2 VAE, frozen, with 16$\times$ spatial downsampling. \\
    MLLM encoder & Model & Frozen Qwen3.5-4B in bf16. \\
    Context projector & Architecture & Zero-initialized linear layer. \\
    Source conditioning & Latent path & VAE-encoded source video tokens are concatenated with noisy video tokens along the token-sequence dimension. \\
    Source conditioning & MLLM path & Images are resized with maximum pixels $147456$ before feeding into Qwen. Frames are sampled at 1 fps with at least 2 frames. \\
    \bottomrule
  \end{tabular}
\end{table*}

%% file: supp/tables/video_training_details.tex
\begin{table*}[!t]
  \caption{\method{} video DiT training parameters. Stage~1 is the
  low-resolution warmup stage; Stage~2 resumes from the Stage~1
  checkpoint stream at the higher pixel budget.}
  \label{tab:appendix:video_training_details}
  \centering
  \footnotesize
  \setlength{\tabcolsep}{4pt}
  \renewcommand{\arraystretch}{1.08}
  \begin{tabular}{L{0.25\textwidth}L{0.14\textwidth}L{0.14\textwidth}L{0.40\textwidth}}
    \toprule
    \rowcolor{benchHeader}
    \benchgroup{Parameter} & \benchgroup{Stage~1} & \benchgroup{Stage~2} & \benchgroup{Notes} \\
    \midrule
    Frames & 81 & 81 & Wan's 4k+1 temporal constraint. \\
    Video pixel budget & 399{,}360 & 921{,}600 & Stage~1 is about $448\times832$; Stage~2 is about $708\times1280$ \\
    Reference VAE budget & 921{,}600 & 921{,}600 & Per reference image and source video frame. \\
    MLLM source budget & 147{,}456 & 147{,}456 & Per sampled source video frame. \\
    MLLM reference budget & 147{,}456 & 147{,}456 & Per reference image. \\
    MLLM video sampling & 1 fps, min 2 frames & 1 fps, min 2 frames &  Qwen video-processor override. \\
    Optimizer & AdamW & AdamW & - \\
    Learning rate and schedule & $1\times10^{-5}$, constant & $1\times10^{-5}$, constant & - \\
    Weight decay and grad clip & 0.01 / 1.0 & 0.01 / 1.0 & - \\
    Precision and distributed recipe & bf16, ZeRO-2 & bf16, ZeRO-2 & - \\
    Effective global batch & 128 & 128 & - \\
    Source conditioning mode & token-sequence concat & token-sequence concat & - \\
    Prompt dropout & 0.1 & 0.1 & Trains the empty-text branch. \\
    Visual dropout given prompt drop & 0.5 & 0.5 & Produces the visual-negative and unconditional subsets. \\
    \bottomrule
  \end{tabular}
\end{table*}

%% file: supp/tables/inference_settings.tex
\begin{table*}[!t]
  \caption{\method{} per-benchmark inference settings. \texttt{cfg\_scale}
  and \texttt{image\_cfg\_scale} correspond to
  $\lambda_{\mathrm{txt}}$ and $\lambda_{\mathrm{img}}$ in
  Eq.~\eqref{eq:three_pass_cfg}; when
  $\lambda_{\mathrm{img}}=1.0$ the pipeline falls back to the two-pass
  CFG path defined in
  Appendix~\ref{sec:appendix:inference_cfg}. The agent VLM tools
  (web image search and mask overlay) are disabled on the two
  already-specified benchmarks, so the agent only rewrites the
  instruction there.}
  \label{tab:appendix:inference_settings}
  \centering
  \footnotesize
  \setlength{\tabcolsep}{4pt}
  \renewcommand{\arraystretch}{1.08}
  \begin{tabular}{L{0.30\textwidth}L{0.20\textwidth}L{0.20\textwidth}L{0.22\textwidth}}
    \toprule
    \rowcolor{benchHeader}
    \benchgroup{Setting} & \benchgroup{\bench{}} & \benchgroup{EditVerse-Bench} & \benchgroup{OpenVE-Bench} \\
    \midrule
    Checkpoint & Stage~2 & Stage~2 & Stage~2 \\
    Cases & 150 (5 edit types) & 120 (12 edit types) & 358 (6 edit types) \\
    \midrule
    \texttt{cfg\_scale} $(\lambda_{\mathrm{txt}})$ & 2.0 & 1.5 & 2.0 \\
    \texttt{image\_cfg\_scale} $(\lambda_{\mathrm{img}})$ & 1.25 & 1.0 & 1.0 \\
    Two-pass CFG fallback when $\lambda_{\mathrm{img}}{=}1$ & disabled & enabled & enabled \\
    Denoising steps & 50 & 50 & 50 \\
    Output frames & 81 (save 81) & 81 (save 64) & 81, temporal-resize \\
    Max video pixels & 921{,}600 & 921{,}600 & 921{,}600 \\
    Negative prompt & empty & empty & empty \\
    \midrule
    VLM agent base & \multicolumn{3}{l}{Qwen3-VL-8B-Instruct~\citep{qwen3vl} + Aurora LoRA adapter} \\
    Agent rewrite ($y'$) used & yes & yes & yes \\
    Agent web image search & enabled & disabled & disabled \\
    Agent mask overlay & enabled & disabled & disabled \\
    \midrule
    Judge model & \multicolumn{3}{l}{Gemini~2.5~Pro~\citep{gemini25pro} at \texttt{temperature}=0} \\
    \bottomrule
  \end{tabular}
\end{table*}

%% file: supp/figs/aurora_sft.tex
\definecolor{sftInk}{HTML}{161B22}
\definecolor{sftMuted}{HTML}{59636E}
\definecolor{sftTagVideo}{HTML}{1F4F8A}
\definecolor{sftTagRef}{HTML}{A65A12}

\providecommand{\SFVideoStrip}[1]{%
  \begin{minipage}{\linewidth}\centering%
    \includegraphics[width=\linewidth,keepaspectratio]{#1}%
  \end{minipage}%
}
\providecommand{\SFTagVideo}{\par\noindent{\fontsize{4.7}{5.3}\selectfont\sffamily\hfill\textcolor{sftTagVideo}{\MakeUppercase{Source video frames}}\hfill\null\par}}

\providecommand{\SFPrompt}[1]{{\fontsize{5.05}{5.55}\selectfont\itshape\textcolor{sftInk}{#1}\par}}
\providecommand{\SFJson}[1]{{\fontsize{4.75}{5.20}\selectfont\textcolor{sftInk}{#1}\par}}
\providecommand{\SFField}[2]{\SFJson{\hspace*{0.55em}\textcolor{sftMuted}{#1}: #2}}
\providecommand{\SFCase}[1]{%
  \begin{minipage}[t]{\linewidth}%
    \raggedright
    \setlength{\parskip}{0pt}%
    \setlength{\baselineskip}{5.40pt}%
    #1%
  \end{minipage}%
}
\providecommand{\SFRow}[3]{%
  \noindent\begin{tabular}{@{}p{0.318\textwidth}@{\hspace{0.023\textwidth}}p{0.318\textwidth}@{\hspace{0.023\textwidth}}p{0.318\textwidth}@{}}
  #1 & #2 & #3
  \end{tabular}\par
  \vspace{1.10ex}
}

\begin{figure}[t]
  \centering
  \begingroup
  \setlength{\tabcolsep}{0pt}%
  \renewcommand{\arraystretch}{1.0}%
  \SFRow
  {\SFCase{%
    \SFVideoStrip{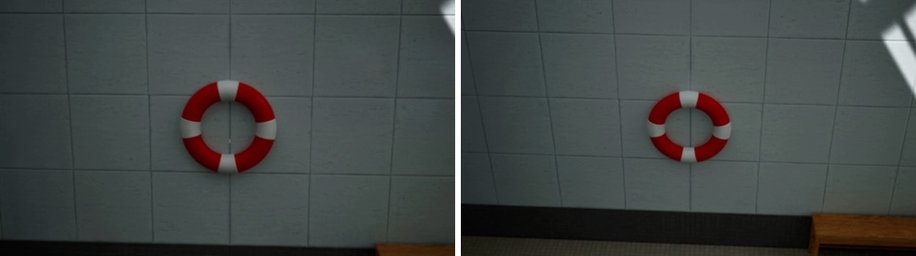}\\[0.10ex]
    \SFTagVideo
    \SFPrompt{User prompt: ``hey, i was just looking at this and wondering if you could maybe take that lifebuoy thing off the wall so it looks plain?''}
    \SFJson{Output plan JSON: \{}
    \SFField{refined\_text\_instruction}{``Remove the red and white striped lifebuoy from the center of the tiled wall. Replace the space it occupied with the continuous pattern of the light gray square tiles\dots''}
    \SFField{subtask}{remove\_object}
    \SFField{image\_search}{false}
    \SFField{mask}{``red and white striped lifebuoy from the center of the tiled wall''}
    \SFJson{\}}
  }}
  {\SFCase{%
    \SFVideoStrip{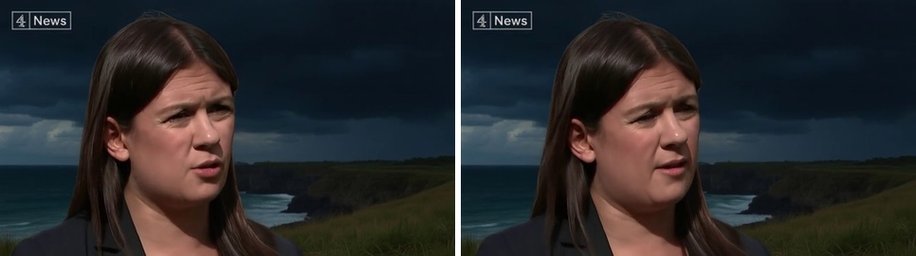}\\[0.10ex]
    \SFTagVideo
    \SFPrompt{User prompt: ``hey could you maybe change the background to 10 downing street and make it look like a still daylight shot instead of the stormy cliff?''}
    \SFJson{Output plan JSON: \{}
    \SFField{refined\_text\_instruction}{``Replace the dynamic stormy coastal cliff with a static outdoor scene of 10 Downing Street under consistent daylight with no background motion.''}
    \SFField{subtask}{change\_background}
    \SFField{image\_search}{``10 Downing Street''}
    \SFField{mask}{false}
    \SFJson{\}}
  }}
  {\SFCase{%
    \SFVideoStrip{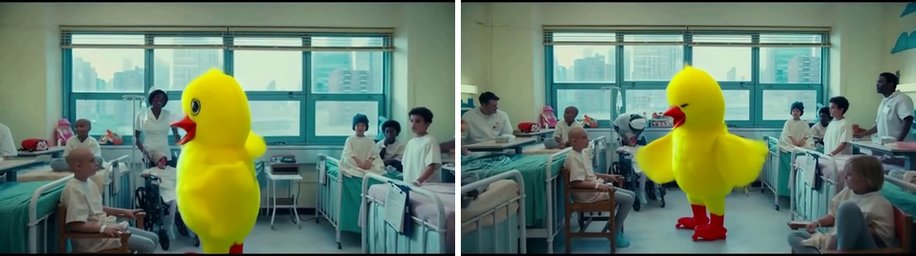}\\[0.10ex]
    \SFTagVideo
    \SFPrompt{User prompt: ``swap duck joker''}
    \SFJson{Output plan JSON: \{}
    \SFField{refined\_text\_instruction}{``Replace the yellow duck mascot on the right side of the frame with a Joker in a white lab coat, plaid shirt, and orange vest.''}
    \SFField{subtask}{replace\_object}
    \SFField{image\_search}{``Joker''}
    \SFField{mask}{false}
    \SFJson{\}}
  }}
  \SFRow
  {\SFCase{%
    \SFVideoStrip{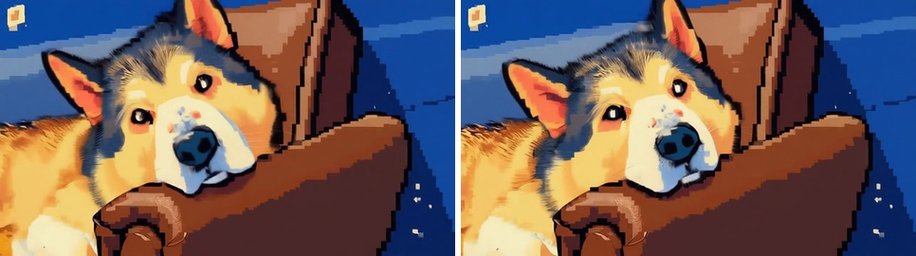}\\[0.10ex]
    \SFTagVideo
    \SFPrompt{User prompt: ``can u make this look real and not like pixel art anymore thx''}
    \SFJson{Output plan JSON: \{}
    \SFField{refined\_text\_instruction}{``Convert this pixel art video back to a live-action, photorealistic format by systematically removing all pixel art characteristics. Restore real-world textures, natural lighting\dots''}
    \SFField{subtask}{global\_style}
    \SFField{image\_search}{false}
    \SFField{mask}{false}
    \SFJson{\}}
  }}
  {\SFCase{%
    \SFVideoStrip{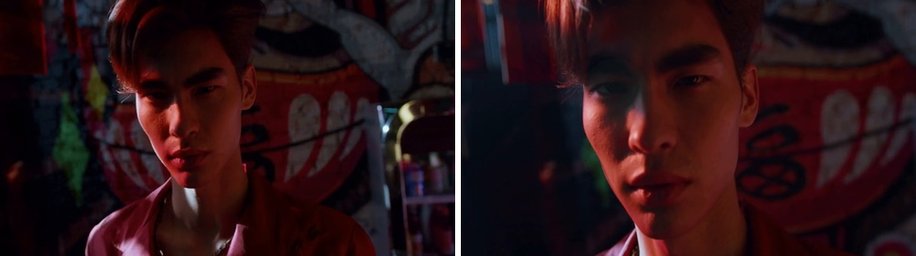}\\[0.10ex]
    \SFTagVideo
    \SFPrompt{User prompt: ``make music video add flares change colors''}
    \SFJson{Output plan JSON: \{}
    \SFField{refined\_text\_instruction}{``Transform the entire video into a cinematic music video aesthetic with slow-motion effects, lens flares, and a color grade that blends deep reds and electric blues into a moody\dots''}
    \SFField{subtask}{combined\_tasks}
    \SFField{image\_search}{false}
    \SFField{mask}{false}
    \SFJson{\}}
  }}
  {\SFCase{%
    \SFVideoStrip{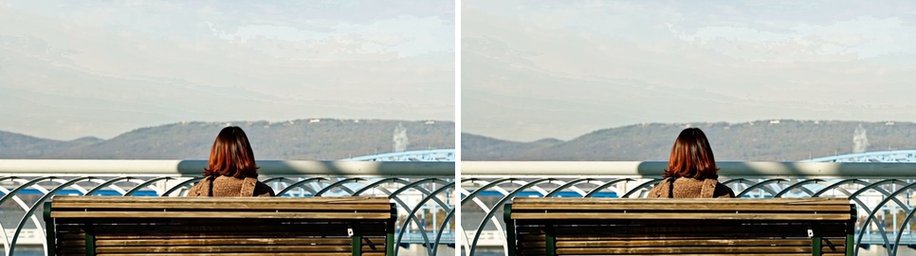}\\[0.10ex]
    \SFTagVideo
    \SFPrompt{User prompt: ``Change the hair to long and light colored with that accessory on the end.''}
    \SFJson{Output plan JSON: \{}
    \SFField{refined\_text\_instruction}{``Change the shoulder-length brown hair into long, silvery-white hair with a single red ribbon tied at the end.''}
    \SFField{subtask}{change\_color}
    \SFField{image\_search}{false}
    \SFField{mask}{false}
    \SFJson{\}}
  }}
  \SFRow
  {\SFCase{%
    \SFVideoStrip{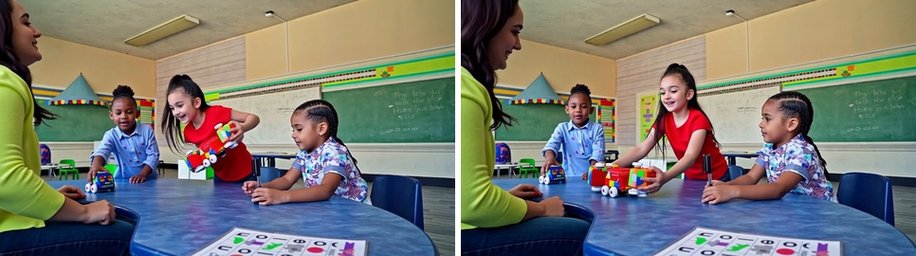}\\[0.10ex]
    \SFTagVideo
    \SFPrompt{User prompt: ``hey i was just thinking it would be super cool if you could maybe make those toy blocks look like thomas the tank engine.''}
    \SFJson{Output plan JSON: \{}
    \SFField{refined\_text\_instruction}{``Transform the colorful toy blocks into a Thomas the Tank Engine toy.''}
    \SFField{subtask}{replace\_object}
    \SFField{image\_search}{``Thomas the Tank Engine toy''}
    \SFField{mask}{false}
    \SFJson{\}}
  }}
  {\SFCase{%
    \SFVideoStrip{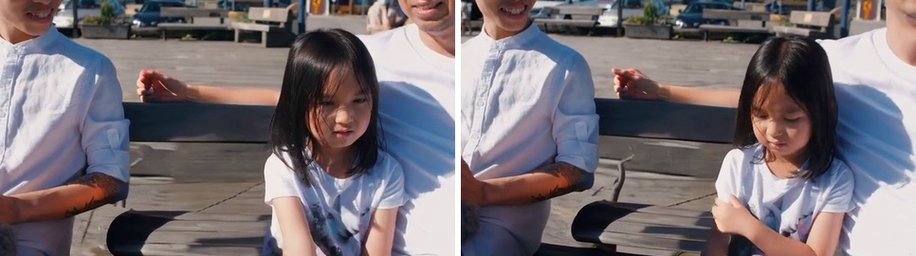}\\[0.10ex]
    \SFTagVideo
    \SFPrompt{User prompt: ``Put a kid in that shirt on the bench between the two people.''}
    \SFJson{Output plan JSON: \{}
    \SFField{refined\_text\_instruction}{``Add a young boy with dark hair, wearing a white t-shirt with two black stripes on the sleeves, looking down, sitting on the bench between the adult and the young girl.''}
    \SFField{subtask}{add\_object}
    \SFField{image\_search}{false}
    \SFField{mask}{false}
    \SFJson{\}}
  }}
  {\SFCase{%
    \SFVideoStrip{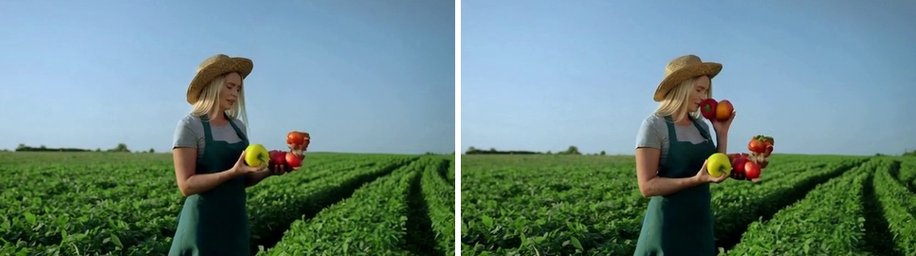}\\[0.10ex]
    \SFTagVideo
    \SFPrompt{User prompt: ``put dwight schrute in a blue shirt and green apron holding red stuff standin next to the lady lookin down pls''}
    \SFJson{Output plan JSON: \{}
    \SFField{refined\_text\_instruction}{``Add Dwight Schrute wearing a blue shirt and a green apron, holding red produce, looking down, standing to the right of the woman in the field.''}
    \SFField{subtask}{add\_object}
    \SFField{image\_search}{``Dwight Schrute''}
    \SFField{mask}{false}
    \SFJson{\}}
  }}
  \SFRow
  {\SFCase{%
    \SFVideoStrip{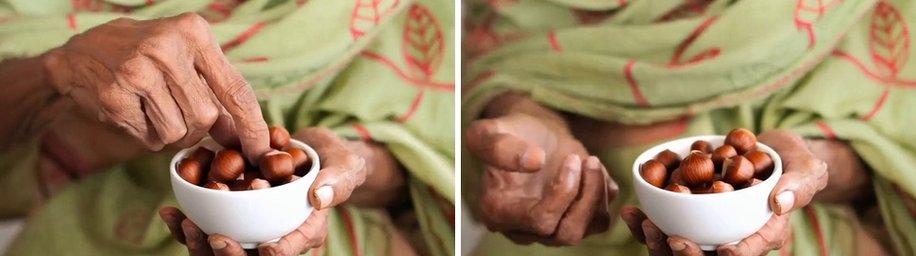}\\[0.10ex]
    \SFTagVideo
    \SFPrompt{User prompt: ``swap bowl for nutella jar''}
    \SFJson{Output plan JSON: \{}
    \SFField{refined\_text\_instruction}{``Replace the white bowl filled with hazelnuts held by hands in the center of the frame with a jar of Nutella.''}
    \SFField{subtask}{replace\_object}
    \SFField{image\_search}{``jar of Nutella''}
    \SFField{mask}{false}
    \SFJson{\}}
  }}
  {\SFCase{%
    \SFVideoStrip{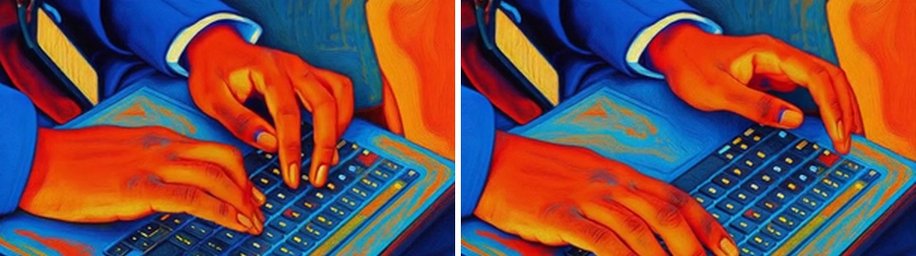}\\[0.10ex]
    \SFTagVideo
    \SFPrompt{User prompt: ``make photorealistic''}
    \SFJson{Output plan JSON: \{}
    \SFField{refined\_text\_instruction}{``Convert this Abstract Art-animated video to a live-action, photorealistic format. Meticulously strip away the abstract shapes, stylized color fields, and gestural animations\dots''}
    \SFField{subtask}{global\_style}
    \SFField{image\_search}{false}
    \SFField{mask}{false}
    \SFJson{\}}
  }}
  {\SFCase{%
    \SFVideoStrip{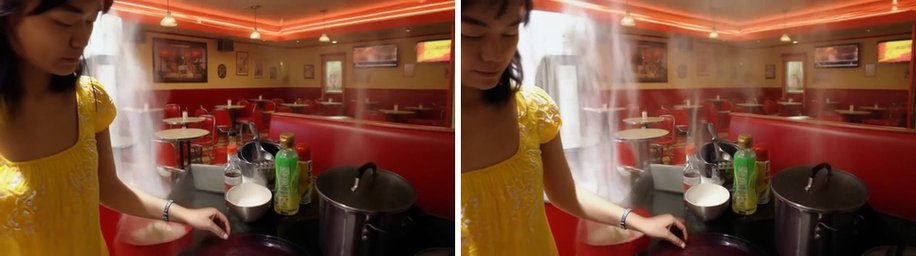}\\[0.10ex]
    \SFTagVideo
    \SFPrompt{User prompt: ``swich the diner background to monica gellers kitchen from friends w the purple walls and teal cabinets pls thx''}
    \SFJson{Output plan JSON: \{}
    \SFField{refined\_text\_instruction}{``Replace the dynamic 1950s diner background with Monica Geller's kitchen from the TV show Friends, making sure to include its iconic purple walls and teal cabinets.''}
    \SFField{subtask}{change\_background}
    \SFField{image\_search}{``Monica Geller's kitchen from Friends''}
    \SFField{mask}{false}
    \SFJson{\}}
  }}
  \SFRow
  {\SFCase{%
    \SFVideoStrip{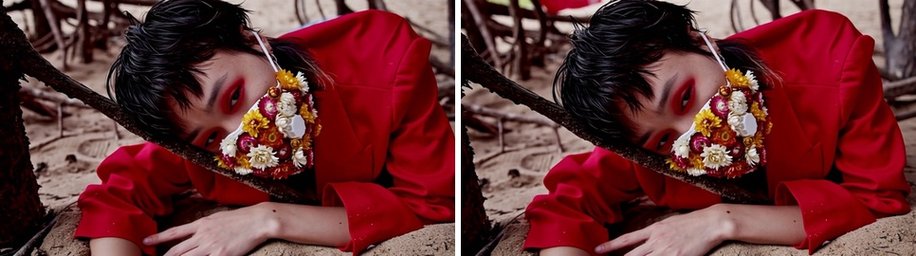}\\[0.10ex]
    \SFTagVideo
    \SFPrompt{User prompt: ``change mask pattern''}
    \SFJson{Output plan JSON: \{}
    \SFField{refined\_text\_instruction}{``Transform the floral face mask into a geometric patterned mask in gold and black.''}
    \SFField{subtask}{change\_color}
    \SFField{image\_search}{false}
    \SFField{mask}{false}
    \SFJson{\}}
  }}
  {\SFCase{%
    \SFVideoStrip{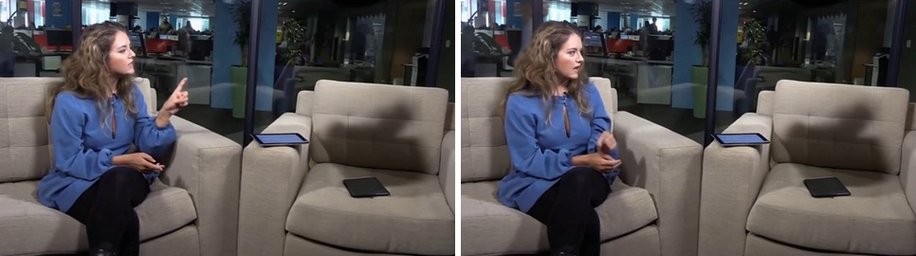}\\[0.10ex]
    \SFTagVideo
    \SFPrompt{User prompt: ``put darth vader on the right chair thx''}
    \SFJson{Output plan JSON: \{}
    \SFField{refined\_text\_instruction}{``Add Darth Vader sitting on the rightmost chair.''}
    \SFField{subtask}{add\_object}
    \SFField{image\_search}{``Darth Vader''}
    \SFField{mask}{false}
    \SFJson{\}}
  }}
  {\SFCase{%
    \SFVideoStrip{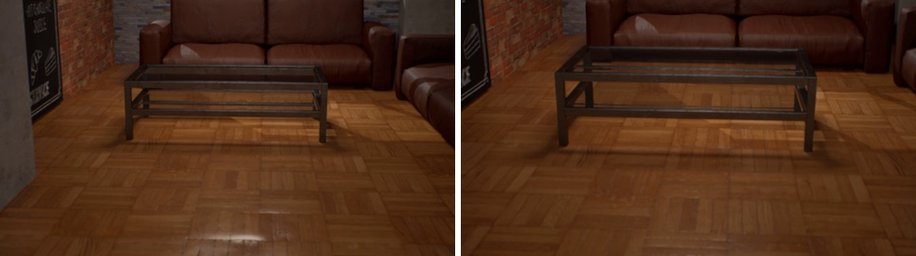}\\[0.10ex]
    \SFTagVideo
    \SFPrompt{User prompt: ``hey, the room feels a bit cramped so could you maybe take out that table and just fill in the floor so it looks empty?''}
    \SFJson{Output plan JSON: \{}
    \SFField{refined\_text\_instruction}{``Remove the dark brown rectangular wooden table frame located in the center of the room. Replace it by continuing the pattern of the polished parquet flooring\dots''}
    \SFField{subtask}{remove\_object}
    \SFField{image\_search}{false}
    \SFField{mask}{``dark brown rectangular wooden table frame located in the center of the room''}
    \SFJson{\}}
  }}
  \endgroup
  \caption{Planning SFT examples.}
  \label{fig:aurora_sft}
\end{figure}

%% file: supp/figs/aurora_select.tex
\definecolor{selInk}{HTML}{161B22}
\definecolor{selMuted}{HTML}{59636E}
\definecolor{selGood}{HTML}{1F7A4D}

\providecommand{\SelStrip}[1]{%
  \begin{minipage}{\linewidth}\centering%
    \includegraphics[width=\linewidth,keepaspectratio]{#1}%
  \end{minipage}%
}
\providecommand{\SelPrompt}[1]{{\fontsize{5.20}{5.70}\selectfont\itshape\textcolor{selInk}{#1}\par}}
\providecommand{\SelText}[1]{{\fontsize{5.05}{5.50}\selectfont\textcolor{selInk}{#1}\par}}
\providecommand{\SelGood}[1]{\textcolor{selGood}{#1}}
\providecommand{\SelCase}[1]{%
  \begin{minipage}[t]{\linewidth}%
    \raggedright
    \setlength{\parskip}{0pt}%
    \setlength{\baselineskip}{5.50pt}%
    #1%
  \end{minipage}%
}
\providecommand{\SelRow}[3]{%
  \noindent\begin{tabular}{@{}p{0.318\textwidth}@{\hspace{0.023\textwidth}}p{0.318\textwidth}@{\hspace{0.023\textwidth}}p{0.318\textwidth}@{}}
  #1 & #2 & #3
  \end{tabular}\par
  \vspace{1.40ex}
}

\begin{figure}[t]
  \centering
  \begingroup
  \setlength{\tabcolsep}{0pt}%
  \renewcommand{\arraystretch}{1.0}%
  \SelRow
  {\SelCase{%
    \SelPrompt{Search prompt: ``A macro photo of the orchid mantis (Hymenopus coronatus) on a flower. A nearby informational card lists the scientific family name and the specific country where the holotype was collected; both must be correct.''}
    \SelStrip{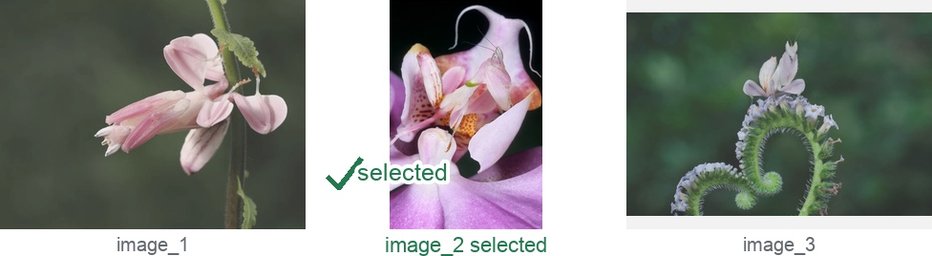}\\[0.10ex]
    \SelText{\SelGood{Selected}: image\_2.}
  }}
  {\SelCase{%
    \SelPrompt{Search prompt: ``A detailed 3D visualization of the metal-organic framework MIL-53. A data sidebar on the screen lists its specific BET surface area in square meters per gram and the city of the university in France where it was first developed; both must be correct.''}
    \SelStrip{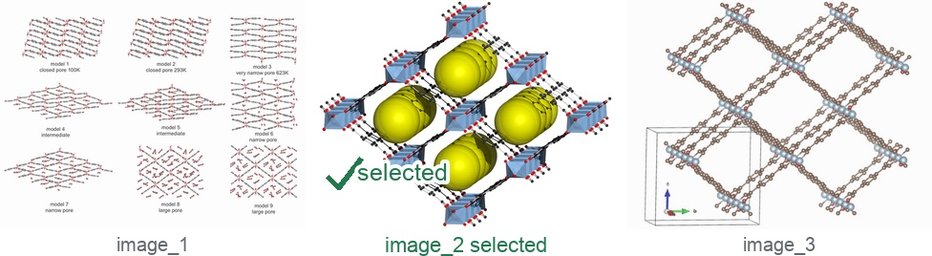}\\[0.10ex]
    \SelText{\SelGood{Selected}: image\_2.}
  }}
  {\SelCase{%
    \SelPrompt{Search prompt: ``A macro photograph of a `Chuvash Khushpu' headpiece resting on a pedestal. A small archival tag identifies the village where this specific piece was acquired and lists the precise number of silver coins sewn onto it.''}
    \SelStrip{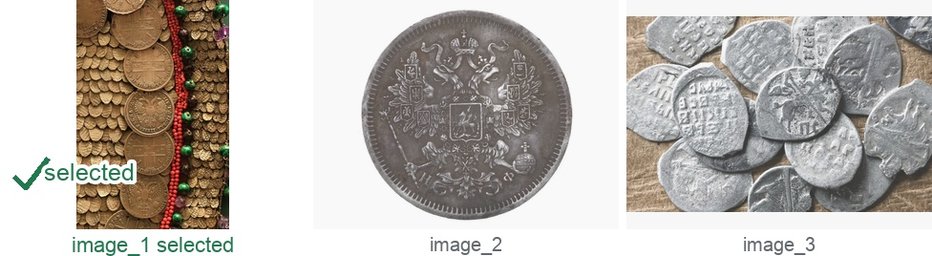}\\[0.10ex]
    \SelText{\SelGood{Selected}: image\_1.}
  }}
  \SelRow
  {\SelCase{%
    \SelPrompt{Search prompt: ``A photo of an exhibit at the Nobel Peace Center focusing on the organization `Nihon Hidankyo'. A wall caption identifies the year the organization was founded and the city where its first general assembly was held.''}
    \SelStrip{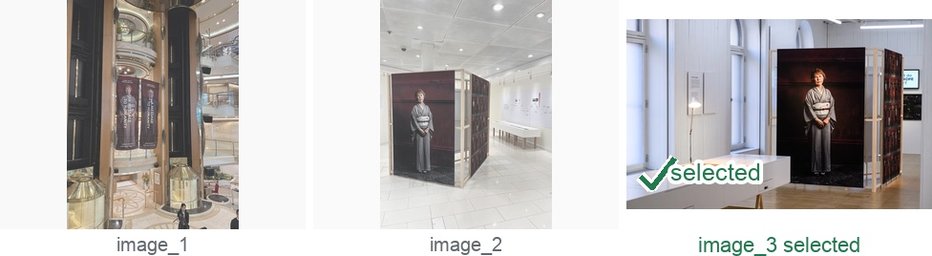}\\[0.10ex]
    \SelText{\SelGood{Selected}: image\_3.}
  }}
  {\SelCase{%
    \SelPrompt{Search prompt: ``A cinematic shot of the `Dungeon Meshi' character `Izutsumi' in her transformed beast-girl form. A side panel lists her archetype and her listed height; both must be correct.''}
    \SelStrip{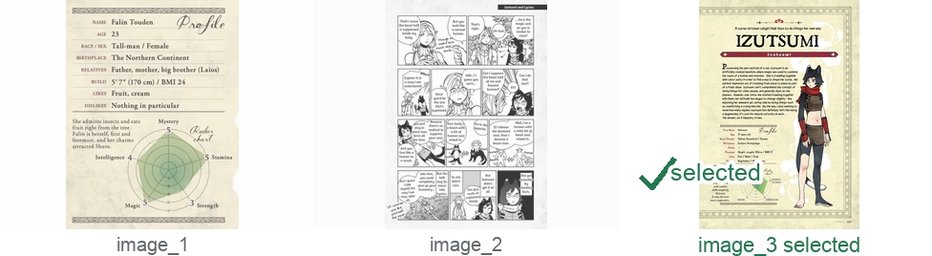}\\[0.10ex]
    \SelText{\SelGood{Selected}: image\_3.}
  }}
  {\SelCase{%
    \SelPrompt{Search prompt: ``A detailed shot of the `Jingdezhen Imperial Kiln Museum' in China, showcasing its brick-vaulted halls. A signage panel lists the architect of record and the year construction was completed.''}
    \SelStrip{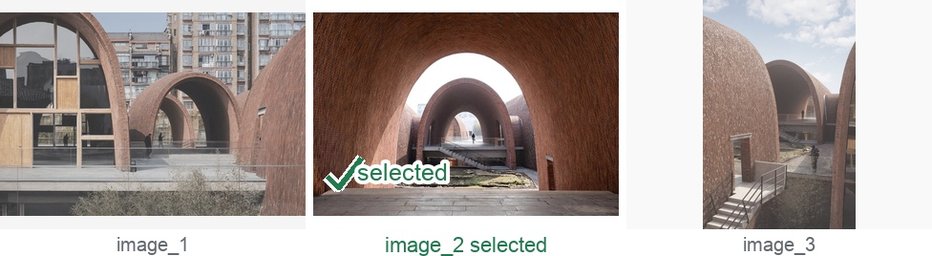}\\[0.10ex]
    \SelText{\SelGood{Selected}: image\_2.}
  }}
  \endgroup
  \caption{Reference-image selection data examples.}
  \label{fig:aurora_select}
\end{figure}

%% file: supp/figs/aurora_dpo.tex
\definecolor{dpoInk}{HTML}{161B22}
\definecolor{dpoMuted}{HTML}{59636E}
\definecolor{dpoGoodEdge}{HTML}{2E7D32}
\definecolor{dpoGoodFill}{HTML}{F1F8E9}
\definecolor{dpoBadEdge}{HTML}{C62828}
\definecolor{dpoBadFill}{HTML}{FDECEA}

\newlength{\DPOGap}\setlength{\DPOGap}{6pt}
\newlength{\DPOColW}\setlength{\DPOColW}{\dimexpr 0.5\textwidth - 0.5\DPOGap\relax}

\providecommand{\DPOPrompt}[1]{{\fontsize{7.60}{8.55}\selectfont\itshape\textcolor{dpoInk}{#1}\par}}
\providecommand{\DPOFrame}[1]{%
  \noindent\includegraphics[width=\linewidth,keepaspectratio]{#1}\par%
}
\providecommand{\DPOJson}[1]{{\fontsize{7.05}{7.95}\selectfont\textcolor{dpoInk}{#1}\par}}
\providecommand{\DPOLine}[2]{\DPOJson{\hspace*{0.45em}\textcolor{dpoMuted}{\texttt{#1}}: #2}}
\providecommand{\DPOHiGood}[1]{\textcolor{dpoGoodEdge}{\textbf{#1}}}
\providecommand{\DPOHiBad}[1]{\textcolor{dpoBadEdge}{\textbf{#1}}}
\providecommand{\DPOBarGood}{\textcolor{dpoGoodEdge}{\rule[-0.05em]{0.22em}{0.85em}}\hspace{0.18em}}
\providecommand{\DPOBarBad}{\textcolor{dpoBadEdge}{\rule[-0.05em]{0.22em}{0.85em}}\hspace{0.18em}}
\providecommand{\DPOLineHiGood}[2]{\DPOJson{\DPOBarGood\textcolor{dpoMuted}{\texttt{#1}}: \DPOHiGood{#2}}}
\providecommand{\DPOLineHiBad}[2]{\DPOJson{\DPOBarBad\textcolor{dpoMuted}{\texttt{#1}}: \DPOHiBad{#2}}}
\providecommand{\DPOTagGood}{{\fontsize{7.30}{8.00}\selectfont\sffamily\textbf{\textcolor{dpoGoodEdge}{\checkmark{} Chosen}}\par}}
\providecommand{\DPOTagBad}{{\fontsize{7.30}{8.00}\selectfont\sffamily\textbf{\textcolor{dpoBadEdge}{\textsf{x}\,Rejected}}\par}}
\providecommand{\DPOPanelGood}[1]{%
  \setlength{\fboxsep}{4pt}\setlength{\fboxrule}{0.5pt}%
  \fcolorbox{dpoGoodEdge}{dpoGoodFill}{%
    \begin{minipage}[t]{\dimexpr\DPOColW-2\fboxsep-2\fboxrule\relax}%
    \DPOTagGood
    \vspace{0.20ex}
    #1%
    \end{minipage}%
  }%
}
\providecommand{\DPOPanelBad}[1]{%
  \setlength{\fboxsep}{4pt}\setlength{\fboxrule}{0.5pt}%
  \fcolorbox{dpoBadEdge}{dpoBadFill}{%
    \begin{minipage}[t]{\dimexpr\DPOColW-2\fboxsep-2\fboxrule\relax}%
    \DPOTagBad
    \vspace{0.20ex}
    #1%
    \end{minipage}%
  }%
}
\providecommand{\DPOPair}[2]{%
  \noindent#1\hspace{\DPOGap}#2\par
}
\providecommand{\DPOCase}[2]{%
  \begin{minipage}[t]{\linewidth}%
    \raggedright
    \setlength{\parskip}{0pt}%
    \setlength{\baselineskip}{7.70pt}%
    #1
    \vspace{0.30ex}
    #2
  \end{minipage}\par
  \vspace{0.80ex}
}

\begin{figure}[t]
  \centering
  \begingroup
  \setlength{\tabcolsep}{0pt}%
  \renewcommand{\arraystretch}{1.0}%
  \begin{minipage}{0.9\linewidth}%
  \setlength{\DPOColW}{\dimexpr 0.5\linewidth - 0.5\DPOGap\relax}%

  \DPOCase
  {\DPOPrompt{User prompt: ``make that place look like an old vintage postcard with sepia tones and blurry edges like an old camera.''}\DPOFrame{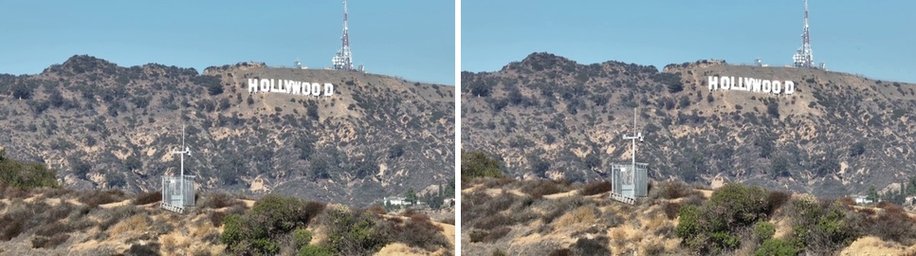}}
  {\DPOPair
    {\DPOPanelGood{%
      \DPOLine{refined\_text\_instruction}{``Transform the video into a vintage Hollywood postcard style with soft sepia tones, film grain, and a slightly blurred edge, as if seen through an old camera lens.''}
      \DPOLineHiGood{subtask}{global\_style}
      \DPOLine{image\_search}{false}
      \DPOLine{mask}{false}
    }}
    {\DPOPanelBad{%
      \DPOLine{refined\_text\_instruction}{``Transform the video into a vintage Hollywood postcard style with soft sepia tones, film grain, and a slightly blurred edge, as if seen through an old camera lens.''}
      \DPOLineHiBad{subtask}{combined\_tasks}
      \DPOLine{image\_search}{false}
      \DPOLine{mask}{false}
    }}
  }

  \DPOCase
  {\DPOPrompt{User prompt: ``hey i was looking at this and i think it would be cool if you could maybe take out the guy sitting on the right.''}\DPOFrame{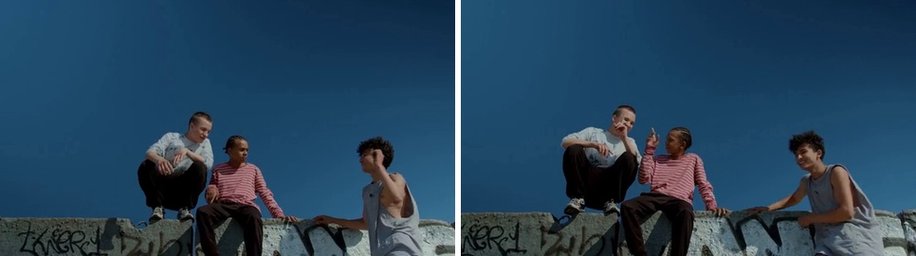}}
  {\DPOPair
    {\DPOPanelGood{%
      \DPOLine{refined\_text\_instruction}{``Remove the person on the right.''}
      \DPOLine{subtask}{remove\_object}
      \DPOLine{image\_search}{false}
      \DPOLineHiGood{mask}{``person on the right''}
    }}
    {\DPOPanelBad{%
      \DPOLine{refined\_text\_instruction}{``Remove the person on the right.''}
      \DPOLine{subtask}{remove\_object}
      \DPOLine{image\_search}{false}
      \DPOLineHiBad{mask}{``person''}
    }}
  }

  \DPOCase
  {\DPOPrompt{User prompt: ``add tomatoes bowl''}\DPOFrame{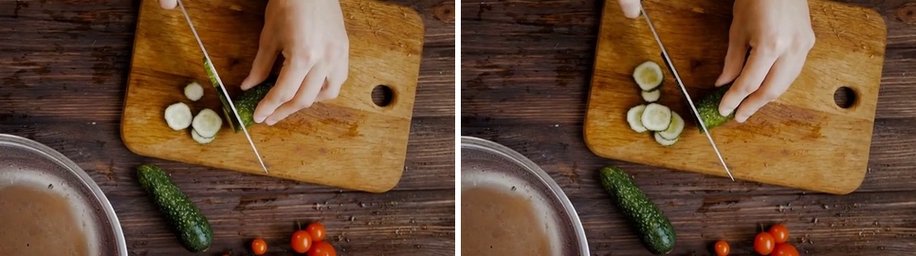}}
  {\DPOPair
    {\DPOPanelGood{%
      \DPOLine{refined\_text\_instruction}{``Add small red cherry tomatoes into the clear bowl.''}
      \DPOLine{subtask}{add\_object}
      \DPOLineHiGood{image\_search}{false}
      \DPOLine{mask}{false}
    }}
    {\DPOPanelBad{%
      \DPOLine{refined\_text\_instruction}{``Add small red cherry tomatoes into the clear bowl.''}
      \DPOLine{subtask}{add\_object}
      \DPOLineHiBad{image\_search}{``NatureSweet Cherubs tomatoes''}
      \DPOLine{mask}{false}
    }}
  }
  \end{minipage}%
  \endgroup
  \caption{DPO preference examples (1/2).}
  \label{fig:aurora_dpo_a}
\end{figure}

\begin{figure}[t]
  \centering
  \begingroup
  \setlength{\tabcolsep}{0pt}%
  \renewcommand{\arraystretch}{1.0}%
  \begin{minipage}{0.9\linewidth}%
  \setlength{\DPOColW}{\dimexpr 0.5\linewidth - 0.5\DPOGap\relax}%

  \DPOCase
  {\DPOPrompt{User prompt: ``change that guy in the center to an old man wearing different clothes.''}\DPOFrame{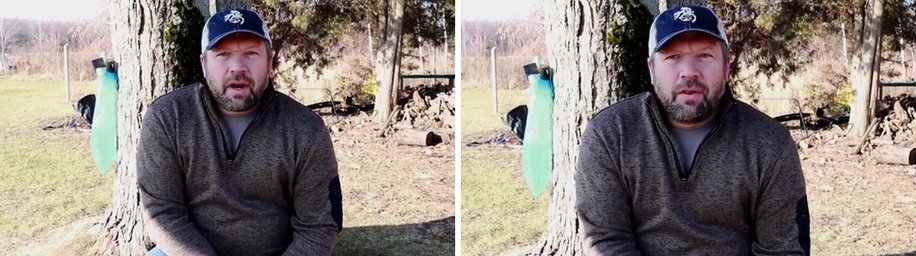}}
  {\DPOPair
    {\DPOPanelGood{%
      \DPOLine{refined\_text\_instruction}{``Replace the man wearing a blue baseball cap and a grey sweater in the center of the frame with an elderly man with a long white beard, a black hat, and a black robe.''}
      \DPOLine{subtask}{replace\_object}
      \DPOLineHiGood{image\_search}{false}
      \DPOLine{mask}{false}
    }}
    {\DPOPanelBad{%
      \DPOLine{refined\_text\_instruction}{``Replace the man wearing a blue baseball cap and a grey sweater in the center of the frame with an elderly man with a long white beard, a black hat, and a black robe.''}
      \DPOLine{subtask}{replace\_object}
      \DPOLineHiBad{image\_search}{``the man wearing a blue baseball cap and a grey sweater''}
      \DPOLine{mask}{false}
    }}
  }

  \DPOCase
  {\DPOPrompt{User prompt: ``hey i was just thinking it would be cool if you could maybe make that dress pink and put a red flower floating above her.''}\DPOFrame{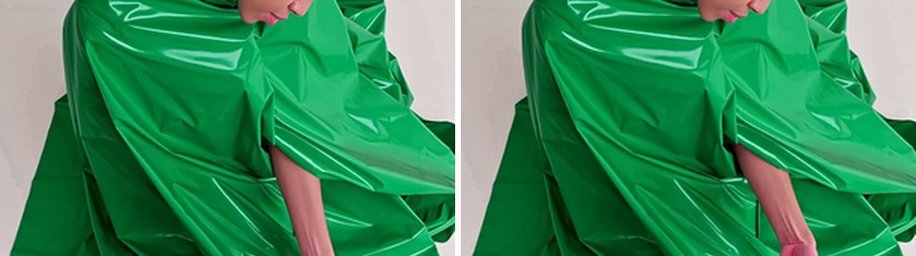}}
  {\DPOPair
    {\DPOPanelGood{%
      \DPOLineHiGood{refined\_text\_instruction}{``Change the green garment to pink and add a single floating red rose near the top of the scene.''}
      \DPOLine{subtask}{combined\_tasks}
      \DPOLine{image\_search}{false}
      \DPOLine{mask}{false}
    }}
    {\DPOPanelBad{%
      \DPOLineHiBad{refined\_text\_instruction}{``Change the green outfit to pink and add a rose near the top.''}
      \DPOLine{subtask}{combined\_tasks}
      \DPOLine{image\_search}{false}
      \DPOLine{mask}{false}
    }}
  }

  \DPOCase
  {\DPOPrompt{User prompt: ``add snow animation''}\DPOFrame{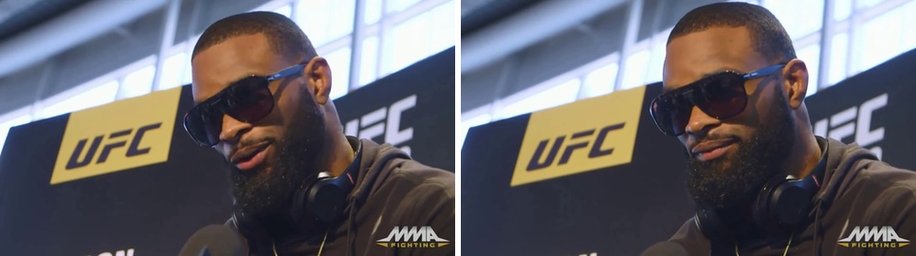}}
  {\DPOPair
    {\DPOPanelGood{%
      \DPOLineHiGood{refined\_text\_instruction}{``Apply a dynamic, snowy animation style to this video, ensuring smooth, frame-by-frame consistency. The final output should feature realistic snowfall, icy textures, and a wintery atmosphere\dots''}
      \DPOLine{subtask}{global\_style}
      \DPOLine{image\_search}{false}
      \DPOLine{mask}{false}
    }}
    {\DPOPanelBad{%
      \DPOLineHiBad{refined\_text\_instruction}{``Apply a snowy animation style.''}
      \DPOLine{subtask}{global\_style}
      \DPOLine{image\_search}{false}
      \DPOLine{mask}{false}
    }}
  }
  \end{minipage}%
  \endgroup
  \caption{DPO preference examples (2/2).}
  \label{fig:aurora_dpo_b}
\end{figure}

%% file: supp/figs/agentedit_bench_examples.tex
\definecolor{benchTagBg}{HTML}{2C3E50}
\providecommand{\benchTag}[1]{\colorbox{benchTagBg}{\textcolor{white}{\footnotesize\strut\,\texttt{#1}\,}}}

\newcommand{\benchcard}[3]{%
  \begin{minipage}[t]{\linewidth}
    \includegraphics[width=\linewidth]{#1}\\[3pt]
    \benchTag{#2}\\[1pt]
    \footnotesize\raggedright #3
  \end{minipage}%
}

\noindent
\begin{minipage}[t]{0.31\linewidth}
  \benchcard{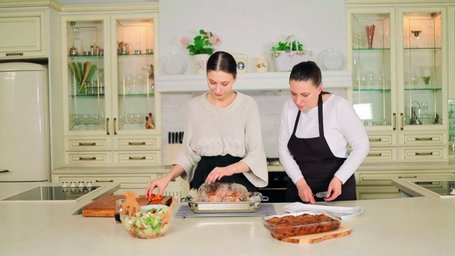}{ip\_object\_replace}{Replace the \benchRegion{glass bowl of salad on the left of the counter} with a \benchTarget{red Le~Creuset Dutch oven}.}
\end{minipage}\hfill%
\begin{minipage}[t]{0.31\linewidth}
  \benchcard{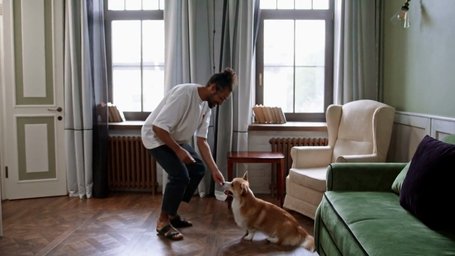}{ip\_object\_replace}{Replace the \benchRegion{white armchair in the background} with a \benchTarget{classic Eames Lounge Chair}.}
\end{minipage}\hfill%
\begin{minipage}[t]{0.31\linewidth}
  \benchcard{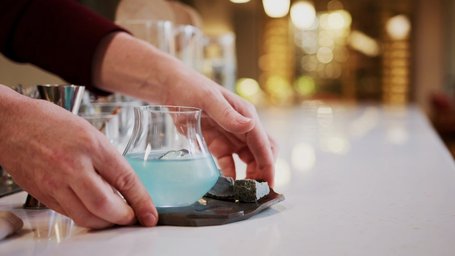}{ip\_object\_add}{Add a bottle of \benchTarget{Hendrick's Gin} on the \benchRegion{white counter to the left of the tray}.}
\end{minipage}

\vspace{8pt}

\noindent
\begin{minipage}[t]{0.31\linewidth}
  \benchcard{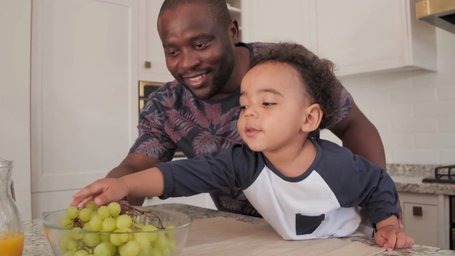}{ip\_object\_add}{Add a \benchTarget{red KitchenAid stand mixer} on the \benchRegion{counter in the background}.}
\end{minipage}\hfill%
\begin{minipage}[t]{0.31\linewidth}
  \benchcard{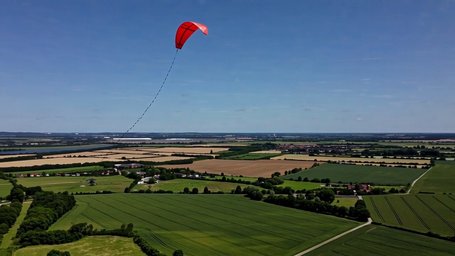}{ip\_object\_add}{Add a \benchTarget{Goodyear blimp} flying in the \benchRegion{empty blue sky to the left of the kite}.}
\end{minipage}\hfill%
\begin{minipage}[t]{0.31\linewidth}
  \benchcard{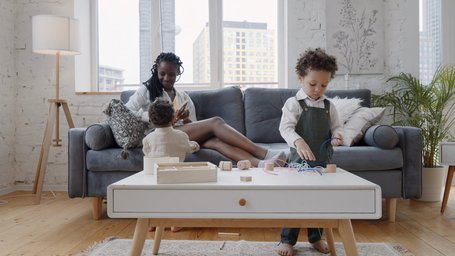}{ip\_background\_change}{Change the \benchRegion{view through the window} to a \benchTarget{New York City skyline at sunset}, keeping the interior unchanged.}
\end{minipage}

\vspace{8pt}

\noindent
\begin{minipage}[t]{0.31\linewidth}
  \benchcard{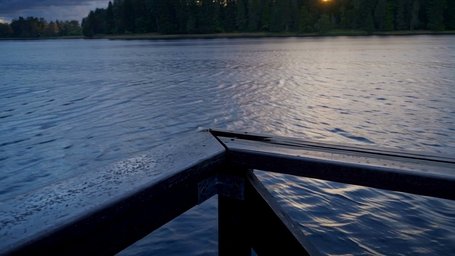}{reasoning}{Turn the lake into a \benchTarget{perfect mirror for the passing clouds} above.}
\end{minipage}\hfill%
\begin{minipage}[t]{0.31\linewidth}
  \benchcard{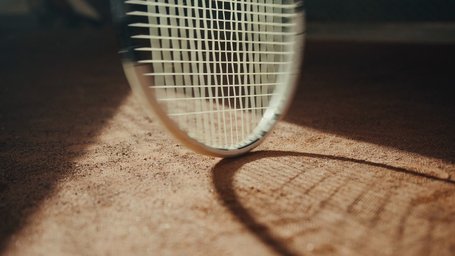}{removal}{Remove the \benchTarget{white tennis racket} on the court, including the \benchRegion{intricate shadow cast by its frame and strings} on the clay surface.}
\end{minipage}\hfill%
\begin{minipage}[t]{0.31\linewidth}
  \benchcard{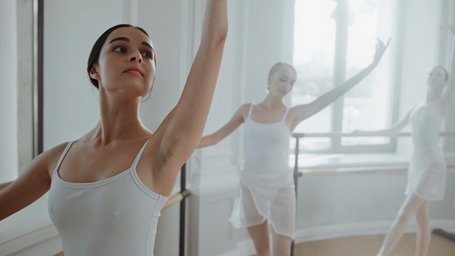}{removal}{Remove the \benchTarget{ballerina in the middle of the room} and reconstruct the \benchRegion{bright window, wall, and barre behind her}.}
\end{minipage}

%% file: supp/tables/agentedit_bench_rubric.tex
\begin{tabular}{clp{0.68\linewidth}}
\toprule
\rowcolor{benchHeader}
\benchgroup{\#} & \benchgroup{Axis} & \benchgroup{What it measures} \\
\midrule
1 & Instruction Following     & Whether the requested edit happened in the named region with the named target. The removal subset adds an anti-replacement clause: an output that replaces the target with a new object is capped at $1$. \\
2 & Edit Region Localization  & Whether the change stays inside the specified region. \\
3 & Source Preservation       & Whether unedited regions, motion, geometry, and lighting are preserved. \\
4 & Visual Quality            & Realism, lighting and shadow match, scale and perspective, absence of paste-on artifacts. \\
5 & Temporal Consistency      & Whether the edited entity would remain stable across the clip. \\
\midrule
6 & IP Presence               & Whether the named entity is visible and recognizable in roughly the right region. \\
7 & IP Identity Match         & Whether the visible entity matches the specific real-world identity (brand colors, logo, signature shape). \\
\bottomrule
\end{tabular}

%% file: supp/figs/agentedit_bench_judge_prompts.tex
\newtcolorbox{judgeprompt}[1]{%
  enhanced, breakable,
  colback=gray!4!white, colframe=gray!55!black,
  colbacktitle=gray!18!white, coltitle=black,
  boxrule=0.4pt, titlerule=0.3pt, arc=2pt,
  left=4pt, right=4pt, top=4pt, bottom=4pt,
  fonttitle=\bfseries\footnotesize,
  title={#1}}

\begin{judgeprompt}{Judge prompt: 7-axis IP rubric (IP edit types)}
\footnotesize\ttfamily
You are a meticulous video editing quality evaluator. Compare the BEFORE
frame and the AFTER frame against the editing instruction and a named
target entity, and score the edit on seven axes.\\[2pt]
Editing instruction: \{prompt\}\\
Target entity:       \{target\_entity\}\\
Edit region:         \{edit\_region\}\\[2pt]
For each axis, return an integer from 0 (worst) to 3 (best) with a brief
justification. Do not exceed 3.\\[2pt]
1. Instruction Following (0--3). Was the requested edit performed in the
named region with the named target?\\
\hspace*{1em}3: requested edit performed correctly in the right region.\\
\hspace*{1em}2: edit performed but with a minor inaccuracy or omission.\\
\hspace*{1em}1: edit partially performed or applied to the wrong region.\\
\hspace*{1em}0: instruction ignored, or the opposite was done.\\
2. Edit Region Localization (0--3). Was the change confined to the
specified region?\\
3. Source Preservation (0--3). Are subject motion, geometry and lighting
outside the edit region preserved?\\
4. Visual Quality (0--3). Realism, seamless integration, lighting and
shadow match, scale and perspective.\\
5. Temporal Consistency (0--3). Inferred from the AFTER frame, would the
edited entity remain stable across the clip without ghosting or pop-in?\\
6. IP Presence (0--3). Is the named entity actually visible and
recognizable in roughly the right region?\\
7. IP Identity Match (0--3). Does the visible entity match the specific
real-world identity, brand colors, logo, and signature shape? You may
rely only on your internal world knowledge of the brand or product.\\[2pt]
Return your evaluation in exactly this format:\\
Instruction Following: [score] - [one-sentence justification]\\
Edit Region Localization: [score] - [one-sentence justification]\\
Source Preservation: [score] - [one-sentence justification]\\
Visual Quality: [score] - [one-sentence justification]\\
Temporal Consistency: [score] - [one-sentence justification]\\
IP Presence: [score] - [one-sentence justification]\\
IP Identity Match: [score] - [one-sentence justification]\\
Total: [sum of the seven scores]
\end{judgeprompt}

\clearpage

\vspace{4pt}
\begin{judgeprompt}{Judge prompt: 5-axis non-IP rubric (reasoning, removal)}
\footnotesize\ttfamily
You are a meticulous video editing quality evaluator. Compare the BEFORE
frame and the AFTER frame against the editing instruction, and score the
edit on five axes.\\[2pt]
Editing instruction: \{prompt\}\\[2pt]
For each axis, return an integer from 0 (worst) to 3 (best) with a brief
justification. Do not exceed 3.\\[2pt]
1. Instruction Following (0--3). Was the requested edit performed?\\
\hspace*{1em}3: edit performed correctly.\\
\hspace*{1em}2: edit performed with a minor inaccuracy.\\
\hspace*{1em}1: edit partially performed or applied to the wrong region.\\
\hspace*{1em}0: instruction ignored, or the opposite was done.\\
\hspace*{1em}\emph{Removal-only clause}: if the model replaced the target
with a new object instead of removing it, give at most $1$.\\
2. Edit Region Localization (0--3). Was the change confined to the
specified region or, for global edits, to the implied scope?\\
3. Source Preservation (0--3). Are subject motion, geometry and lighting
outside the edit region preserved?\\
4. Visual Quality (0--3). Realism, seamless integration, lighting and
shadow match, scale and perspective.\\
5. Temporal Consistency (0--3). Inferred from the AFTER frame, would the
result remain stable across the clip?\\[2pt]
Return your evaluation in exactly this format:\\
Instruction Following: [score] - [one-sentence justification]\\
Edit Region Localization: [score] - [one-sentence justification]\\
Source Preservation: [score] - [one-sentence justification]\\
Visual Quality: [score] - [one-sentence justification]\\
Temporal Consistency: [score] - [one-sentence justification]\\
Total: [sum of the five scores]
\end{judgeprompt}